%% file: ms.tex
\documentclass{pbmlarxiv} \pdfoutput=1

\usepackage{cprotect}
\usepackage{xcolor}
\usepackage{footnote}\makesavenoteenv{tabular} 
\usepackage{subcaption} 
\usepackage{gnuplot-lua-tikz}

\def\recommend#1{\textsl{#1}}

\def\footurl#1{\footnote{\url{#1}}}

\def\Sref#1{Section~\ref{#1}}
\def\Tref#1{Table~\ref{#1}}
\def\Fref#1{Figure~\ref{#1}}

\def\perscite#1{\citet{#1}}
\def\parcite#1{\cite{#1}}

\begin{document}
\title{Training Tips for the Transformer Model}
\institute{}{Charles University, Faculty of Mathematics and Physics, Institute of Formal and Applied Linguistics,\\
Prague, Czechia}

\author{
  firstname=Martin,
  surname=Popel,
  corresponding=yes,
  email={popel@ufal.mff.cuni.cz},
  address={Institute of Formal and Applied Linguistics\\
           Faculty of Mathematics and Physics, Charles University\\
           Malostranské náměstí 25,
           118 00 Praha 1\\
           Czech Republic}
}
\author{firstname=Ondřej, surname=Bojar}

\PBMLmaketitle

\begin{abstract}
This article describes our experiments in neural machine translation
 using the recent Tensor2Tensor framework
 and the Transformer sequence-to-sequence model \citep{vaswani-et-al:2017}. 
We examine some of the critical parameters
 that affect the final translation quality, memory usage, training stability and training time,
 concluding each experiment with a set of recommendations for fellow researchers.
In addition to confirming the general mantra ``more data and larger models'',
 we address scaling to multiple GPUs and provide practical tips for improved training
 regarding batch size, learning rate, warmup steps, maximum sentence length and checkpoint averaging.
We hope that our observations will allow others
 to get better results given their particular hardware and data constraints.
\end{abstract}

\section{Introduction}

It has been already clearly established that neural machine translation (NMT) is
the new state of the art in machine translation, see e.g. the most recent
evaluation campaigns \parcite{bojar-EtAl:2017:WMT1,iwslt:overview:2017}. Many fundamental
changes of the underlying neural network architecture are nevertheless still
frequent and it is very difficult to predict which of the architectures has the
best combination of properties to win in the long term, considering all relevant
criteria like translation quality, model size, stability and speed of
training, interpretability but also practical availability of good
implementations.
A considerable part of a model's success in translation quality
consists in the training data, the model's sensitivity to noise in the data but also on a
wide range of hyper-parameters that affect the training. Having the right
setting of them turns out to be often a critical component for the success.

In this article, we experiment with a relatively new NMT model, called Transformer
\citep{vaswani-et-al:2017} as implemented in the Tensor2Tensor\footurl{https://github.com/tensorflow/tensor2tensor} (abbreviated T2T) toolkit, version 1.2.9.
The model and the toolkit have been released shortly after the
evaluation campaign at WMT2017\footurl{http://www.statmt.org/wmt17} and its
behavior on large-data news translation is not yet fully explored.
We want to empirically explore some of the important hyper-parameters.
Hopefully, our observations will be useful also for other
researchers considering this model and framework.

While investigations into the effect of hyper-parameters
 like learning rate and batch size are available in the deep-learning community
 \citep[e.g.][]{bottou-et-al:2016,smith:le:generalization:2017,jastrzebski-et-al:2017},
 these are either mostly theoretic or experimentally supported
 from domains like image recognition rather than machine translation.
In this article, we fill the gap by focusing exclusively on MT
 and on the Transformer model only,
 providing hopefully the best practices for this particular setting.

Some of our observations confirm the general wisdom
 (e.g. larger training data are generally better)
 and quantify the behavior on English-to-Czech translation experiments.
Some of our observations are somewhat surprising,
 e.g. that two GPUs are more than three times faster than a single GPU,
 or our findings about the interaction between maximum sentence length, learning
 rate and batch size.

The article is structured as follows.
In \Sref{sec:eval}, we discuss our evaluation methodology and main criteria:
translation quality and speed of training.
\Sref{sec:data} describes our dataset and its preparations.
\Sref{experiments} is the main contribution of the article:
 a set of commented experiments, each with a set of recommendations. 
Finally, \Sref{wmt-comparison} compares our best Transformer run with systems participating in WMT17.
We conclude in \Sref{conclusion}.

\section{Evaluation Methodology}
\label{sec:eval}

Machine translation can be evaluated in many ways and some forms of human
judgment should be always used for the ultimate resolution in any final
application. The common practice in MT research is to evaluate the model performance on a test
set against one or more human reference translations. The most widespread
automatic metric is undoubtedly the BLEU score \parcite{papineni:2002}, despite
its acknowledged problems and better-performing alternatives
\parcite{bojar-graham-kamran:2017:WMT}.
For simplicity, we stick to BLEU, too
 (we evaluated all our results also with \textsc{chrF} \citep{popovic-2015},
  but found no substantial differences from BLEU).
In particular, we use the case-insensitive
sacréBLEU\footnote{
 \url{https://github.com/awslabs/sockeye/tree/master/contrib/sacrebleu}\\
 The signature of the BLEU scores reported in this paper is
 \texttt{BLEU+case.lc+lang.en-cs+numrefs.1+smooth. exp+test.wmt13+tok.intl+version.1.2.3}.
} which uses a fixed tokenization 
(identical to \texttt{mteval-v14.pl -{}-in\-ter\-na\-tional-tok\-en\-i\-za\-tion})
and automatically downloads the reference translation for a given WMT testset.

\subsection{Considerations on Stopping Criterion}
The situation in NMT is further complicated by the fact
 that the training of NMT systems is usually non-deterministic,\footnote{
  Even if we fix the random seed (which was not done properly in T2T v1.2.9),
   a change of some hyper-parameters may affect the results
   not because of the change itself, but because it influenced the random initialization.
 }
 and (esp. with the most recent models) hardly ever converges
 or starts overfitting\footnote{
  By overfitting we mean here that the translation quality (test-set BLEU) begins to worsen,
   while the training loss keeps improving.
 } on reasonably big datasets.
This leads to learning curves that never fully flatten let alone start
 decreasing (see \Sref{sec:train-data-size}).
The common practice of machine learning to evaluate the model on a final test set
 when it started overfitting (or a bit sooner) is thus not applicable in practice.
 
Many papers in neural machine translation do not specify any stopping criteria
whatsoever. Sometimes, they mention only an approximate number of days the model
was trained for, e.g. \perscite{bahdanau:etal:attention:iclr:2015}, sometimes
the exact number of training steps is given but no indication on ``how much
converged'' the model was at that point, e.g. \perscite{vaswani-et-al:2017}.
Most probably, the training was run 
until no further improvements were clearly apparent on the development test set,
and the model was evaluated at that point.
Such an approximate stopping criterion is rather risky: it is conceivable
that different setups were stopped at different stages of training and their
comparison is not fair.

A somewhat more reliable method is to keep training for a specified number of iterations
 or a certain number of epochs.
This is however not a perfect solution either, if the models are not quite
converged at that time and the difference in their performance is not
sufficiently large. It is quite possible that e.g. a more complex model would
need a few more epochs and eventually arrived at a higher score than its
competitor.
Also, the duration of one training step (or one epoch) differs between models
 (see \Sref{sec:computation-speed}) and from the practical point of view,
 we are mostly interested in the wall-clock time.

When we tried the standard technique of early stopping,
when $N$ subsequent evaluations on the development test set do not give
improvements larger than a given delta,
we saw a big variance in the training time and final BLEU,
even for experiments with the same hyper-parameters and just a different random
seed.
Moreover to get the best results, we would have had to use a very large $N$ and
a very small delta.

\subsection{Our Final Choice: Full Learning Curves}
Based on the discussion above, we decided to report always the full learning curves
 and not just single scores.
This solution does not fully prevent the risk of premature judgments,
 but the readers can at least judge for themselves
 if they would expect any sudden twist in the results or not.

In all cases, we plot the case-insensitive BLEU score against the wall-clock time in hours.
This solution obviously depends on the hardware chosen,
 so we always used the same equipment:
 one up to eight GeForce GTX 1080 Ti GPUs with NVIDIA driver 375.66.
Some variation in the measurements is unfortunately unavoidable
 because we could not fully isolate the computation
 from different processes on the same machine and from general network traffic,
 but based on our experiments with replicated experiments
 such variation is negligible.


\subsection{Terminology}\label{sec:terminology}

For clarity, we define the following terms and adhere to them for the rest of
the paper:
\label{terms}

\begin{description}
\item[Translation quality] is an automatic estimate of how well the translation
carried out by a particular fixed model expresses the meaning of the source. We
estimate translation quality solely by BLEU score against one reference
translation.

\item[Training Steps] denote the number of iterations, i.e. the number of times
the optimizer update was run. This number also equals the number of (mini)batches that
were processed.

\item[Batch Size] is the number of training examples used by one GPU in one training step.
In sequence-to-sequence models, batch size is usually specified as the number of \emph{sentence pairs}.
However, the parameter \verb|batch_size| in T2T translation specifies the 
 approximate number of \emph{tokens} (subwords) in one batch.\footnote{
  For this purpose, the number of tokens in a sentence is defined
   as the maximum of source and target subwords.
  T2T also does reordering and bucketing of the sentences by their length
   to minimize the use of padding symbols.
  However, some padding is still needed, thus \texttt{batch\_size} only approximates
   the actual number of (non-padding) subwords in a batch.
 }
This allows to use a higher number of short sentences in one batch
 or a smaller number of long sentences.

\item[Effective Batch Size] is the number of training examples consumed in one training
step.
When training on multiple GPUs, the parameter \verb|batch_size| is interpreted per GPU.
That is, with \verb|batch_size=1500| and 8 GPUs, the system actually digests
12k subwords of each language in one step.

\item[Training Epoch] corresponds to one complete pass over the training data.
Unfortunately, it is not easy to measure the number of training epochs in
T2T.\footurl{https://github.com/tensorflow/tensor2tensor/issues/415}
T2T reports only the number of training steps.
In order to convert training steps to epochs,
 we need to multiply the steps by the effective batch size
 and divide by the number of subwords in the training data (see \Sref{sec:train-data-preproc}).
The segmentation of the training data into subwords is usually hidden to the user
 and the number of subwords must be thus computed by a special script.

\item[Computation Speed] is simply the observed number of training steps per hour.
Computation speed obviously depends on the hardware (GPU speed, GPU-CPU communication)
 and software (driver version, CUDA library version, implementation).
The main parameters affecting computation speed are the model size, optimizer
 and other settings that directly modify the formula of the neural network.

\item[Training Throughput] is the amount of training data digested by the
training. We report training throughput in subwords per hour.
Training Throughput equals to the Computation Speed multiplied by the effective batch size.

\item[Convergence Speed] or \textbf{BLEU Convergence} is the increase in BLEU
divided by time. Convergence speed changes heavily during training, starting
very high and decreasing as the training progresses. A converged model should
have convergence speed of zero.

\item[Time Till Score] is the training time needed to
 achieve a certain level of translation quality, in our case BLEU.
 We use this as an informal measure because it is not clear
  how to define the moment of ``achieving'' a given BLEU score.
 We define it as time after which the BLEU never falls below the given level.\footnote{
  Such definition of Time Till Score leads to a high variance of its values
   because of the relatively high BLEU variance between subsequent checkpoints
   (visible as a ``flickering'' of the learning curves in the figures).
  To decrease the variation one can use a bigger development test set.
 }
\item[Examples Till Score] is the number of training examples (in subwords) needed to
 achieve a certain level of BLEU.
 It equals to the Time Till Score multiplied by Training Throughput.
\end{description}

\subsection{Tools for Evaluation within Tensor2Tensor}
\label{sec:tools}

T2T, being implemented in TensorFlow, provides nice TensorBoard visualizations of the training progress. The original implementation was optimized towards speed of evaluation rather than towards following the standards of the field.
T2T thus reports ``approx-bleu'' by default, which is
computed on the internal subwords (never exposed to the user, actually)
instead of words (according to BLEU tokenization).
As a result, ``approx-bleu''
is usually about 1.2--1.8 times higher than the real BLEU. Due to its dependence on the training data (for the subword vocabulary), it is not easily reproducible in varying experiments and thus not suitable for reporting in publications.

We implemented a helper script
\texttt{t2t-bleu} which computes the ``real'' BLEU
 (giving the same result as sacréBLEU with \verb|--tokenization intl|).
Our script can be used in two ways:

\begin{itemize}
\item To evaluate one translated file:\\
   \verb|t2t-bleu --translation=my-wmt13.de --reference=wmt13_deen.de|
\item To evaluate all translations in a given directory
   (created e.g. by \verb|t2t-translate-all|)
   and store the results in a TensorBoard events file.
   All the figures in this article were created this way.
\end{itemize}

We also implemented \verb|t2t-translate-all| and \verb|t2t-avg-all| scripts,
 which translate all checkpoints in a given directory
 and average a window of N subsequent checkpoints, respectively.\footnote{
  All three scripts are now merged in the T2T master.
  All three scripts can be used while the training is still in progress,
  i.e. they wait a given number of minutes for new checkpoints to appear.}
For details on averaging see \Sref{sec:averaging}.

\section{Data Selection and Preprocessing}
\label{sec:data}

We focused on the English-to-Czech translation direction.
Most of our training data comes from the CzEng parallel treebank,
 version 1.7 (57M sentence pairs),\footnote{
  \url{http://ufal.mff.cuni.cz/czeng/czeng17},
  which is a subset of CzEng 1.6 \parcite{czeng16:2016}.
 }
 and the rest (1M sentence pairs) comes from three smaller sources
 (Europarl, News Commentary, Common Crawl)
 as detailed in Table~\ref{tab:data}.

\begin{table}\centering
\begin{tabular}{lccc}
                    & sentences & EN words & CS words \\\hline
CzEng 1.7           & 57 M      & 618 M    & 543 M  \\
europarl-v7         & 647 k     & 15 M     & 13 M   \\
news-commentary-v11 & 190 k     & 4.1 M    & 3.7 M  \\
commoncrawl         & 161 k     & 3.3 M    & 2.9 M  \\\hline
Total               & 58 M      & 640 M    & 563 M  \\\hline
\end{tabular}
\caption{Training data resources}\label{tab:data}
\end{table}

We use this dataset of 58M sentence pairs for most our experiments.
In some experiments (in Sections \ref{sec:train-data-size} and \ref{sec:learning-rate}),
 we substitute CzEng 1.7 with an older and considerably smaller CzEng~1.0 \parcite{czeng10:lrec2012}
 containing 15M sentence pairs (233M/206M of en/cs words).

To plot the performance throughout the training, we use WMT newstest2013
 as a development set (not overlapping with the training data).
In \Sref{wmt-comparison}, we apply our best model
 (judged from the performance on the development set) to the WMT newstest2017,
 for comparison with the state-of-the-art systems.

\subsection{Training Data Preprocessing}
\label{sec:train-data-preproc}

Data preprocessing such as tokenization and truecasing has always been a very important part of the setup of statistical machine translation systems.
A huge leap in scaling NMT to realistic data size has been achieved by the introduction of subword units \parcite{sennrich-haddow-birch:2016:P16-12}, but
the long-term vision of the deep-learning community is to leave all these ``technicalities'' up to the trained neural network and feed it with as original input as possible
 \citep[see e.g.][]{lee:etal:charlevel:2016}.

T2T adopts this vision and while it supports the use of external subword units, it comes with its own built-in method similar to the word-piece algorithm by \perscite{google:bridging:gap:2016:arxiv} and does not expect the input to be even tokenized. Based on a small sample of the training data, T2T will train a subword vocabulary and apply it to all the training and later evaluation data.

We follow the T2T default and provide raw plain text training sentences.
We use the default parameters: shared source and target (English and Czech) subword vocabulary
 of size 32k.\footnote{
  More details on T2T with BPE subword units by \perscite{sennrich-haddow-birch:2016:P16-12}
   vs. the internal implementation can be found in the technical report
   ``Morphological and Language-Agnostic Word Segmentation for NMT'' attached to the Deliverable 2.3
   of the project QT21: \url{http://www.qt21.eu/resources/}.
 }
After this preprocessing, the total number of subwords in our main training data is 992 millions
 (taking the maximum of English and Czech lengths for each sentence pair,
  as needed for computing the number of epochs, see \Sref{sec:terminology}).
The smaller dataset CzEng~1.0 has 327 million subwords.
In both cases the average number of subwords per (space-delimited) word is about 1.5.
%

Even when following the defaults, there are some important details that should be considered. We thus provide our first set of technical tips here:

\subsubsection*{Tips on Training Data Preprocessing}
\begin{itemize}
\item Make sure that the subword vocabulary is trained on a sufficiently large sample of the training data.\cprotect\footnote{%
  This is controlled by a \verb|file_byte_budget| constant,
   which must be changed directly in the source code in T2T v1.2.9.
  A sign of too small training data for the subword vocabulary is
   that the \verb|min_count| as reported in the logs is too low,
   so the vocabulary is estimated from words seen only once or twice.
 }
\item As discussed in Section~\ref{sec:batch-size}, a higher batch size may be beneficial for the training
  and the batch size can be higher when excluding training sentences longer than a given threshold.
 This can be controlled with parameter \verb|max_length| (see \Sref{sec:max-len}),
  but it may be a good idea to exclude too long sentences
  even before preparing the training data using \verb|t2t-datagen|.
 This way the TFRecords training files will be smaller and their processing a bit faster.\footnote{
  We did no such pre-filtering in our experiments.
 }
\end{itemize}

\section{Experiments}
\label{experiments}

In this section, we present several experiments, always summarizing the
 observations and giving some generally applicable tips that we learned.
All experiments were done with T2T v1.2.9 unless stated otherwise.

We experiment with two sets of hyper-parameters pre-defined in T2T:
 \texttt{trans\-for\-mer\_big\_sin\-gle\_gpu} (BIG) and \texttt{trans\-for\-mer\_ba\-se\_sin\-gle\_gpu} (BASE),
 which differ mainly in the size of the model.
Note that \texttt{trans\-for\-mer\_big\_sin\-gle\_gpu} and \texttt{trans\-for\-mer\_ba\-se\_sin\-gle\_gpu}
 are just names of a set of hyper-parameters,
 which can be applied even when training on multiple GPUs, as we do in our experiments, see Section~\ref{sec:gpus}.\footnote{
  According to our experiments (not reported here),
   \texttt{trans\-for\-mer\_big\_sin\-gle\_gpu} is better than \texttt{trans\-for\-mer\_big}
   even when training on 8 GPUs,
   although the naming suggests that the T2T authors had an opposite experience.
}

Our baseline setting uses the BIG model with its default hyper-parameters except
for:
\begin{itemize}
\item \verb|batch_size=1500| (see the discussion of different sizes in
\Sref{sec:batch-size}),
\item \verb|--train_steps=6000000|, i.e. high enough, so we can stop each
experiment manually as needed,
\item \verb|--save_checkpoints_secs=3600| which forces checkpoint saving each
hour (see \Sref{sec:averaging}),
\item \verb|--schedule=train| which disables the internal evaluation with \verb|approx_bleu|
      and thus makes training a bit faster (see \Sref{sec:eval}).\cprotect\footnote{%
      Also there are some problems with the alternative schedules
        \verb|train_and_evaluate| (it needs more memory) and \verb|continuous_train_and_eval|
        (see \url{https://github.com/tensorflow/tensor2tensor/issues/556}).
      }
\end{itemize}

\subsection{Computation Speed and Training Throughput}
\label{sec:computation-speed}

We are primarily interested in the translation quality (BLEU learning curves and Time Till Score)
 and we discuss it in the following sections \ref{sec:train-data-size}--\ref{sec:averaging}.
In this section, we focus however only on the \emph{computation speed} and \emph{training throughput}.
Both are affected by three important factors: batch size, number of used GPUs and model size.
The speed is usually almost constant for a given experiment.\footnote{
 TensorBoard shows \texttt{global\_step/sec} statistics, i.e. the computation speed curve.
 These curves in our experiments are almost constant for the whole training
  with variation within 2\%, except for moments
  when a checkpoint is being saved (and the computation speed is thus much slower).
}

\begin{table}[htb]\centering
\begin{subfigure}[b]{0.45\textwidth}
 \begin{tabular}{r|rr}\hline
            & \multicolumn{2}{c}{model} \\
 batch\_size&  BASE & BIG \\\hline
  500       & 43.4k & 23.6k \\
 1000       & 30.2k & 13.5k \\
 1500       & 22.3k &  9.8k \\
 2000       & 16.8k &  7.5k \\
 2500       & 14.4k &  6.5k \\
 3000       & 12.3k & OOM \\
 4500       &  8.2k & OOM \\
 6000       &  6.6k & OOM \\\hline
 \end{tabular}
 \caption{Computation speed (steps/hour)}
 \label{tab:speed-a}
\end{subfigure}
\begin{subfigure}[b]{0.45\textwidth}
 \begin{tabular}{r|rr}\hline
            & \multicolumn{2}{c}{model} \\
 batch\_size&  BASE & BIG \\\hline
  500       & 21.7M & 11.9M \\
 1000       & 30.2M & 13.5M \\
 1500       & 33.4M & 14.7M \\
 2000       & 33.7M & 15.0M \\
 2500       & 36.0M & 16.2M \\
 3000       & 37.0M & OOM \\
 4500       & 36.7M & OOM \\
 6000       & 39.4M & OOM \\\hline
 \end{tabular}
 \caption{Training throughput (subwords/hour)}
 \label{tab:speed-b}
\end{subfigure}
\caption{Computation speed and training throughput for a single GPU.}
\label{tab:speed}
\end{table}

\Tref{tab:speed} shows the computation speed and training throughput for a single GPU
 and various batch sizes and model sizes (BASE and BIG).
The BASE model allows for using a higher batch size than the BIG model.
The cells where the BIG model resulted in out-of-memory errors are marked with ``OOM''.\footnote{
  For these experiments, we used \texttt{max\_length=50} in order to be able to test bigger batch sizes.
  However, in additional experiments we checked that \texttt{max\_length} does not affect the training throughput itself.
 }
We can see that the computation speed decreases with increasing batch size
 because not all operations in GPU are fully batch-parallelizable.
The training throughput grows sub-linearly with increasing batch size,
 so based on these experiments only, there is just a small advantage
 when setting the batch size to the maximum value.
We will return to this question in \Sref{sec:batch-size},
 while taking into account the translation quality.

We can also see the BASE model has approximately two times bigger throughput as well as computation speed
 relative to the BIG model.

\begin{table}[htb]\centering
\begin{tabular}{rrr}\hline
GPUs & steps/hour & subwords/hour \\\hline
1    &       9.8k & 14.7M \\
2    &       7.4k & 22.2M \\
6    &       5.4k & 48.6M \\
8    &       5.6k & 67.2M \\\hline
\end{tabular}
\caption{Computation speed and training throughput for various numbers of GPUs,
 with the BIG model and \texttt{batch\_size=1500}.}
\label{tab:speed-gpus}
\end{table}

\Tref{tab:speed-gpus} uses the BIG model and \texttt{batch\_size=1500},
 while varying the number of GPUs.
The overhead in GPU synchronization is apparent from the decreasing computation speed.
Nevertheless, the training throughput still grows with more GPUs,
 so e.g. with 6 GPUs we process 3.2 times more training data per hour relative to a single GPU
 (while without any overhead we would hypothetically expect 6 times more data).

The overhead when scaling to multiple GPUs is smaller
 than the overhead when scaling to a higher batch size.
Scaling from a single GPU to 6 GPUs increases the throughput 3.2 times,
 but scaling from batch size 1000 to 6000 on a single GPU increases the throughput 1.3 times.

\subsection{Training Data Size}
\label{sec:train-data-size}

\begin{figure}
\input{plots/8GPU-czeng1-vs-czeng57.tex}\vspace{-5mm}
\cprotect\caption{Training data size effect.
 BLEU learning curves for our main training dataset with 58 million sentence pairs
 and an alternative training dataset with 16 million sentence pairs.
 Both trained with 8 GPUs, BIG model and \verb|batch_size=1500|.
}
\label{fig:czeng1-vs-czeng57}
\end{figure}
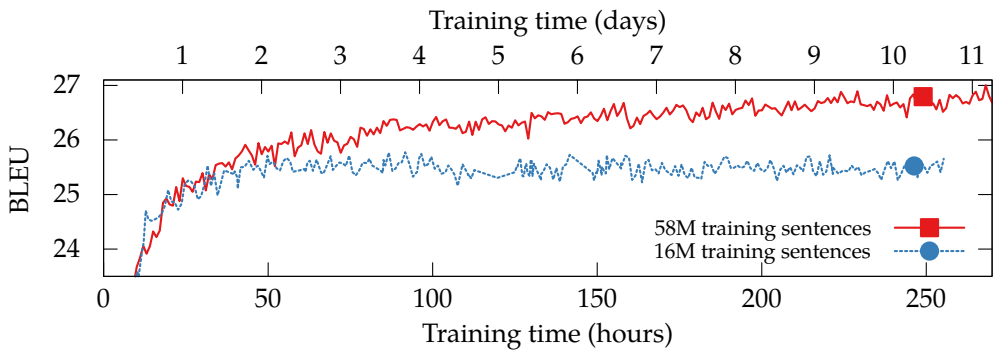

For this experiment, we substituted CzEng 1.7 with CzEng 1.0 in the training data,
 so the total training size is 16 million sentence pairs (255M / 226M of English/Czech words).
Figure~\ref{fig:czeng1-vs-czeng57} compares the BLEU learning curves of two experiments
 which differ only in the training data: the baseline CzEng 1.7 versus the smaller CzEng 1.0.
Both are trained on the same hardware with the same hyper-parameters
 (8 GPUs, BIG, \verb|batch_size=1500|).
Training on the smaller dataset (2.5 times smaller in the number of words) converges
 to BLEU of about 25.5 after two days of training and does not improve over the next week of training.
Training on the bigger dataset gives slightly worse results
 in the first eight hours of training (not shown in the graph)
 but clearly better results after two days of training,
 reaching over 26.5 BLEU after eight days.\footnote{
  We compared the two datasets also in another experiment with two GPUs,
   where CzEng 1.7 gave slightly worse results than CzEng 1.0 during the first two days of training
   but clearly better results after eight days.
  We hypothesize CzEng 1.0 is somewhat cleaner than CzEng 1.7.
 }

With \verb|batch_size=1500| and 8 GPUs,
 training one epoch of the smaller dataset (with CzEng 1.0) takes 27k steps (5 hours of training),
 compared to 83k steps (15 hours) for the bigger dataset (with CzEng 1.7).
This means \recommend{about 10 epochs in the smaller dataset were needed for reaching the convergence}
 and this is also the moment when the bigger dataset starts being clearly better.
However, \recommend{even 18 epochs in the bigger dataset were not enough to reach the convergence}.
\textit{enough to reach the convergence}

\subsubsection*{Tips on Training Data Size}
\begin{itemize}
  \item For comparing different datasets (e.g. smaller and cleaner vs. bigger and noisier),
    we need to train long enough because
    \recommend{results after first hours (or days if training on a single GPU) may be misleading}.
  \item For large training data (as CzEng 1.7 which has over half a gigaword),
   \recommend{BLEU improves even after one week of training on eight GPUs}
    (or after 20 days of training on two GPUs in another experiment).
  \item \recommend{We cannot easily interpolate one dataset results to another dataset.}
   While the smaller training data (with CzEng 1.0) converged after 2 days,
    the main training data (with CzEng 1.7), which is 2.5 times bigger,
    continues improving even after 2.5$\times$2 days.\footnote{
     Although such an expectation may seem na\"ive, we can find it in literature.
     For example, \citet{bottou-2012} in Section 4.2 writes:
     ``\emph{Expect the validation performance to plateau after a number of epochs roughly comparable
     to the number of epochs needed to reach this point on the small training set.}''
    }
\end{itemize}

\subsection{Model Size}
\label{sec:model-size}

Choosing the right model size is important for practical reasons:
 larger models may not fit any more on your GPU or they may require to use a very small batch size.

We experiment with two models,\footnote{
 We tried also a model three times as large as BASE (1.5 times as large as BIG),
  but it did not reach better results than BIG, so we don't report it here.
}
 as pre-defined in Tensor2Tensor --
 \texttt{trans\-for\-mer\-\_big\-\_sin\-gle\-\_gpu} (BIG) and
 \texttt{trans\-for\-mer\_ba\-se\_sin\-gle\_gpu} (BASE),
 which differ in four hyper-parameters summarized in \Tref{tab:big-vs-base}.

\begin{table}[htb]\centering
\begin{tabular}{lrrrr}\hline
model & hidden\_size & filter\_size & num\_heads & adam\_beta2 \\\hline
BASE  &          512 &         2048 &          8 & 0.980 \\
BIG   &         1024 &         4096 &         16 & 0.998 \\\hline
\end{tabular}
\caption{\texttt{transformer\_big\_single\_gpu} (BIG)
and \texttt{transformer\_base\_single\_gpu} (BASE) hyper-parameter differences.}
\label{tab:big-vs-base}
\end{table}

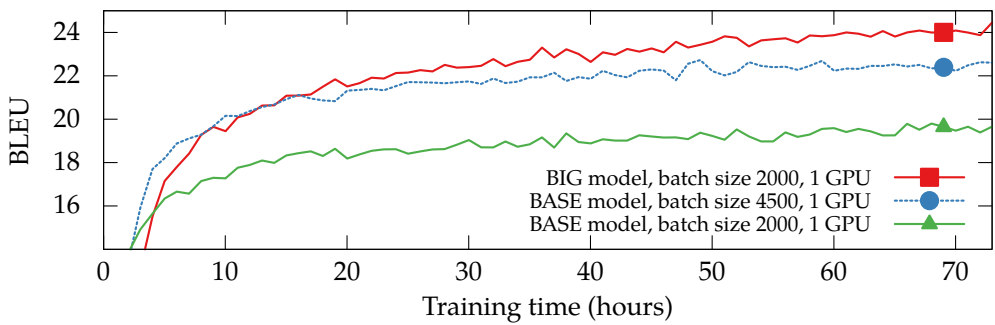
\begin{figure}
\input{plots/1GPU-big2000-base4500-base2000.tex}\vspace{-5mm}
\caption{Effect of model size and batch size on a single GPU.}
\label{fig:1GPU-big2000-base4500-base2000}
\end{figure}

Figure~\ref{fig:1GPU-big2000-base4500-base2000} shows
 that on a single GPU, the BIG model becomes clearly better than the BASE model
 after 4 hours of training if we keep the batch size the same -- 2000
 (and we have confirmed it with 1500 in other experiments).
However, the BASE model takes less memory, so we can afford a higher batch size,
 in our case 4500 (with no \verb|max_length| restriction, see the next section),
 which improves the BLEU (see \Sref{sec:batch-size}).
But even so, after less than one day of training, BIG with batch size 2000 becomes better
 than BASE with batch size 4500 (or even 6000 with \verb|max_length|=70 in another experiment)
 and the difference grows up to 1.8 BLEU after three days of training.

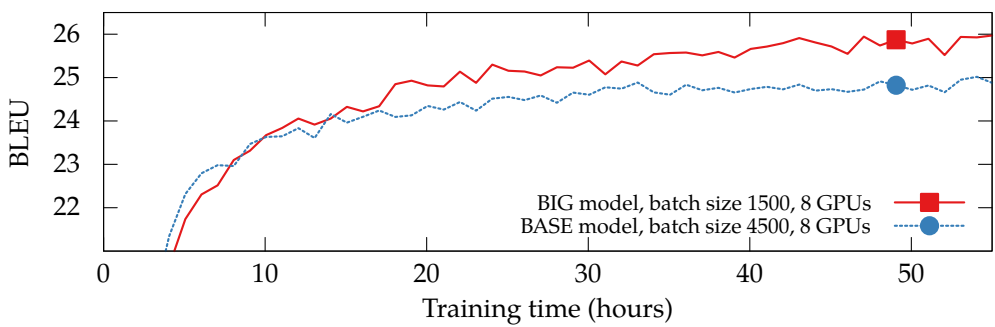
\begin{figure}
\input{plots/8GPU-big1500-base4500.tex}\vspace{-5mm}
\caption{Effect of model size and batch size on 8 GPUs.}
\label{fig:8GPU-big1500-base4500}
\end{figure}

Figure~\ref{fig:8GPU-big1500-base4500} confirms this with 8 GPUs
 -- here BIG with batch size 1500 becomes clearly better than BASE with batch size 4500
 after 18 hours of training.

\subsubsection*{Tips on Model Size}
\begin{itemize}
\item \recommend{Prefer the BIG over the BASE model}
 if you plan to train longer than one day and have 11 GB (or more) memory available on GPU.
\item With less memory you should benchmark BIG and BASE with the maximum possible batch size.
\item For fast debugging (of model-size-unrelated aspects) use a model called \texttt{trans\-for\-mer\_ti\-ny}.
\end{itemize}

\subsection{Maximum Training Sentence Length}\label{sec:max-len}
The parameter \verb|max_length| specifies the maximum length of a sentence in subwords.
Longer sentences (either in source or target language) are excluded from the training completely.
If no \verb|max_length| is specified (which is the default),
 \verb|batch_size| is used instead.
Lowering the \verb|max_length| allows to use a higher batch size or a bigger model.
Since the Transformer implementation in T2T can suddenly run out of memory
 even after several hours of training,
 it is good to know how large batch size fits in your GPU.
\Tref{batch-sizes-for-sent-length} presents what we empirically measured for the BASE and BIG
 models with Adam and Adafactor\footnote{
  The Adafactor optimizer \citep{adafactor} is available only in T2T 1.4.2 or newer and
   has three times smaller models than Adam
   because it does not store first and second moments for all weights.
  We leave further experiments with Adafactor for future work.
 } optimizers and various \verb|max_length| values.

\begin{table}\centering
\begin{tabular}{r|ccc|cc}\hline
            &  \multicolumn{3}{c}{maximum batch size} & \multicolumn{2}{|c}{longer sentences}\\
max\_length & BIG+Adam & BIG+Adafactor & BASE+Adam & train & test\\\hline
none        & 2040     & 2550          & 4950      & 0.0\% & 0.0\% \\
150         & 2230     & 2970          & 5430      & 0.2\% & 0.0\% \\
100         & 2390     & 3280          & 5990      & 0.7\% & 0.3\% \\
70          & 2630     & 3590          & 6290      & 2.1\% & 2.2\% \\
50          & 2750     & 3770          & 6430      & 5.0\% & 9.1\% \\\hline
\end{tabular}
\caption{Maximum batch size which fits into 11GB memory for various combinations of
 \texttt{max\_length} (maximum sentence length in subwords),
 model size (base or big)
 and optimizer (Adam or Adafactor).
 The last two columns show the percentage of sentences in the train (CzEng~1.7) and test (wmt13) data
  that are longer than a given threshold.
}
\label{batch-sizes-for-sent-length}
\end{table}

\begin{figure}
\input{plots/1GPU-max-length-b1500.tex}\vspace{-5mm}
\cprotect\caption{Effect of restricting the training data to various \verb|max_length| values.
 All trained on a single GPU with the BIG model and \verb|batch_size|=1500.
 An experiment without any \verb|max_length| is not shown,
  but it has the same curve as \verb|max_length|=400.
}
\label{fig:max-length1500}
\end{figure}
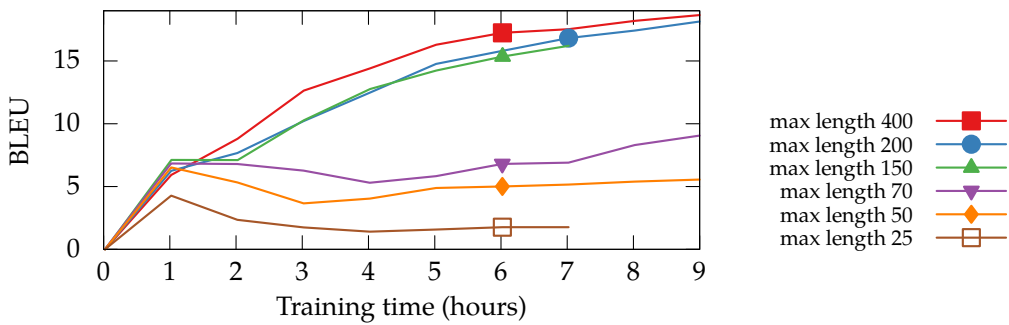

Setting \verb|max_length| too low would result in excluding too many training sentences
 and biasing the translation towards shorter sentences,
 which would hurt the translation quality.
The last two columns in \Tref{batch-sizes-for-sent-length} show
 that setting \verb|max_length| to 70 (resp. 100) results in excluding only
 2.1\% (resp. 0.7\%) of sentences in the training data,
 and only 2.2\% (resp. 0.3\%) sentences in the development test data are longer,
 so the detrimental effect of smaller training data and length bias should be minimal in this setting.
However, our experiments with \verb|batch_size|=1500 in \Fref{fig:max-length1500}
 show a strange drop in BLEU after one hour of training for all experiments with \verb|max_length| 70 or lower.
Even with \verb|max_length| 150 or 200 the BLEU learning curve is worse than with \verb|max_length|=400,
 which finally gives the same result as not using any \verb|max_length| restriction.
The training loss of \verb|max_length|=25 (and 50 and 70) has high variance
 and stops improving after the first hour of training
 but shows no sudden increase
 (as in the case of diverged training discussed in \Sref{sec:learning-rate} when the learning rate is too high).
We have no explanation for this phenomenon.\footnote{
 \url{https://github.com/tensorflow/tensor2tensor/issues/582}
}

We did another set of experiments with varying \verb|max_length|,
 but this time with \verb|batch_size|=2000 instead of 1500.
In this case, \verb|max_length| 25 and 50 still results in slower growing BLEU curves,
 but 70 and higher has the same curve as no \verb|max_length| restriction.
So in our case, \recommend{if the batch size is high enough, the \texttt{max\_length} has almost no effect on BLEU},
 but this should be checked for each new dataset.

We trained several models with various \verb|max_length| for three days
 and observed that \recommend{they are not able to produce longer translations
 than what was the maximum length used in training},
 even if we change the decoding parameter alpha.

\subsubsection*{Tips on \texttt{max\_length}}
\begin{itemize}
\item \recommend{Set (a reasonably low) \texttt{max\_length}}.
 This allows to use a higher batch size and prevents out-of-memory errors after several hours of training.
 Also, with a higher percentage of training sentences that are almost \verb|max_length| long,
  there is a higher chance
  that the training will fail either immediately (if the batch size is too high) or never (otherwise).,
\item \recommend{Set a reasonably high \texttt{max\_length}}.
 Consider the percentage of sentences excluded from training and from the targeted development test set
 and also watch for unexpected drops (or stagnations) of the BLEU curve in the first hours of training.
\end{itemize}

\subsection{Batch Size}\label{sec:batch-size}
The default \verb|batch_size| value in recent T2T versions is 4096 subwords for all models
 except for \verb|transformer_base_single_gpu|, where the default is 2048.
However, we recommend to always set the batch size explicitly\cprotect\footnote{ 
  e.g. \scriptsize\verb|--hparams="batch_size=1500,learning_rate=0.20,learning_rate_warmup_steps=16000"|\\
  As the batch size is specified in subwords, we see no advantage in using power-of-two values.
 }
 or at least make a note what was the default in a given T2T version when reporting experimental results.

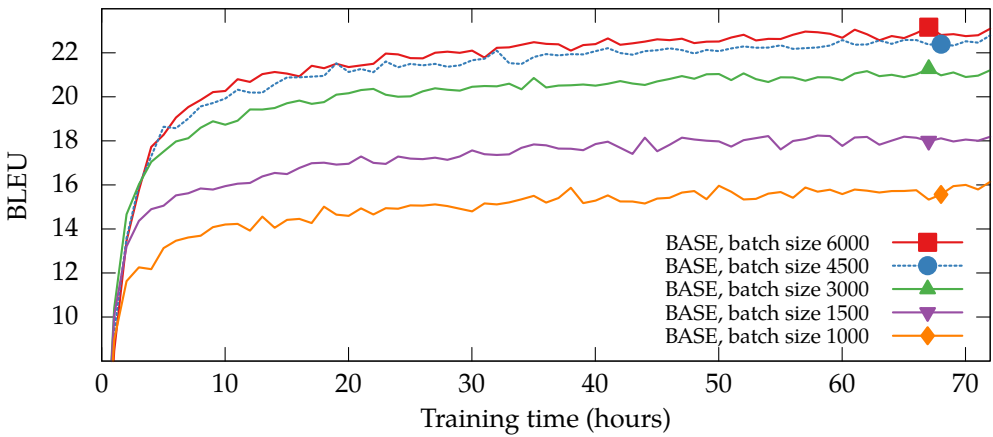
\begin{figure}
\input{plots/1GPU-base-m70-batch-size.tex}\vspace{-5mm}
\caption{Effect of the batch size with the BASE model. All trained on a single GPU.}
\label{fig:base-batch-size}
\end{figure}

Figure~\ref{fig:base-batch-size} shows learning curves for five different batch sizes
 (1000, 1500, 3000, 4500 and 6000) for experiments with a single GPU and the BASE model.\cprotect\footnote{All
   the experiments in \Fref{fig:base-batch-size} use \verb|max_length|=70,
   but we have got the same curves when re-running without any \verb|max_length| restrictions,
   except for \verb|batch_size|=6000 which failed with OOM.
 } 
A higher batch size \emph{up to 4500} is clearly better in terms of BLEU 
 as measured by Time Till Score
 and Examples Till Score metrics defined in \Sref{sec:computation-speed}.
For example, to get over BLEU of 18
 with \verb|batch_size|=3000, we need 7 hours (260M examples),
 and with \verb|batch_size|=1500, we need about 3 days (2260M examples)
 i.e. 10 times longer (9 time more examples).
From \Tref{tab:speed-a} we know that bigger batches have slower computation speed,
 so when re-plotting \Fref{fig:base-batch-size} with steps instead of time on the x-axis,
 the difference between the curves would be even bigger.
From \Tref{tab:speed-b} we know that bigger batches have slightly higher training throughput,
 so when re-plotting with number of examples processed on the x-axis,
 the difference will be smaller, but still visible.
The only exception is the difference between batch size 4500 and 6000,
 which is very small and can be fully explained by the fact
 that batch size 6000 has 7\% higher throughput than batch size 4500.

So \recommend{for the BASE model, a higher batch size gives better results}, although with diminishing returns.
This observation goes against the common knowledge in other NMT frameworks and
 deep learning in general \parcite{keskar:etal:minibatches:iclr:2017}
 that smaller batches proceed slower (training examples per hour)
 but result in better generalization (higher test-set BLEU) in the end.
In our experiments with the BASE model in T2T,
 bigger batches are not only faster in training throughput (as could be expected),
 but also faster in convergence speed, Time Till Score and Examples Till Score.

\begin{figure}
\input{plots/1GPU-big-m70-batch-size.tex}\vspace{-5mm}
\caption{Effect of the batch size with the BIG model. All trained on a single GPU.}
\label{fig:1GPU-czeng57-batch-size}
\end{figure}
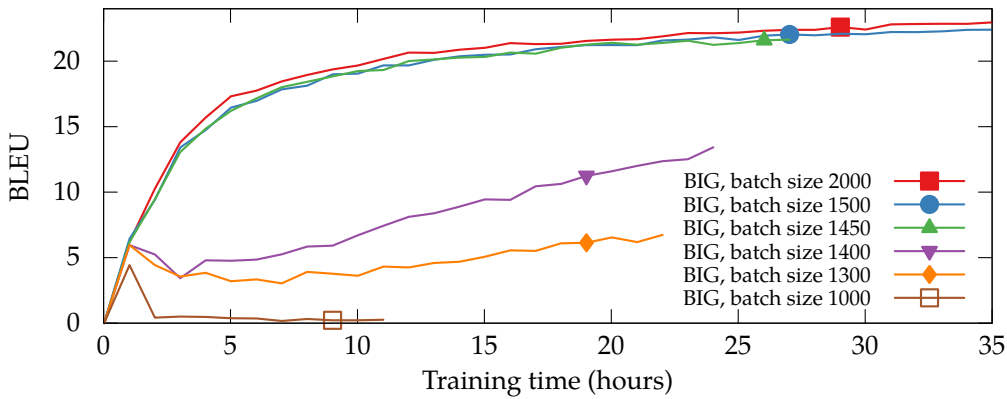

Interestingly, when replicating these experiments \recommend{with the BIG model,
 we see quite different results}, as shown in \Fref{fig:1GPU-czeng57-batch-size}.
The BIG model needs a certain minimal batch size to start converging at all,
 but for higher batch sizes there is almost no difference in the BLEU curves
 (but still, bigger batch never makes the BLEU worse in our experiments).
In our case, the sharp difference is between batch size 1450, which trains well,
 and 1400, which drops off after two hours of training, recovering only slowly.

According to \citet{smith:le:generalization:2017} and \citet{smith:etal:lr:batchsize:arxiv:2017},
 the \emph{gradient noise scale}, i.e. scale of random fluctuations in the SGD (or Adam etc.) dynamics,
 is proportional to learning rate divided by the batch size (cf. \Sref{sec:scaling}).
Thus when lowering the batch size, we increase the noise scale
 and the training may \emph{diverge}.
This may be either permanent,
 as in the case of batch size 1000 in \Fref{fig:1GPU-czeng57-batch-size},
 or temporary, as in the case of batch size 1300 and 1400,
 where the BLEU continues to grow after the temporary drop,
 but much more slowly than the non-diverged curves.

We are not sure what causes the difference between the BASE and BIG models
 with regards to the sensitivity to batch size.
One hypothesis is that the BIG model is more difficult to initialize
 and thus more sensitive to divergence in the early training phase.
Also while for BASE, increasing the batch size was highly helpful until 4500,
 for BIG this limit may be below 1450,
 i.e. below the minimal batch size needed for preventing diverged training.

\subsubsection*{Tip on Batch Size}
\begin{itemize}
\item \recommend{Batch size should be set as high as possible}
      while keeping a reserve for not hitting the out-of-memory errors.
      It is advisable to establish the largest
      possible batch size before starting the main and long training.
\end{itemize}

\subsection{Learning Rate and Warmup Steps on a Single GPU}\label{sec:learning-rate}

\begin{figure}
\input{plots/1GPU-czeng1-learning-rate.tex}\vspace{-5mm}
\caption{Effect of the learning rate on a single GPU.
 All trained on CzEng~1.0 with the default batch size (1500) and warmup steps (16k).}
\label{fig:learning-rate}
\end{figure}
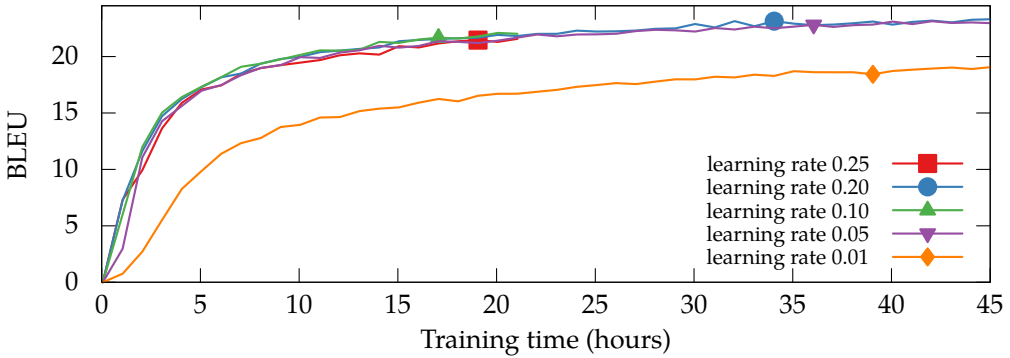

The default learning rate in T2T translation models is 0.20.
Figure~\ref{fig:learning-rate} shows that varying the value within range 0.05--0.25
 makes almost no difference.
Setting the learning rate too low (0.01) results in notably slower convergence.
Setting the learning rate too high (0.30, not shown in the figure)
 results in \emph{diverged} training,
 which means in this case that the learning curve starts growing as usual,
 but at one moment drops down almost to zero
 and stays there forever.

A common solution to prevent diverged training is
 to decrease the \texttt{learning\_ra\-te} parameter
 or increase \verb|learning_rate_warmup_steps|
 or introduce gradient clipping.
The \texttt{learning\_rate\_warmup\_steps} parameter 
 configures a \texttt{linear\_warmup\_rsqrt\_decay} schedule\footnote{
 The schedule was called \texttt{noam} in T2T versions older than 1.4.4.
}
 and it is set to  16~000 by default (for the BIG model),
 meaning that within the first 16k steps the learning rate grows linearly
 and then follows an inverse square root decay ($t^{-0.5}$, cf. \Sref{sec:lr-schedule}).
At 16k steps, the actual learning rate is thus the highest.

If a divergence is to happen, it usually happens within the first few hours of
training, when the actual learning rate becomes the highest.
Once we increased the warmup steps from 16k to 32k,
 we were able to train with the learning rate of 0.30 and even 0.50 without any divergence.
The learning curves looked similarly to the baseline one
 (with default values of 16k warmup steps and learning rate 0.20).
When trying learning rate 1.0, we had to increase warmup steps to 60k
 (with 40k the training diverged after one hour)
 -- this resulted in a slower convergence at first
 (about 3 BLEU lower than the baseline after 8 hours of training),
 but after 3--4 days of training having the same curve as the baseline.

\begin{figure}
\input{plots/1GPU-czeng1-warmup-steps.tex}\vspace{-5mm}
\caption{Effect of the warmup steps on a single GPU.
 All trained on CzEng~1.0 with the default batch size (1500) and learning rate (0.20).}
\label{fig:warmup-steps}
\end{figure}
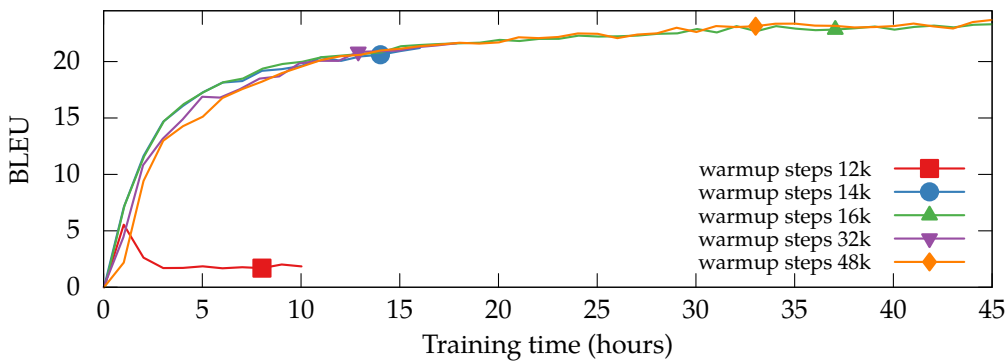

Figure~\ref{fig:warmup-steps} shows the effect of different warmup steps
 with a fixed learning rate (the default 0.20).
Setting warmup steps too low (12k) results in diverged training.
Setting them too high (48k, green curve) results in a slightly slower convergence at first,
 but matching the baseline after a few hours of training.

We can conclude that for a single GPU and the BIG model,
 there is a relatively large range of learning rate and warmup steps values
 that achieve the optimal results.
The default values \verb|learning_rate|=0.20 and \verb|learning_rate_warmup_steps|=16000 are within this range.

\subsection*{Tips on Learning Rate and Warmup Steps}
\begin{itemize}
\item \recommend{In case of diverged training, try gradient clipping and/or more warmup steps.}
\item If that does not help (or if the warmup steps are too high relative to the expected total training steps),
 try decreasing the learning rate.
\item Note that when you decrease warmup steps (and keep learning rate),
 you also increase the maximum actual learning rate
 because of the way how the \texttt{linear\_warmup\_rsqrt\_decay} (aka \texttt{noam}) schedule
 is implemented.\footnote{This holds at least in T2T versions 1.2.9--1.5.2,
  but as it is somewhat unexpected/unintuitive for some users,
  it may be fixed in future, see \url{https://github.com/tensorflow/tensor2tensor/issues/517}.
 } 
\end{itemize}

\subsection{Number of GPUs}\label{sec:gpus}

T2T allows to train with multiple GPUs on the same machine
 simply using the parameter \verb|--worker_gpus|.\cprotect\footnote{ %
  and making sure environment variable \verb|CUDA_VISIBLE_DEVICES| is set so enough cards are visible.
  T2T allows also distributed training (on multiple machines),
   but we have not experimented with it.
  Both single-machine multi-gpu and distributed training use synchronous Adam updates by default.
  }
As explained in Section~\ref{terms},
  the parameter \verb|batch_size| is interpreted per GPU,
  so with 8 GPUs, the \textit{effective batch size} is 8 times bigger.

A single-GPU experiment with batch size 4000,
 should give exactly the same results as two GPUs and batch size 2000
 and as four GPUs and batch size 1000
 because the effective batch size is 4000 in all three cases.
We have confirmed this empirically.
By the ``same results'' we mean BLEU (or train loss) versus training steps on the x-axis.
When considering time, the four-GPU experiment will be the fastest one,
 as explained in \Sref{sec:computation-speed}.

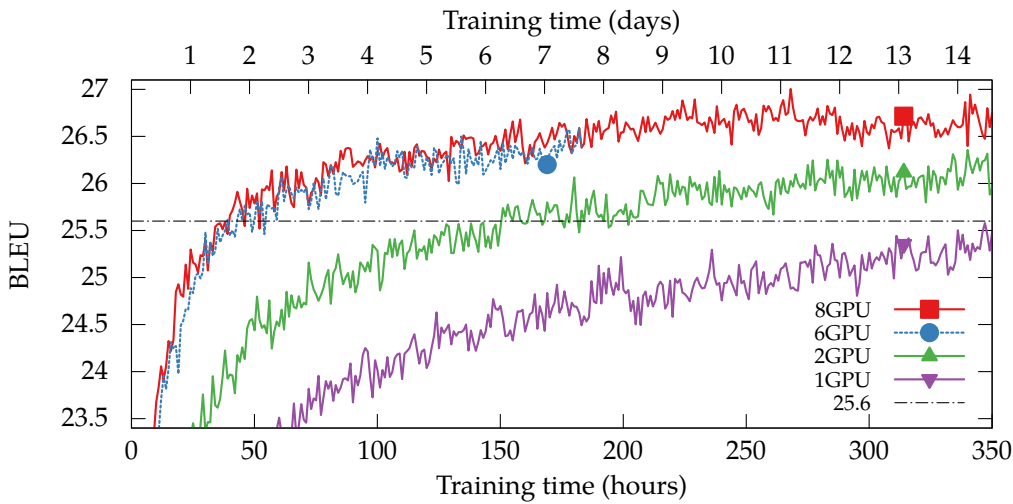
\begin{figure}
\input{plots/1268GPU-czeng57-b15.tex}\vspace{-5mm}
\caption{Effect of the number of GPUs. BLEU=25.6 is marked with a black line.}
\label{fig:gpus-czeng57}
\end{figure}

\bigskip
Figure~\ref{fig:gpus-czeng57} shows BLEU curves for different numbers of GPUs and the BIG model
 with batch size, learning rate and warmup steps fixed on their default values (1500, 0.20 and 16k, respectively).
As could be expected, training with more GPUs converges faster.
What is interesting is the Time Till Score. 
\Tref{tab:gpus-tts} lists
the approximate training time
and number of training examples (in millions of subwords)
needed to ``surpass'' (i.e. achieve and never again fall below) BLEU of 25.6.

\begin{table}[ht]
\begin{center}
\begin{tabular}{rrr}\hline
\# GPUs & hours & subwords (M)\\\hline
1       & $>$ 600 &        $>$ 9000 \\
2       & 203 & 2322$\cdot$2 = 4644 \\
6       &  56 &  451$\cdot$6 = 2706 \\
8       &  40 &  341$\cdot$8 = 2728 \\\hline
\end{tabular}
\end{center}
\caption{Time and training data consumed to reach BLEU of 25.6,\
 i.e. Time Till Score and Examples Till Score.
Note that the experiment on 1 GPU was ended after 25 days of training
 without clearly surpassing the threshold
 (already outside of Figure~\ref{fig:gpus-czeng57}).
}
\label{tab:gpus-tts}
\end{table}

We can see that \recommend{two GPUs are more than three times faster than a single GPU}
 when measuring the Time Till Score
 and need much less training examples (i.e. they have lower Examples Till Score).
Similarly, \recommend{eight GPUs are more than five times faster than two GPUs}
 and 1.7 times less training data is needed.

Recall that in \Fref{fig:1GPU-czeng57-batch-size} we have shown
 that increasing the batch size from 1450 to 2000 has almost no effect on the BLEU curve.
However, when increasing the effective batch size by using more GPUs,
 the improvement is higher than could be expected from the higher throughput.\footnote{
  It would be interesting to try simulating multi-GPU training
   on a single GPU, simply by doing the update once after N batches
   (and summing the gradients).
  This is similar to the \emph{ghost batches} of \citet{hoffer-et-al-2017},
   but using ghost batch size higher than the actual batch size.
   We leave this for future work.
 }
We find this quite surprising,
 especially considering the fact
 that we have not tuned the learning rate and warmup steps (see the next section).

\subsubsection*{Tips on the Number of GPUs}
\begin{itemize}
\item For the fastest BLEU convergence \recommend{use as many GPUs as available} (in our experiments up to 8).
\item This holds \recommend{even when there are more experiments} to be done.
      For example, it is better to run one 8-GPUs experiment after another,
      rather than running two 4-GPUs experiments in parallel
      or eight single-GPU experiments in parallel.
\end{itemize}

\subsection{Learning Rate and Warmup Steps on Multiple GPUs}
\label{sec:scaling}

\subsubsection{Related Work}
There is a growing number of papers on scaling deep learning
 to multiple machines with synchronous SGD (or its variants)
 by increasing the effective batch size.
We will focus mostly on the question how to adapt the learning rate schedule,
 when scaling from one GPU (or any device, in general) to $k$ GPUs.

\citet{krizhevsky:2014} says
 ``\emph{Theory suggests that when multiplying the batch size by $k$, one should multiply the
         learning rate  by $\sqrt{k}$ to keep the variance in the gradient expectation constant}'',
 without actually explaining which theory suggests so.
However, in the experimental part he reports that what worked the best,
 was a \emph{linear scaling heuristics}, i.e. multiplying the learning rate by $k$,
 again without any explanation nor details on the difference between $\sqrt{k}$ scaling and $k$ scaling.

The linear scaling heuristics become popular,
 leading to good scaling results in practice \citep{goyal-et-al:2017,smith:etal:lr:batchsize:arxiv:2017}
 and also theoretical explanations \citep{bottou-et-al:2016,smith:le:generalization:2017,jastrzebski-et-al:2017}.
\citet{smith:le:generalization:2017} interpret SGD (and its variants) as a stochastic differential equation
 and show that the \emph{gradient noise scale} $g = \epsilon \left(\frac{N}{B}-1\right)$,
 where $\epsilon$ is the learning rate,
 $N$ is the training set size,
 and $B$ is the effective batch size.
This noise ``\emph{drives SGD away from sharp minima, and therefore there is an optimal batch size
 which maximizes the test set accuracy}''.
In other words for keeping the optimal level of gradient noise
 (which leads to ``flat minima'' that generalize well),
 we need to scale the learning rate linearly when increasing the effective batch size.

However, \citet{hoffer-et-al-2017} suggest to use $\sqrt{k}$ scaling instead of the linear scaling
 and provide both theoretical and empirical support for this claim.
They show that $\textrm{cov}(\Delta w, \Delta w) \propto \frac{\epsilon^2}{NB}$,
 thus if we want to keep the the covariance matrix of the parameters update step $\Delta w$
 in the same range for any effective batch size $B$,
 we need to scale the learning rate proportionally to the square root of $B$.
They found that $\sqrt{k}$ scaling works better than linear scaling on CIFAR10.\footnote{
 To close the gap between small-batch training and large-batch training,
  \citet{hoffer-et-al-2017} introduce (in addition to $\sqrt{k}$ scaling) so-called \emph{ghost batch normalization}
  and \emph{adapted training regime},
  which means decaying the learning rate after a given number of steps instead of epochs.
}
\citet{you-et-al:2017} confirm linear scaling does not perform well on ImageNet
 and suggest to use Layer-wise Adaptive Rate Scaling.

We can see that large-batch training is still an open research question. 
Most of the papers cited above have experimental support only from the image recognition tasks (usually ImageNet)
 and convolutional networks (e.g. ResNet),
 so it is not clear whether their suggestions can be applied also on sequence-to-sequence tasks (NMT)
 with self-attentional networks (Transformer).
There are several other differences as well:
Modern convolutional networks are usually trained with \emph{batch normalization} \citep{batch-norm},
 which seems to be important for the scaling,
 while Transformer uses \emph{layer normalization} \citep{layer-norm}.\footnote{
  Applying batch normalization on RNN is difficult.
  Transformer does not use RNN, but still we were not successful
   in switching to batch normalization (and possibly ghost batch normalization)
   due to NaN loss errors.
 }
Also, Transformer uses Adam together with an inverse-square-root learning-rate decay,
 while most ImageNet papers use SGD with momentum and piecewise-constant learning-rate decay.

\subsubsection{Our Experiments}
We decided to find out empirically the optimal learning rate for training on 8 GPUs.
Increasing the learning rate from 0.20 to 0.30
 resulted in diverged training (BLEU dropped to almost 0 after two hours of training).
Similarly to our single-GPU experiments (\Sref{sec:learning-rate}),
 we were able prevent the divergence by increasing the warmup steps
 or by introducing gradient clipping
 (e.g. with \verb|clip_grad_norm=1.0|, we were able to use learning rate 0.40,
 but increasing it further to 0.60 led to divergence anyway).
However, \recommend{none of these experiments led to any improvements over the default learning rate} --
 all had about the same BLEU curve after few hours of training.

\citet{jastrzebski-et-al:2017} shows that
 ``\emph{the invariance under simultaneous rescaling of learning rate and batch size
  breaks down if the learning rate gets too large or the batch size gets too small}''.
A similar observation was reported e.g. by \citet{bottou-et-al:2016}.
Thus our initial hypothesis was that 0.20 (or 0.25) is the maximal learning rate
 suitable for stable training in our experiments
 even when we scale from a single GPU to 8 GPUs.
Considering this initial hypothesis,
 we were surprised that we were able to achieve so good Time Till Score with 8 GPUs
 (more than 8 times smaller relative to a single GPU, as reported in \Tref{tab:gpus-tts}).
To answer this riddle we need to understand how learning rate schedules are implemented in T2T.

\subsubsection{Parametrization of Learning Rate Schedules in T2T}\label{sec:lr-schedule}
In most works on learning rate schedules\cprotect\footnote{~Examples of learning rate schedules are
 inverse-square-root decay, inverse-time decay, exponential decay, piecewise-constant decay,
 see \url{https://www.tensorflow.org/api_guides/python/train#Decaying_the_learning_rate}
 for TF implementations.
}
 the ``time'' parameter is actually interpreted as the number of epochs or training examples.
For example a popular setup for piecewise-constant decay in ImageNet training \citep[e.g.][]{goyal-et-al:2017}
 is to divide the learning rate by a factor of 10 at the 30-th, 60-th, and 80-th epoch.

However, in T2T, it is the \verb|global_step| variable that is used as the ``time'' parameter.
So when increasing the effective batch size 8 times, e.g. by using 8 GPUs instead of a single GPU,
 the actual learning rate\footnote{
  By \emph{actual} learning rate
   we mean the learning rate after applying the decay schedule.
   The \texttt{learning\_rate} parameter stays the same in this case.
 } achieves a given value after the same number of steps,
 but this means after 8 times less training examples.
For the inverse-square-root decay, we have
 $\textit{actual\_lr}(\textit{steps}) =
  c \cdot \textit{steps}^{-0.5} =
  \frac{1}{\sqrt{8}} \cdot \textit{actual\_lr}(\textit{steps} \cdot 8)$,
 where $c$ is a constant containing also the \verb|learning_rate| parameter.
So with 8 GPUs, if we divide the \verb|learning_rate| parameter by $\sqrt{8}$,
 we achieve the same actual learning rate after a given number of training examples
 as in the original single-GPU setting.

This explains the riddle from the previous section.
\recommend{By keeping the \texttt{learning\_rate} parameter the same
 when scaling to $k$ times bigger effective batch,
 we actually increase the actual learning rate $\sqrt{k}$ times},
 in accordance with the suggestion of \citet{hoffer-et-al-2017}.\footnote{
  In addition to suggesting the $\sqrt{k}$ learning-rate scaling,
   \citet{hoffer-et-al-2017} show that to fully close the ``generalization gap'',
   we need to train longer because the absolute number of steps (updates) matters.
  So from this point of view, using steps instead of epochs as the time parameter
   for learning rate schedules may not be a completely wrong idea.
 }
This holds only for the \texttt{linear\_warmup\_rsqrt\_decay} (aka \texttt{noam}) schedule
 and ignoring the warmup steps.

If we want to keep the same learning rate also in the warmup phase,
 we would need to divide the warmup steps by $k$.
However, this means that the maximum actual learning rate will be $\sqrt{k}$ times higher,
 relative to the single-GPU maximal actual learning rate
 and this leads to divergence in our experiments.
In deed, many researchers \citep[e.g.][]{goyal-et-al:2017} suggest to use a warmup
 when scaling to more GPUs in order to prevent divergence.
Transformer uses learning rate warmup by default even for single-GPU training
 (cf. \Sref{sec:learning-rate}),
 but it makes sense to use more warmup training examples in multi-GPU setting.

In our experiments with 8 GPUs and the default learning rate 0.20,
 using 8k warmup steps instead of the default 16k had no effect on the BLEU curve
 (it was a bit higher in the first few hours, but the same afterwards).
Further decreasing the warmup steps resulted in a retarded BLEU curve (for 6k)
 or a complete divergence (for 2k).

\subsection*{Tips on Learning Rate and Warmup Steps on Multiple GPUs}
\begin{itemize}
\item Keep the \texttt{learning\_rate} parameter at its optimal value found in single-GPU experiments.
\item You can try decreasing the warmup steps, but less than linearly
 and you should not expect to improve the final BLEU this way.
\end{itemize}

\subsection{Resumed Training}\label{sec:resumed-training}
T2T allows to resume training from a checkpoint,
 simply by pointing the \verb|output_dir| parameter to a directory
 with an existing checkpoint (specified in the \verb|checkpoint| file).
This may be useful when the training fails (e.g. because of hardware error),
 when we need to continue training on a different machine
 or during hyper-parameter search,
 when we want to continue with the most promising setups.
T2T saves also Adam momentum into the checkpoint,
 so the training continues almost as if it had not been stopped.
However, it does not store the position in the training data
 -- it starts from a random position.
Also the relative time (and wall-clock time) in TensorBoard graphs
 will be influenced by the stopping.

Resumed training can also be exploited for changing some hyper-parameters,
 which cannot be meta-parametrized by the number of steps.
For example, \citet{smith:etal:lr:batchsize:arxiv:2017} suggest
 to increase the effective batch size (and number of GPUs) during training,
 instead of decaying the learning rate.

Yet another usage is to do domain adaptation
 by switching from (large) general-domain training data
 to (small) target-domain training data
 for the few last epochs.
In this case, consider editing also the learning rate or learning rate schedule
 (or faking the \verb|global_step| stored in the checkpoint)
 to make sure the learning rate is not too small.

\subsection{Checkpoint Averaging}\label{sec:averaging}
\citet{vaswani-et-al:2017} suggest to average the last 20 checkpoints saved in 10-minute intervals
 (using \verb|utils/avg_checkpoints.py|).
According to our experiments slightly better results are achieved
 with averaging checkpoints saved in 1-hour intervals.
This has also the advantage that less time is spent with checkpoint saving, so the training is faster.

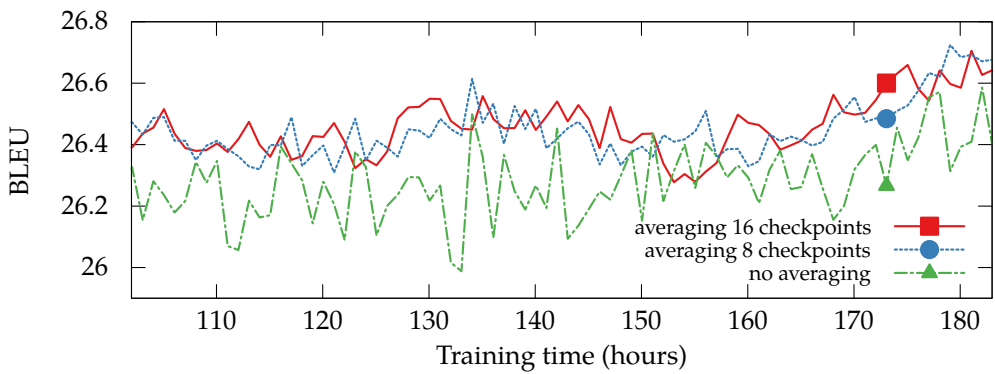
\begin{figure}
\input{plots/6GPU-czeng57-averaging.tex}\vspace{-5mm}
\caption{Effect of checkpoint averaging. All trained on 6 GPUs.}
\label{fig:averaging}
\end{figure}

Figure~\ref{fig:averaging} shows the effect of averaging is twofold:
 the averaged curve has lower variance (flickering) from checkpoint to checkpoint
 and it is almost always better than the baseline without averaging
 (usually by about 0.2 BLEU).
In some setups, we have seen improvements due to averaging over 1 BLEU.
In the early phases of training, while the (baseline) learning curve grows fast,
 it is better to use fewer checkpoints for averaging.
In later phases (as shown in Figure~\ref{fig:averaging}, after 4.5--7.5 days of training),
 it seems that 16 checkpoints (covering last 16 hours) give
 slightly better results on average than 8 checkpoints,
 but we have not done any proper evaluation for significance
 (using paired bootstrap testing for each hour and then summarizing the results).

The fact that resumed training starts from a random position in the training data
 (cf. \Sref{sec:resumed-training})
 can be actually exploited for ``forking'' a training
 to get two (or more) copies of the model,
 which are trained for the same number of steps,
 but independently in the later stages
 and thus ending with different weights saved in the final checkpoint.
These semi-independent models can be averaged in the same way
 as checkpoints from the same run, as described above.
Our preliminary results show this helps a bit (on top of checkpoint averaging).

\subsubsection*{Tips on Checkpoint Averaging}
\begin{itemize}
\item Use it.
 Averaging 8 checkpoints takes about 5 minutes, so it is a ``BLEU boost for free''
  (compared with the time needed for the whole training).
\item See the tools for automatic checkpoint averaging and evaluation described in \Sref{sec:tools}.
\end{itemize}

\section{Comparison with WMT17 Systems}
\label{wmt-comparison}

\begin{table}[t]
\def\fst#1{\bf #1}
\def\snd#1{\it #1}
\begin{center}
\footnotesize
\begin{tabular}{lcc|ccccc}
\multicolumn{3}{c|}{Manual} & \multicolumn{4}{c}{Automatic Scores} \\
\#    	& Ave \%    	& Ave z      	& BLEU      	& TER        	& CharacTER  	& BEER       	& System \\
\hline
     -- &         --    &         --    & \fst{23.8}    & \fst{0.662}   & \fst{0.582}   & \fst{0.543}   & T2T 8 GPUs 8 days\\ 
\hline
     1	& \fst{62.0}	& \fst{0.308}	& \snd{22.8}	& \snd{0.667}	& \snd{0.588}	& \snd{0.540}	& uedin-nmt \\
\hline
     2	& \snd{59.7}	& \snd{0.240}	&       20.1	&       0.703	&       0.612	&       0.519	& online-B \\
\hline
     3	&       55.9	&       0.111	&       20.2	&     {0.696}	&       0.607	&     {0.524}	& limsi-factored \\
      	&       55.2	&       0.102	&       20.0	&       0.699	& -          	& -          	& LIUM-FNMT \\
      	&       55.2	&       0.090	&       20.2	&       0.701	& \snd{0.605}	&       0.522	& LIUM-NMT \\
      	&       54.1	&       0.050	&     {20.5}	&     {0.696}	&       0.624	&       0.523	& CU-Chimera \\
      	&       53.3	&       0.029	&       16.6	&       0.743	&       0.637	&       0.503	& online-A \\
\hline
     8	&       41.9	&      -0.327	&       16.2	&       0.757	&       0.697	&       0.485	& PJATK \\
\end{tabular}
\caption{WMT17 systems for English-to-Czech and our best T2T training run. Manual scores are from the official WMT17 ranking.
Automatic metrics were provided by \protect\url{http://matrix.statmt.org/}. For *TER
metrics, lower is better. Best results in bold, second-best in italics.}
\label{tab:wmt17results}
\end{center}
\end{table}

\Tref{tab:wmt17results} provides the results of WMT17 English-to-Czech news translation task,
 with our best Transformer model (BIG trained on 8 GPUs for 8 days, averaging 8 checkpoints)
 evaluated using the exact same implementation of automatic metrics.
While the automatic evaluation is not fully reliable
 (see e.g. the high BLEU score for CU-Chimera despite its lower manual rank),
 we see that the Transformer model outperforms the best system in BLEU, TER, CharacTER and BEER,
 despite it does not use any back-translated data,
 reranking with other models (e.g. right-to-left reranking) nor ensembling (as is the case of uedin-nmt and other systems).
Note that our Transformer uses a subset of the constrained training data for WMT17,
 so the results are comparable.

\section{Conclusion}
\label{conclusion}

We presented a broad range of basic experiments with the Transformer model
\parcite{vaswani-et-al:2017} for English-to-Czech neural machine translation.
While we limit our exploration to the more or less basic parameter settings, we
believe this report can be useful for other researchers.
In sum, experiments done for this article took about 4 years of GPU time.

Among other practical observations, we've seen that for the Transformer model, larger batch sizes lead not only to
faster training but more importantly better translation quality.
Given at least a day and a 11GB GPU for training, the larger setup (BIG)
should be always preferred.
The Transformer model and its implementation in Tensor2Tensor is also best fit
for ``intense training'': using as many GPUs as possible and running experiments
one after another should be preferred over running several single-GPU experiments concurrently.

The best performing model we obtained on 8 GPUs trained for 8 days has outperformed the WMT17 winner in a number of automatic metrics.

\section*{Acknowledgements}
This research was supported by the grants
18-24210S of the Czech Science Foundation, 
H2020-ICT-2014-1-645452 (QT21) of the EU,
SVV~260~453, 
 and using language resources distributed by the LINDAT/CLARIN project of the
 Ministry of Education, Youth and Sports of the Czech Republic (LM2015071).

\bibliography{biblio}

\correspondingaddress
\end{document}

%% file: plots/8GPU-czeng1-vs-czeng57.tex
\begin{tikzpicture}[gnuplot]
\path (0.000,0.000) rectangle (13.600,4.500);
\gpcolor{color=gp lt color border}
\gpsetlinetype{gp lt border}
\gpsetlinewidth{1.00}
\draw[gp path] (1.320,1.345)--(1.500,1.345);
\draw[gp path] (13.047,1.345)--(12.867,1.345);
\node[gp node right] at (1.136,1.345) { 24};
\draw[gp path] (1.320,2.064)--(1.500,2.064);
\draw[gp path] (13.047,2.064)--(12.867,2.064);
\node[gp node right] at (1.136,2.064) { 25};
\draw[gp path] (1.320,2.784)--(1.500,2.784);
\draw[gp path] (13.047,2.784)--(12.867,2.784);
\node[gp node right] at (1.136,2.784) { 26};
\draw[gp path] (1.320,3.503)--(1.500,3.503);
\draw[gp path] (13.047,3.503)--(12.867,3.503);
\node[gp node right] at (1.136,3.503) { 27};
\draw[gp path] (1.320,0.985)--(1.320,1.165);
\node[gp node center] at (1.320,0.677) { 0};
\draw[gp path] (3.492,0.985)--(3.492,1.165);
\node[gp node center] at (3.492,0.677) { 50};
\draw[gp path] (5.663,0.985)--(5.663,1.165);
\node[gp node center] at (5.663,0.677) { 100};
\draw[gp path] (7.835,0.985)--(7.835,1.165);
\node[gp node center] at (7.835,0.677) { 150};
\draw[gp path] (10.007,0.985)--(10.007,1.165);
\node[gp node center] at (10.007,0.677) { 200};
\draw[gp path] (12.178,0.985)--(12.178,1.165);
\node[gp node center] at (12.178,0.677) { 250};
\draw[gp path] (2.362,3.575)--(2.362,3.395);
\node[gp node center] at (2.362,3.883) { 1};
\draw[gp path] (3.405,3.575)--(3.405,3.395);
\node[gp node center] at (3.405,3.883) { 2};
\draw[gp path] (4.447,3.575)--(4.447,3.395);
\node[gp node center] at (4.447,3.883) { 3};
\draw[gp path] (5.490,3.575)--(5.490,3.395);
\node[gp node center] at (5.490,3.883) { 4};
\draw[gp path] (6.532,3.575)--(6.532,3.395);
\node[gp node center] at (6.532,3.883) { 5};
\draw[gp path] (7.574,3.575)--(7.574,3.395);
\node[gp node center] at (7.574,3.883) { 6};
\draw[gp path] (8.617,3.575)--(8.617,3.395);
\node[gp node center] at (8.617,3.883) { 7};
\draw[gp path] (9.659,3.575)--(9.659,3.395);
\node[gp node center] at (9.659,3.883) { 8};
\draw[gp path] (10.702,3.575)--(10.702,3.395);
\node[gp node center] at (10.702,3.883) { 9};
\draw[gp path] (11.744,3.575)--(11.744,3.395);
\node[gp node center] at (11.744,3.883) { 10};
\draw[gp path] (12.786,3.575)--(12.786,3.395);
\node[gp node center] at (12.786,3.883) { 11};
\draw[gp path] (1.320,3.575)--(1.320,0.985)--(13.047,0.985)--(13.047,3.575)--cycle;
\node[gp node center,rotate=-270] at (0.246,2.280) {BLEU};
\node[gp node center] at (7.183,0.215) {Training time (hours)};
\node[gp node center] at (7.183,4.344) {Training time (days)};
\node[gp node right,font={\fontsize{8pt}{9.6pt}\selectfont}] at (11.579,1.627) {58M training sentences};
\gpcolor{rgb color={0.894,0.102,0.110}}
\gpsetlinetype{gp lt plot 0}
\gpsetlinewidth{2.00}
\draw[gp path] (11.763,1.627)--(12.679,1.627);
\draw[gp path] (1.736,0.985)--(1.757,1.109)--(1.800,1.229)--(1.844,1.385)--(1.887,1.284)%
  --(1.931,1.383)--(1.974,1.578)--(2.017,1.503)--(2.061,1.588)--(2.104,1.955)--(2.148,2.012)%
  --(2.191,1.934)--(2.235,1.917)--(2.278,2.163)--(2.322,1.977)--(2.365,2.279)--(2.408,2.178)%
  --(2.452,2.165)--(2.495,2.100)--(2.539,2.235)--(2.582,2.228)--(2.626,2.347)--(2.669,2.118)%
  --(2.712,2.331)--(2.756,2.265)--(2.799,2.452)--(2.843,2.472)--(2.886,2.479)--(2.930,2.433)%
  --(2.973,2.489)--(3.016,2.397)--(3.060,2.540)--(3.103,2.579)--(3.147,2.635)--(3.190,2.720)%
  --(3.234,2.647)--(3.277,2.580)--(3.320,2.458)--(3.364,2.743)--(3.407,2.596)--(3.451,2.691)%
  --(3.494,2.631)--(3.538,2.708)--(3.581,2.438)--(3.625,2.738)--(3.668,2.731)--(3.711,2.764)%
  --(3.755,2.732)--(3.798,2.884)--(3.842,2.569)--(3.885,2.683)--(3.929,2.713)--(3.972,2.871)%
  --(4.015,2.730)--(4.059,2.914)--(4.102,2.725)--(4.146,2.633)--(4.189,2.602)--(4.233,2.889)%
  --(4.276,2.747)--(4.319,2.870)--(4.363,2.773)--(4.406,2.753)--(4.450,2.614)--(4.493,2.712)%
  --(4.537,2.734)--(4.580,2.825)--(4.623,2.793)--(4.667,2.693)--(4.710,2.946)--(4.754,2.929)%
  --(4.797,2.840)--(4.841,2.924)--(4.884,2.857)--(4.928,2.946)--(4.971,2.874)--(5.014,3.033)%
  --(5.058,2.970)--(5.101,3.049)--(5.145,3.055)--(5.188,3.049)--(5.232,2.910)--(5.275,2.930)%
  --(5.318,2.952)--(5.362,2.944)--(5.405,2.918)--(5.449,2.960)--(5.492,2.993)--(5.536,2.962)%
  --(5.579,2.989)--(5.622,2.906)--(5.666,3.023)--(5.709,3.087)--(5.753,2.978)--(5.796,3.031)%
  --(5.840,3.054)--(5.883,2.946)--(5.926,2.944)--(5.970,2.931)--(6.013,2.980)--(6.057,2.874)%
  --(6.100,2.890)--(6.144,3.026)--(6.187,2.850)--(6.231,3.025)--(6.274,2.917)--(6.317,3.056)%
  --(6.361,3.083)--(6.404,2.995)--(6.448,3.032)--(6.491,3.028)--(6.535,2.988)--(6.578,2.976)%
  --(6.621,2.898)--(6.665,2.974)--(6.708,2.945)--(6.752,3.043)--(6.795,2.976)--(6.839,3.011)%
  --(6.882,3.027)--(6.925,2.801)--(6.969,3.141)--(7.012,3.069)--(7.056,3.109)--(7.099,3.095)%
  --(7.143,3.049)--(7.186,3.151)--(7.229,3.123)--(7.273,3.132)--(7.316,3.136)--(7.360,3.128)%
  --(7.403,2.998)--(7.447,3.017)--(7.490,3.062)--(7.534,3.072)--(7.577,3.095)--(7.620,3.006)%
  --(7.664,3.080)--(7.707,3.093)--(7.751,3.148)--(7.794,3.067)--(7.838,3.033)--(7.881,3.101)%
  --(7.924,3.196)--(7.968,3.044)--(8.011,3.162)--(8.055,3.258)--(8.098,3.100)--(8.142,3.169)%
  --(8.185,3.272)--(8.228,3.065)--(8.272,2.937)--(8.315,2.971)--(8.359,3.079)--(8.402,2.993)%
  --(8.446,3.101)--(8.489,3.126)--(8.532,3.169)--(8.576,3.106)--(8.619,3.058)--(8.663,3.092)%
  --(8.706,3.143)--(8.750,3.209)--(8.793,3.120)--(8.837,3.171)--(8.880,3.247)--(8.923,3.217)%
  --(8.967,3.239)--(9.010,3.177)--(9.054,3.065)--(9.097,3.018)--(9.141,3.147)--(9.184,3.099)%
  --(9.227,3.208)--(9.271,3.057)--(9.314,3.086)--(9.358,3.276)--(9.401,3.159)--(9.445,3.062)%
  --(9.488,3.239)--(9.531,3.179)--(9.575,3.262)--(9.618,3.122)--(9.662,3.180)--(9.705,3.197)%
  --(9.749,3.207)--(9.792,3.304)--(9.835,3.244)--(9.879,3.332)--(9.922,3.103)--(9.966,3.173)%
  --(10.009,3.174)--(10.053,3.251)--(10.096,3.225)--(10.140,3.196)--(10.183,3.203)--(10.226,3.152)%
  --(10.270,3.107)--(10.313,3.196)--(10.357,3.290)--(10.400,3.176)--(10.444,3.293)--(10.487,3.182)%
  --(10.530,3.211)--(10.574,3.249)--(10.617,3.205)--(10.661,3.163)--(10.704,3.199)--(10.748,3.297)%
  --(10.791,3.270)--(10.834,3.283)--(10.878,3.337)--(10.921,3.270)--(10.965,3.345)--(11.008,3.280)%
  --(11.052,3.418)--(11.095,3.346)--(11.138,3.268)--(11.182,3.293)--(11.225,3.281)--(11.269,3.426)%
  --(11.312,3.253)--(11.356,3.243)--(11.399,3.230)--(11.443,3.250)--(11.486,3.180)--(11.529,3.289)%
  --(11.573,3.253)--(11.616,3.169)--(11.660,3.374)--(11.703,3.233)--(11.747,3.261)--(11.790,3.278)%
  --(11.833,3.212)--(11.877,3.257)--(11.920,3.081)--(11.964,3.334)--(12.007,3.385)--(12.051,3.270)%
  --(12.094,3.340)--(12.137,3.355)--(12.181,3.309)--(12.224,3.251)--(12.268,3.267)--(12.311,3.187)%
  --(12.355,3.314)--(12.398,3.155)--(12.441,3.204)--(12.485,3.379)--(12.528,3.334)--(12.572,3.376)%
  --(12.615,3.339)--(12.659,3.265)--(12.702,3.286)--(12.746,3.261)--(12.789,3.371)--(12.832,3.421)%
  --(12.876,3.308)--(12.919,3.324)--(12.963,3.506)--(13.006,3.364)--(13.047,3.277);
\gpsetpointsize{8.00}
\gppoint{gp mark 5}{(12.137,3.355)}
\gppoint{gp mark 5}{(12.221,1.627)}
\gpcolor{color=gp lt color border}
\node[gp node right,font={\fontsize{8pt}{9.6pt}\selectfont}] at (11.579,1.319) {16M training sentences};
\gpcolor{rgb color={0.216,0.494,0.722}}
\gpsetlinetype{gp lt plot 3}
\draw[gp path] (11.763,1.319)--(12.679,1.319);
\draw[gp path] (1.747,0.985)--(1.770,1.044)--(1.770,0.985);
\draw[gp path] (1.783,0.985)--(1.830,1.338)--(1.834,1.355)--(1.848,1.423)--(1.876,1.851)%
  --(1.912,1.734)--(1.955,1.718)--(1.999,1.738)--(2.042,1.761)--(2.088,1.816)--(2.129,1.916)%
  --(2.172,2.124)--(2.215,1.972)--(2.259,1.962)--(2.303,1.864)--(2.350,1.944)--(2.389,2.196)%
  --(2.434,2.248)--(2.476,2.235)--(2.520,2.174)--(2.563,2.002)--(2.607,2.041)--(2.650,2.196)%
  --(2.693,2.445)--(2.737,2.178)--(2.780,2.342)--(2.824,2.053)--(2.868,2.214)--(2.911,2.383)%
  --(3.044,2.331)--(3.049,2.370)--(3.085,2.290)--(3.090,2.142)--(3.129,2.422)--(3.180,2.430)%
  --(3.237,2.510)--(3.295,2.390)--(3.353,2.445)--(3.381,2.403)--(3.436,2.287)--(3.483,2.575)%
  --(3.511,2.464)--(3.543,2.492)--(3.565,2.488)--(3.604,2.408)--(3.649,2.383)--(3.692,2.505)%
  --(3.737,2.546)--(3.782,2.499)--(3.818,2.364)--(3.911,2.384)--(3.969,2.401)--(3.984,2.362)%
  --(4.011,2.488)--(4.042,2.510)--(4.083,2.377)--(4.131,2.524)--(4.191,2.353)--(4.220,2.339)%
  --(4.266,2.445)--(4.317,2.570)--(4.345,2.388)--(4.391,2.444)--(4.431,2.486)--(4.474,2.478)%
  --(4.517,2.440)--(4.561,2.578)--(4.604,2.362)--(4.648,2.316)--(4.698,2.390)--(4.740,2.484)%
  --(4.798,2.479)--(4.828,2.537)--(4.864,2.448)--(4.978,2.378)--(5.005,2.349)--(5.015,2.520)%
  --(5.038,2.448)--(5.084,2.617)--(5.131,2.376)--(5.173,2.407)--(5.212,2.478)--(5.256,2.489)%
  --(5.301,2.622)--(5.354,2.474)--(5.403,2.299)--(5.459,2.365)--(5.480,2.533)--(5.513,2.600)%
  --(5.560,2.435)--(5.603,2.466)--(5.646,2.561)--(5.690,2.403)--(5.733,2.446)--(5.777,2.338)%
  --(5.831,2.355)--(5.869,2.447)--(5.910,2.397)--(5.951,2.327)--(5.994,2.180)--(6.058,2.463)%
  --(6.137,2.318)--(6.134,2.327)--(6.173,2.277)--(6.212,2.387)--(6.518,2.284)--(6.749,2.359)%
  --(6.773,2.440)--(6.783,2.433)--(6.800,2.499)--(6.821,2.542)--(6.833,2.416)--(6.850,2.306)%
  --(6.866,2.305)--(6.884,2.426)--(6.899,2.470)--(6.912,2.366)--(6.933,2.321)--(6.943,2.308)%
  --(6.956,2.509)--(6.969,2.524)--(6.980,2.307)--(7.000,2.470)--(7.094,2.254)--(7.093,2.330)%
  --(7.123,2.366)--(7.167,2.463)--(7.211,2.419)--(7.255,2.262)--(7.297,2.413)--(7.341,2.236)%
  --(7.384,2.435)--(7.427,2.463)--(7.471,2.588)--(7.775,2.348)--(7.793,2.249)--(7.823,2.358)%
  --(7.836,2.342)--(7.852,2.511)--(7.881,2.395)--(7.893,2.320)--(7.897,2.431)--(7.932,2.360)%
  --(7.949,2.341)--(7.961,2.584)--(7.994,2.399)--(8.036,2.545)--(8.151,2.406)--(8.150,2.491)%
  --(8.166,2.458)--(8.205,2.576)--(8.254,2.550)--(8.296,2.257)--(8.341,2.316)--(8.383,2.470)%
  --(8.427,2.225)--(8.473,2.490)--(8.513,2.522)--(8.557,2.384)--(8.602,2.400)--(8.645,2.342)%
  --(8.689,2.482)--(8.734,2.387)--(8.776,2.353)--(8.817,2.321)--(8.863,2.440)--(8.908,2.311)%
  --(8.948,2.461)--(8.991,2.318)--(9.037,2.282)--(9.080,2.417)--(9.121,2.291)--(9.166,2.285)%
  --(9.208,2.271)--(9.252,2.495)--(9.296,2.441)--(9.339,2.453)--(9.385,2.338)--(9.428,2.400)%
  --(9.471,2.479)--(9.515,2.565)--(9.556,2.546)--(9.600,2.412)--(9.642,2.418)--(9.686,2.448)%
  --(9.732,2.414)--(9.773,2.432)--(9.816,2.508)--(9.860,2.408)--(9.906,2.376)--(9.949,2.481)%
  --(9.990,2.400)--(10.036,2.368)--(10.081,2.409)--(10.121,2.520)--(10.164,2.413)--(10.207,2.246)%
  --(10.251,2.393)--(10.295,2.444)--(10.333,2.354)--(10.381,2.425)--(10.425,2.330)--(10.470,2.450)%
  --(10.512,2.420)--(10.551,2.371)--(10.598,2.559)--(10.643,2.566)--(10.685,2.368)--(10.729,2.308)%
  --(10.773,2.503)--(10.816,2.325)--(10.859,2.367)--(10.902,2.582)--(10.946,2.335)--(10.932,2.436)%
  --(10.976,2.386)--(11.019,2.339)--(11.063,2.371)--(11.106,2.378)--(11.150,2.369)--(11.193,2.446)%
  --(11.237,2.391)--(11.280,2.426)--(11.323,2.319)--(11.367,2.352)--(11.410,2.367)--(11.454,2.381)%
  --(11.497,2.401)--(11.541,2.348)--(11.584,2.258)--(11.628,2.396)--(11.671,2.506)--(11.714,2.411)%
  --(11.758,2.501)--(11.801,2.256)--(11.845,2.333)--(11.888,2.437)--(11.932,2.479)--(11.975,2.467)%
  --(12.018,2.441)--(12.062,2.292)--(12.105,2.479)--(12.149,2.440)--(12.192,2.350)--(12.236,2.443)%
  --(12.279,2.467)--(12.322,2.500)--(12.366,2.358)--(12.409,2.536);
\gppoint{gp mark 7}{(12.018,2.441)}
\gppoint{gp mark 7}{(12.221,1.319)}
\gpcolor{color=gp lt color border}
\gpsetlinetype{gp lt border}
\gpsetlinewidth{1.00}
\draw[gp path] (1.320,3.575)--(1.320,0.985)--(13.047,0.985)--(13.047,3.575)--cycle;
\gpdefrectangularnode{gp plot 1}{\pgfpoint{1.320cm}{0.985cm}}{\pgfpoint{13.047cm}{3.575cm}}
\end{tikzpicture}

%% file: plots/1GPU-big2000-base4500-base2000.tex
\begin{tikzpicture}[gnuplot]
\path (0.000,0.000) rectangle (13.600,4.500);
\gpcolor{color=gp lt color border}
\gpsetlinetype{gp lt border}
\gpsetlinewidth{1.00}
\draw[gp path] (1.320,1.557)--(1.500,1.557);
\draw[gp path] (13.047,1.557)--(12.867,1.557);
\node[gp node right] at (1.136,1.557) { 16};
\draw[gp path] (1.320,2.129)--(1.500,2.129);
\draw[gp path] (13.047,2.129)--(12.867,2.129);
\node[gp node right] at (1.136,2.129) { 18};
\draw[gp path] (1.320,2.701)--(1.500,2.701);
\draw[gp path] (13.047,2.701)--(12.867,2.701);
\node[gp node right] at (1.136,2.701) { 20};
\draw[gp path] (1.320,3.273)--(1.500,3.273);
\draw[gp path] (13.047,3.273)--(12.867,3.273);
\node[gp node right] at (1.136,3.273) { 22};
\draw[gp path] (1.320,3.845)--(1.500,3.845);
\draw[gp path] (13.047,3.845)--(12.867,3.845);
\node[gp node right] at (1.136,3.845) { 24};
\draw[gp path] (1.320,0.985)--(1.320,1.165);
\draw[gp path] (1.320,4.131)--(1.320,3.951);
\node[gp node center] at (1.320,0.677) { 0};
\draw[gp path] (2.926,0.985)--(2.926,1.165);
\draw[gp path] (2.926,4.131)--(2.926,3.951);
\node[gp node center] at (2.926,0.677) { 10};
\draw[gp path] (4.533,0.985)--(4.533,1.165);
\draw[gp path] (4.533,4.131)--(4.533,3.951);
\node[gp node center] at (4.533,0.677) { 20};
\draw[gp path] (6.139,0.985)--(6.139,1.165);
\draw[gp path] (6.139,4.131)--(6.139,3.951);
\node[gp node center] at (6.139,0.677) { 30};
\draw[gp path] (7.746,0.985)--(7.746,1.165);
\draw[gp path] (7.746,4.131)--(7.746,3.951);
\node[gp node center] at (7.746,0.677) { 40};
\draw[gp path] (9.352,0.985)--(9.352,1.165);
\draw[gp path] (9.352,4.131)--(9.352,3.951);
\node[gp node center] at (9.352,0.677) { 50};
\draw[gp path] (10.959,0.985)--(10.959,1.165);
\draw[gp path] (10.959,4.131)--(10.959,3.951);
\node[gp node center] at (10.959,0.677) { 60};
\draw[gp path] (12.565,0.985)--(12.565,1.165);
\draw[gp path] (12.565,4.131)--(12.565,3.951);
\node[gp node center] at (12.565,0.677) { 70};
\draw[gp path] (1.320,4.131)--(1.320,0.985)--(13.047,0.985)--(13.047,4.131)--cycle;
\node[gp node center,rotate=-270] at (0.246,2.558) {BLEU};
\node[gp node center] at (7.183,0.215) {Training time (hours)};
\node[gp node right,font={\fontsize{8pt}{9.6pt}\selectfont}] at (11.579,1.935) {BIG model, batch size 2000, 1 GPU};
\gpcolor{rgb color={0.894,0.102,0.110}}
\gpsetlinetype{gp lt plot 0}
\gpsetlinewidth{2.00}
\draw[gp path] (11.763,1.935)--(12.679,1.935);
\draw[gp path] (1.866,0.985)--(1.966,1.420)--(2.126,1.886)--(2.287,2.072)--(2.448,2.247)%
  --(2.608,2.494)--(2.769,2.600)--(2.930,2.543)--(3.090,2.725)--(3.251,2.772)--(3.412,2.881)%
  --(3.572,2.884)--(3.733,3.011)--(3.894,3.012)--(4.054,3.027)--(4.215,3.130)--(4.376,3.227)%
  --(4.536,3.133)--(4.697,3.176)--(4.858,3.248)--(5.018,3.237)--(5.179,3.310)--(5.340,3.316)%
  --(5.500,3.348)--(5.661,3.331)--(5.822,3.417)--(5.982,3.381)--(6.143,3.388)--(6.303,3.405)%
  --(6.464,3.494)--(6.625,3.400)--(6.785,3.458)--(6.946,3.485)--(7.107,3.645)--(7.267,3.515)%
  --(7.428,3.623)--(7.589,3.561)--(7.749,3.456)--(7.910,3.583)--(8.071,3.550)--(8.231,3.627)%
  --(8.392,3.592)--(8.553,3.634)--(8.713,3.583)--(8.874,3.722)--(9.035,3.646)--(9.195,3.682)%
  --(9.356,3.723)--(9.517,3.794)--(9.677,3.774)--(9.838,3.661)--(9.999,3.740)--(10.159,3.755)%
  --(10.320,3.768)--(10.480,3.712)--(10.641,3.805)--(10.802,3.795)--(10.962,3.809)--(11.123,3.845)%
  --(11.284,3.828)--(11.444,3.789)--(11.605,3.863)--(11.766,3.791)--(11.926,3.845)--(12.087,3.870)%
  --(12.248,3.844)--(12.408,3.846)--(12.569,3.870)--(12.730,3.843)--(12.891,3.809)--(13.047,3.975);
\gpsetpointsize{8.00}
\gppoint{gp mark 5}{(12.408,3.846)}
\gppoint{gp mark 5}{(12.221,1.935)}
\gpcolor{color=gp lt color border}
\node[gp node right,font={\fontsize{8pt}{9.6pt}\selectfont}] at (11.579,1.627) {BASE model, batch size 4500, 1 GPU};
\gpcolor{rgb color={0.216,0.494,0.722}}
\gpsetlinetype{gp lt plot 3}
\draw[gp path] (11.763,1.627)--(12.679,1.627);
\draw[gp path] (1.684,0.985)--(1.805,1.534)--(1.965,2.043)--(2.126,2.188)--(2.286,2.381)%
  --(2.447,2.447)--(2.608,2.497)--(2.768,2.607)--(2.929,2.746)--(3.090,2.742)--(3.250,2.808)%
  --(3.411,2.861)--(3.572,2.895)--(3.732,2.965)--(3.893,3.021)--(4.054,2.974)--(4.214,2.949)%
  --(4.375,2.939)--(4.536,3.078)--(4.696,3.087)--(4.857,3.101)--(5.018,3.082)--(5.178,3.137)%
  --(5.339,3.191)--(5.500,3.188)--(5.660,3.184)--(5.821,3.175)--(5.981,3.188)--(6.142,3.199)%
  --(6.303,3.166)--(6.463,3.238)--(6.624,3.178)--(6.785,3.196)--(6.945,3.253)--(7.106,3.253)%
  --(7.267,3.314)--(7.427,3.205)--(7.588,3.256)--(7.749,3.234)--(7.909,3.339)--(8.070,3.283)%
  --(8.231,3.253)--(8.391,3.340)--(8.552,3.357)--(8.713,3.342)--(8.873,3.215)--(9.034,3.432)%
  --(9.195,3.481)--(9.355,3.334)--(9.516,3.279)--(9.676,3.324)--(9.837,3.454)--(9.998,3.404)%
  --(10.158,3.387)--(10.319,3.393)--(10.480,3.351)--(10.640,3.405)--(10.801,3.470)--(10.962,3.340)%
  --(11.122,3.368)--(11.283,3.365)--(11.444,3.404)--(11.604,3.403)--(11.765,3.424)--(11.926,3.396)%
  --(12.086,3.417)--(12.247,3.371)--(12.408,3.385)--(12.568,3.339)--(12.729,3.412)--(12.890,3.452)%
  --(13.047,3.446);
\gppoint{gp mark 7}{(12.408,3.385)}
\gppoint{gp mark 7}{(12.221,1.627)}
\gpcolor{color=gp lt color border}
\node[gp node right,font={\fontsize{8pt}{9.6pt}\selectfont}] at (11.579,1.319) {BASE model, batch size 2000, 1 GPU};
\gpcolor{rgb color={0.302,0.686,0.290}}
\gpsetlinetype{gp lt plot 0}
\draw[gp path] (11.763,1.319)--(12.679,1.319);
\draw[gp path] (1.669,0.985)--(1.804,1.246)--(1.965,1.455)--(2.126,1.655)--(2.286,1.746)%
  --(2.447,1.719)--(2.608,1.885)--(2.768,1.928)--(2.929,1.921)--(3.090,2.061)--(3.250,2.098)%
  --(3.411,2.156)--(3.572,2.125)--(3.732,2.223)--(3.893,2.253)--(4.054,2.277)--(4.214,2.216)%
  --(4.375,2.311)--(4.536,2.182)--(4.696,2.236)--(4.857,2.284)--(5.017,2.303)--(5.178,2.305)%
  --(5.339,2.246)--(5.499,2.276)--(5.660,2.302)--(5.821,2.307)--(5.981,2.364)--(6.142,2.426)%
  --(6.303,2.330)--(6.463,2.329)--(6.624,2.407)--(6.785,2.338)--(6.945,2.370)--(7.106,2.461)%
  --(7.267,2.327)--(7.427,2.514)--(7.588,2.400)--(7.749,2.382)--(7.909,2.436)--(8.070,2.419)%
  --(8.230,2.419)--(8.391,2.489)--(8.552,2.473)--(8.712,2.459)--(8.873,2.461)--(9.034,2.438)%
  --(9.194,2.523)--(9.355,2.480)--(9.516,2.430)--(9.676,2.566)--(9.837,2.474)--(9.998,2.408)%
  --(10.158,2.408)--(10.319,2.527)--(10.480,2.468)--(10.640,2.499)--(10.801,2.573)--(10.961,2.583)%
  --(11.122,2.532)--(11.283,2.573)--(11.443,2.541)--(11.604,2.487)--(11.765,2.488)--(11.925,2.640)%
  --(12.086,2.562)--(12.247,2.643)--(12.407,2.600)--(12.568,2.550)--(12.729,2.602)--(12.889,2.526)%
  --(13.047,2.603);
\gppoint{gp mark 9}{(12.407,2.600)}
\gppoint{gp mark 9}{(12.221,1.319)}
\gpcolor{color=gp lt color border}
\gpsetlinetype{gp lt border}
\gpsetlinewidth{1.00}
\draw[gp path] (1.320,4.131)--(1.320,0.985)--(13.047,0.985)--(13.047,4.131)--cycle;
\gpdefrectangularnode{gp plot 1}{\pgfpoint{1.320cm}{0.985cm}}{\pgfpoint{13.047cm}{4.131cm}}
\end{tikzpicture}

%% file: plots/8GPU-big1500-base4500.tex
\begin{tikzpicture}[gnuplot]
\path (0.000,0.000) rectangle (13.600,4.500);
\gpcolor{color=gp lt color border}
\gpsetlinetype{gp lt border}
\gpsetlinewidth{1.00}
\draw[gp path] (1.320,1.557)--(1.500,1.557);
\draw[gp path] (13.047,1.557)--(12.867,1.557);
\node[gp node right] at (1.136,1.557) { 22};
\draw[gp path] (1.320,2.129)--(1.500,2.129);
\draw[gp path] (13.047,2.129)--(12.867,2.129);
\node[gp node right] at (1.136,2.129) { 23};
\draw[gp path] (1.320,2.701)--(1.500,2.701);
\draw[gp path] (13.047,2.701)--(12.867,2.701);
\node[gp node right] at (1.136,2.701) { 24};
\draw[gp path] (1.320,3.273)--(1.500,3.273);
\draw[gp path] (13.047,3.273)--(12.867,3.273);
\node[gp node right] at (1.136,3.273) { 25};
\draw[gp path] (1.320,3.845)--(1.500,3.845);
\draw[gp path] (13.047,3.845)--(12.867,3.845);
\node[gp node right] at (1.136,3.845) { 26};
\draw[gp path] (1.320,0.985)--(1.320,1.165);
\draw[gp path] (1.320,4.131)--(1.320,3.951);
\node[gp node center] at (1.320,0.677) { 0};
\draw[gp path] (3.452,0.985)--(3.452,1.165);
\draw[gp path] (3.452,4.131)--(3.452,3.951);
\node[gp node center] at (3.452,0.677) { 10};
\draw[gp path] (5.584,0.985)--(5.584,1.165);
\draw[gp path] (5.584,4.131)--(5.584,3.951);
\node[gp node center] at (5.584,0.677) { 20};
\draw[gp path] (7.717,0.985)--(7.717,1.165);
\draw[gp path] (7.717,4.131)--(7.717,3.951);
\node[gp node center] at (7.717,0.677) { 30};
\draw[gp path] (9.849,0.985)--(9.849,1.165);
\draw[gp path] (9.849,4.131)--(9.849,3.951);
\node[gp node center] at (9.849,0.677) { 40};
\draw[gp path] (11.981,0.985)--(11.981,1.165);
\draw[gp path] (11.981,4.131)--(11.981,3.951);
\node[gp node center] at (11.981,0.677) { 50};
\draw[gp path] (1.320,4.131)--(1.320,0.985)--(13.047,0.985)--(13.047,4.131)--cycle;
\node[gp node center,rotate=-270] at (0.246,2.558) {BLEU};
\node[gp node center] at (7.183,0.215) {Training time (hours)};
\node[gp node right,font={\fontsize{8pt}{9.6pt}\selectfont}] at (11.579,1.627) {BIG model, batch size 1500, 8 GPUs};
\gpcolor{rgb color={0.894,0.102,0.110}}
\gpsetlinetype{gp lt plot 0}
\gpsetlinewidth{2.00}
\draw[gp path] (11.763,1.627)--(12.679,1.627);
\draw[gp path] (2.253,0.985)--(2.399,1.407)--(2.612,1.732)--(2.825,1.852)--(3.038,2.186)%
  --(3.251,2.308)--(3.465,2.514)--(3.678,2.609)--(3.891,2.733)--(4.104,2.653)--(4.317,2.732)%
  --(4.531,2.887)--(4.744,2.827)--(4.957,2.895)--(5.170,3.186)--(5.384,3.232)--(5.597,3.170)%
  --(5.810,3.156)--(6.023,3.351)--(6.236,3.204)--(6.450,3.444)--(6.663,3.363)--(6.876,3.353)%
  --(7.089,3.301)--(7.303,3.409)--(7.516,3.403)--(7.729,3.498)--(7.942,3.316)--(8.155,3.485)%
  --(8.369,3.432)--(8.582,3.581)--(8.795,3.597)--(9.008,3.603)--(9.222,3.566)--(9.435,3.611)%
  --(9.648,3.537)--(9.861,3.651)--(10.074,3.682)--(10.288,3.727)--(10.501,3.794)--(10.714,3.736)%
  --(10.927,3.683)--(11.140,3.586)--(11.354,3.813)--(11.567,3.696)--(11.780,3.772)--(11.993,3.723)%
  --(12.207,3.785)--(12.420,3.570)--(12.633,3.809)--(12.846,3.803)--(13.047,3.828);
\gpsetpointsize{8.00}
\gppoint{gp mark 5}{(11.780,3.772)}
\gppoint{gp mark 5}{(12.221,1.627)}
\gpcolor{color=gp lt color border}
\node[gp node right,font={\fontsize{8pt}{9.6pt}\selectfont}] at (11.579,1.319) {BASE model, batch size 4500, 8 GPUs};
\gpcolor{rgb color={0.216,0.494,0.722}}
\gpsetlinetype{gp lt plot 3}
\draw[gp path] (11.763,1.319)--(12.679,1.319);
\draw[gp path] (2.139,0.985)--(2.185,1.181)--(2.399,1.738)--(2.612,2.011)--(2.825,2.118)%
  --(3.038,2.107)--(3.252,2.397)--(3.465,2.490)--(3.678,2.499)--(3.891,2.606)--(4.105,2.476)%
  --(4.318,2.792)--(4.531,2.679)--(4.744,2.756)--(4.958,2.839)--(5.171,2.754)--(5.384,2.775)%
  --(5.597,2.899)--(5.810,2.851)--(6.024,2.951)--(6.237,2.837)--(6.450,2.995)--(6.663,3.018)%
  --(6.877,2.976)--(7.090,3.037)--(7.303,2.941)--(7.516,3.075)--(7.730,3.047)--(7.943,3.145)%
  --(8.156,3.127)--(8.369,3.208)--(8.583,3.078)--(8.796,3.047)--(9.009,3.179)--(9.222,3.106)%
  --(9.435,3.138)--(9.649,3.076)--(9.862,3.122)--(10.075,3.151)--(10.288,3.118)--(10.502,3.181)%
  --(10.715,3.101)--(10.928,3.120)--(11.141,3.085)--(11.355,3.115)--(11.568,3.223)--(11.781,3.172)%
  --(11.994,3.112)--(12.208,3.169)--(12.421,3.080)--(12.634,3.245)--(12.847,3.283)--(13.047,3.204);
\gppoint{gp mark 7}{(11.781,3.172)}
\gppoint{gp mark 7}{(12.221,1.319)}
\gpcolor{color=gp lt color border}
\gpsetlinetype{gp lt border}
\gpsetlinewidth{1.00}
\draw[gp path] (1.320,4.131)--(1.320,0.985)--(13.047,0.985)--(13.047,4.131)--cycle;
\gpdefrectangularnode{gp plot 1}{\pgfpoint{1.320cm}{0.985cm}}{\pgfpoint{13.047cm}{4.131cm}}
\end{tikzpicture}

%% file: plots/1GPU-max-length-b1500.tex
\begin{tikzpicture}[gnuplot]
\path (0.000,0.000) rectangle (13.600,4.500);
\gpcolor{color=gp lt color border}
\gpsetlinetype{gp lt border}
\gpsetlinewidth{1.00}
\draw[gp path] (1.320,0.985)--(1.500,0.985);
\draw[gp path] (9.187,0.985)--(9.007,0.985);
\node[gp node right] at (1.136,0.985) { 0};
\draw[gp path] (1.320,1.813)--(1.500,1.813);
\draw[gp path] (9.187,1.813)--(9.007,1.813);
\node[gp node right] at (1.136,1.813) { 5};
\draw[gp path] (1.320,2.641)--(1.500,2.641);
\draw[gp path] (9.187,2.641)--(9.007,2.641);
\node[gp node right] at (1.136,2.641) { 10};
\draw[gp path] (1.320,3.469)--(1.500,3.469);
\draw[gp path] (9.187,3.469)--(9.007,3.469);
\node[gp node right] at (1.136,3.469) { 15};
\draw[gp path] (1.320,0.985)--(1.320,1.165);
\draw[gp path] (1.320,4.131)--(1.320,3.951);
\node[gp node center] at (1.320,0.677) { 0};
\draw[gp path] (2.194,0.985)--(2.194,1.165);
\draw[gp path] (2.194,4.131)--(2.194,3.951);
\node[gp node center] at (2.194,0.677) { 1};
\draw[gp path] (3.068,0.985)--(3.068,1.165);
\draw[gp path] (3.068,4.131)--(3.068,3.951);
\node[gp node center] at (3.068,0.677) { 2};
\draw[gp path] (3.942,0.985)--(3.942,1.165);
\draw[gp path] (3.942,4.131)--(3.942,3.951);
\node[gp node center] at (3.942,0.677) { 3};
\draw[gp path] (4.816,0.985)--(4.816,1.165);
\draw[gp path] (4.816,4.131)--(4.816,3.951);
\node[gp node center] at (4.816,0.677) { 4};
\draw[gp path] (5.691,0.985)--(5.691,1.165);
\draw[gp path] (5.691,4.131)--(5.691,3.951);
\node[gp node center] at (5.691,0.677) { 5};
\draw[gp path] (6.565,0.985)--(6.565,1.165);
\draw[gp path] (6.565,4.131)--(6.565,3.951);
\node[gp node center] at (6.565,0.677) { 6};
\draw[gp path] (7.439,0.985)--(7.439,1.165);
\draw[gp path] (7.439,4.131)--(7.439,3.951);
\node[gp node center] at (7.439,0.677) { 7};
\draw[gp path] (8.313,0.985)--(8.313,1.165);
\draw[gp path] (8.313,4.131)--(8.313,3.951);
\node[gp node center] at (8.313,0.677) { 8};
\draw[gp path] (9.187,0.985)--(9.187,1.165);
\draw[gp path] (9.187,4.131)--(9.187,3.951);
\node[gp node center] at (9.187,0.677) { 9};
\draw[gp path] (1.320,4.131)--(1.320,0.985)--(9.187,0.985)--(9.187,4.131)--cycle;
\node[gp node center,rotate=-270] at (0.246,2.558) {BLEU};
\node[gp node center] at (5.253,0.215) {Training time (hours)};
\node[gp node right,font={\fontsize{8pt}{9.6pt}\selectfont}] at (12.131,2.679) {max length 400};
\gpcolor{rgb color={0.894,0.102,0.110}}
\gpsetlinetype{gp lt plot 0}
\gpsetlinewidth{2.00}
\draw[gp path] (12.315,2.679)--(13.231,2.679);
\draw[gp path] (1.340,0.985)--(2.213,1.966)--(3.088,2.443)--(3.962,3.078)--(4.836,3.371)%
  --(5.710,3.683)--(6.584,3.840)--(7.458,3.888)--(8.333,3.998)--(9.187,4.074);
\gpsetpointsize{8.00}
\gppoint{gp mark 5}{(6.584,3.840)}
\gppoint{gp mark 5}{(12.773,2.679)}
\gpcolor{color=gp lt color border}
\node[gp node right,font={\fontsize{8pt}{9.6pt}\selectfont}] at (12.131,2.371) {max length 200};
\gpcolor{rgb color={0.216,0.494,0.722}}
\draw[gp path] (12.315,2.371)--(13.231,2.371);
\draw[gp path] (1.338,0.985)--(2.212,2.020)--(3.086,2.255)--(3.960,2.676)--(4.834,3.051)%
  --(5.708,3.430)--(6.582,3.602)--(7.457,3.771)--(8.331,3.870)--(9.187,3.989);
\gppoint{gp mark 7}{(7.457,3.771)}
\gppoint{gp mark 7}{(12.773,2.371)}
\gpcolor{color=gp lt color border}
\node[gp node right,font={\fontsize{8pt}{9.6pt}\selectfont}] at (12.131,2.063) {max length 150};
\gpcolor{rgb color={0.302,0.686,0.290}}
\draw[gp path] (12.315,2.063)--(13.231,2.063);
\draw[gp path] (1.342,0.985)--(2.215,2.164)--(3.090,2.162)--(3.964,2.687)--(4.838,3.099)%
  --(5.712,3.343)--(6.586,3.527)--(7.460,3.669);
\gppoint{gp mark 9}{(6.586,3.527)}
\gppoint{gp mark 9}{(12.773,2.063)}
\gpcolor{color=gp lt color border}
\node[gp node right,font={\fontsize{8pt}{9.6pt}\selectfont}] at (12.131,1.755) {max length  70};
\gpcolor{rgb color={0.596,0.306,0.639}}
\draw[gp path] (12.315,1.755)--(13.231,1.755);
\draw[gp path] (1.337,0.986)--(2.211,2.117)--(3.085,2.110)--(3.959,2.023)--(4.833,1.863)%
  --(5.708,1.950)--(6.582,2.111)--(7.456,2.128)--(8.330,2.360)--(9.187,2.485);
\gppoint{gp mark 11}{(6.582,2.111)}
\gppoint{gp mark 11}{(12.773,1.755)}
\gpcolor{color=gp lt color border}
\node[gp node right,font={\fontsize{8pt}{9.6pt}\selectfont}] at (12.131,1.447) {max length  50};
\gpcolor{rgb color={1.000,0.498,0.000}}
\draw[gp path] (12.315,1.447)--(13.231,1.447);
\draw[gp path] (1.338,0.985)--(2.212,2.065)--(3.086,1.867)--(3.960,1.591)--(4.834,1.654)%
  --(5.709,1.794)--(6.583,1.814)--(7.457,1.839)--(8.331,1.878)--(9.187,1.905);
\gppoint{gp mark 13}{(6.583,1.814)}
\gppoint{gp mark 13}{(12.773,1.447)}
\gpcolor{color=gp lt color border}
\node[gp node right,font={\fontsize{8pt}{9.6pt}\selectfont}] at (12.131,1.139) {max length  25};
\gpcolor{rgb color={0.651,0.337,0.157}}
\draw[gp path] (12.315,1.139)--(13.231,1.139);
\draw[gp path] (1.338,0.985)--(2.212,1.693)--(3.086,1.374)--(3.960,1.274)--(4.834,1.218)%
  --(5.709,1.246)--(6.583,1.277)--(7.457,1.277);
\gppoint{gp mark 4}{(6.583,1.277)}
\gppoint{gp mark 4}{(12.773,1.139)}
\gpcolor{color=gp lt color border}
\gpsetlinetype{gp lt border}
\gpsetlinewidth{1.00}
\draw[gp path] (1.320,4.131)--(1.320,0.985)--(9.187,0.985)--(9.187,4.131)--cycle;
\gpdefrectangularnode{gp plot 1}{\pgfpoint{1.320cm}{0.985cm}}{\pgfpoint{9.187cm}{4.131cm}}
\end{tikzpicture}

%% file: plots/1GPU-base-m70-batch-size.tex
\begin{tikzpicture}[gnuplot]
\path (0.000,0.000) rectangle (13.600,6.000);
\gpcolor{color=gp lt color border}
\gpsetlinetype{gp lt border}
\gpsetlinewidth{1.00}
\draw[gp path] (1.320,1.566)--(1.500,1.566);
\draw[gp path] (13.047,1.566)--(12.867,1.566);
\node[gp node right] at (1.136,1.566) { 10};
\draw[gp path] (1.320,2.147)--(1.500,2.147);
\draw[gp path] (13.047,2.147)--(12.867,2.147);
\node[gp node right] at (1.136,2.147) { 12};
\draw[gp path] (1.320,2.727)--(1.500,2.727);
\draw[gp path] (13.047,2.727)--(12.867,2.727);
\node[gp node right] at (1.136,2.727) { 14};
\draw[gp path] (1.320,3.308)--(1.500,3.308);
\draw[gp path] (13.047,3.308)--(12.867,3.308);
\node[gp node right] at (1.136,3.308) { 16};
\draw[gp path] (1.320,3.889)--(1.500,3.889);
\draw[gp path] (13.047,3.889)--(12.867,3.889);
\node[gp node right] at (1.136,3.889) { 18};
\draw[gp path] (1.320,4.470)--(1.500,4.470);
\draw[gp path] (13.047,4.470)--(12.867,4.470);
\node[gp node right] at (1.136,4.470) { 20};
\draw[gp path] (1.320,5.050)--(1.500,5.050);
\draw[gp path] (13.047,5.050)--(12.867,5.050);
\node[gp node right] at (1.136,5.050) { 22};
\draw[gp path] (1.320,0.985)--(1.320,1.165);
\draw[gp path] (1.320,5.631)--(1.320,5.451);
\node[gp node center] at (1.320,0.677) { 0};
\draw[gp path] (2.949,0.985)--(2.949,1.165);
\draw[gp path] (2.949,5.631)--(2.949,5.451);
\node[gp node center] at (2.949,0.677) { 10};
\draw[gp path] (4.578,0.985)--(4.578,1.165);
\draw[gp path] (4.578,5.631)--(4.578,5.451);
\node[gp node center] at (4.578,0.677) { 20};
\draw[gp path] (6.206,0.985)--(6.206,1.165);
\draw[gp path] (6.206,5.631)--(6.206,5.451);
\node[gp node center] at (6.206,0.677) { 30};
\draw[gp path] (7.835,0.985)--(7.835,1.165);
\draw[gp path] (7.835,5.631)--(7.835,5.451);
\node[gp node center] at (7.835,0.677) { 40};
\draw[gp path] (9.464,0.985)--(9.464,1.165);
\draw[gp path] (9.464,5.631)--(9.464,5.451);
\node[gp node center] at (9.464,0.677) { 50};
\draw[gp path] (11.093,0.985)--(11.093,1.165);
\draw[gp path] (11.093,5.631)--(11.093,5.451);
\node[gp node center] at (11.093,0.677) { 60};
\draw[gp path] (12.721,0.985)--(12.721,1.165);
\draw[gp path] (12.721,5.631)--(12.721,5.451);
\node[gp node center] at (12.721,0.677) { 70};
\draw[gp path] (1.320,5.631)--(1.320,0.985)--(13.047,0.985)--(13.047,5.631)--cycle;
\node[gp node center,rotate=-270] at (0.246,3.308) {BLEU};
\node[gp node center] at (7.183,0.215) {Training time (hours)};
\node[gp node right,font={\fontsize{8pt}{9.6pt}\selectfont}] at (11.579,2.551) {BASE, batch size 6000};
\gpcolor{rgb color={0.894,0.102,0.110}}
\gpsetlinetype{gp lt plot 0}
\gpsetlinewidth{2.00}
\draw[gp path] (11.763,2.551)--(12.679,2.551);
\draw[gp path] (1.477,0.985)--(1.485,1.119)--(1.648,2.571)--(1.811,3.238)--(1.974,3.809)%
  --(2.137,3.969)--(2.300,4.196)--(2.463,4.335)--(2.626,4.428)--(2.789,4.532)--(2.952,4.547)%
  --(3.114,4.702)--(3.277,4.666)--(3.440,4.769)--(3.603,4.797)--(3.766,4.776)--(3.929,4.739)%
  --(4.092,4.879)--(4.255,4.847)--(4.417,4.905)--(4.580,4.862)--(4.743,4.884)--(4.906,4.905)%
  --(5.069,5.040)--(5.232,5.029)--(5.395,4.980)--(5.558,4.977)--(5.721,5.051)--(5.883,5.065)%
  --(6.046,5.050)--(6.209,5.079)--(6.372,4.987)--(6.535,5.116)--(6.698,5.122)--(6.861,5.156)%
  --(7.024,5.190)--(7.187,5.168)--(7.349,5.162)--(7.512,5.079)--(7.675,5.152)--(7.838,5.165)%
  --(8.001,5.240)--(8.164,5.156)--(8.327,5.175)--(8.490,5.198)--(8.652,5.228)--(8.815,5.217)%
  --(8.978,5.236)--(9.141,5.178)--(9.304,5.197)--(9.467,5.199)--(9.630,5.252)--(9.793,5.288)%
  --(9.956,5.213)--(10.119,5.234)--(10.281,5.232)--(10.444,5.288)--(10.607,5.330)--(10.770,5.321)%
  --(10.933,5.301)--(11.096,5.248)--(11.259,5.355)--(11.422,5.294)--(11.585,5.215)--(11.747,5.271)%
  --(11.910,5.244)--(12.073,5.310)--(12.236,5.388)--(12.399,5.285)--(12.562,5.296)--(12.725,5.270)%
  --(12.888,5.282)--(13.047,5.368);
\gpsetpointsize{8.00}
\gppoint{gp mark 5}{(12.236,5.388)}
\gppoint{gp mark 5}{(12.221,2.551)}
\gpcolor{color=gp lt color border}
\node[gp node right,font={\fontsize{8pt}{9.6pt}\selectfont}] at (11.579,2.243) {BASE, batch size 4500};
\gpcolor{rgb color={0.216,0.494,0.722}}
\gpsetlinetype{gp lt plot 3}
\draw[gp path] (11.763,2.243)--(12.679,2.243);
\draw[gp path] (1.462,0.985)--(1.485,1.375)--(1.648,2.605)--(1.811,3.299)--(1.974,3.686)%
  --(2.137,4.075)--(2.300,4.056)--(2.463,4.183)--(2.626,4.343)--(2.789,4.388)--(2.951,4.447)%
  --(3.114,4.564)--(3.277,4.526)--(3.440,4.525)--(3.603,4.637)--(3.766,4.725)--(3.929,4.727)%
  --(4.092,4.736)--(4.255,4.746)--(4.417,4.911)--(4.580,4.797)--(4.743,4.837)--(4.906,4.794)%
  --(5.069,4.936)--(5.232,4.859)--(5.395,4.904)--(5.558,4.885)--(5.721,4.904)--(5.883,4.866)%
  --(6.046,4.885)--(6.209,4.952)--(6.372,4.972)--(6.535,5.079)--(6.698,4.914)--(6.861,4.903)%
  --(7.024,4.995)--(7.186,5.033)--(7.349,5.016)--(7.512,5.032)--(7.675,5.029)--(7.838,5.072)%
  --(8.001,5.111)--(8.164,5.047)--(8.327,5.025)--(8.490,5.072)--(8.652,5.084)--(8.815,5.109)%
  --(8.978,5.086)--(9.141,5.041)--(9.304,5.088)--(9.467,5.072)--(9.630,5.108)--(9.793,5.134)%
  --(9.956,5.117)--(10.118,5.119)--(10.281,5.149)--(10.444,5.102)--(10.770,5.121)--(10.933,5.146)%
  --(11.096,5.218)--(11.259,5.157)--(11.422,5.160)--(11.584,5.214)--(11.747,5.168)--(11.910,5.220)%
  --(12.073,5.218)--(12.236,5.159)--(12.399,5.164)--(12.562,5.145)--(12.725,5.202)--(12.888,5.184)%
  --(13.047,5.280);
\gppoint{gp mark 7}{(12.399,5.164)}
\gppoint{gp mark 7}{(12.221,2.243)}
\gpcolor{color=gp lt color border}
\node[gp node right,font={\fontsize{8pt}{9.6pt}\selectfont}] at (11.579,1.935) {BASE, batch size 3000};
\gpcolor{rgb color={0.302,0.686,0.290}}
\gpsetlinetype{gp lt plot 0}
\draw[gp path] (11.763,1.935)--(12.679,1.935);
\draw[gp path] (1.448,0.985)--(1.485,1.681)--(1.648,2.920)--(1.811,3.314)--(1.974,3.612)%
  --(2.137,3.748)--(2.300,3.881)--(2.463,3.924)--(2.626,4.062)--(2.788,4.147)--(2.951,4.102)%
  --(3.114,4.155)--(3.277,4.303)--(3.440,4.302)--(3.603,4.322)--(3.766,4.385)--(3.929,4.420)%
  --(4.092,4.377)--(4.254,4.400)--(4.417,4.498)--(4.580,4.517)--(4.743,4.559)--(4.906,4.574)%
  --(5.069,4.497)--(5.232,4.471)--(5.395,4.476)--(5.557,4.542)--(5.720,4.582)--(5.883,4.563)%
  --(6.046,4.551)--(6.209,4.602)--(6.372,4.612)--(6.535,4.608)--(6.698,4.643)--(6.861,4.569)%
  --(7.023,4.718)--(7.186,4.593)--(7.349,4.618)--(7.512,4.622)--(7.675,4.632)--(7.838,4.617)%
  --(8.001,4.642)--(8.164,4.677)--(8.326,4.645)--(8.489,4.626)--(8.652,4.673)--(8.815,4.705)%
  --(8.978,4.743)--(9.141,4.708)--(9.304,4.767)--(9.467,4.770)--(9.629,4.690)--(9.792,4.779)%
  --(9.955,4.703)--(10.118,4.665)--(10.281,4.727)--(10.444,4.725)--(10.607,4.682)--(10.770,4.728)%
  --(10.933,4.726)--(11.095,4.688)--(11.258,4.777)--(11.421,4.808)--(11.584,4.742)--(11.747,4.760)%
  --(11.910,4.729)--(12.073,4.767)--(12.236,4.840)--(12.398,4.753)--(12.561,4.791)--(12.724,4.731)%
  --(12.887,4.749)--(13.047,4.820);
\gppoint{gp mark 9}{(12.236,4.840)}
\gppoint{gp mark 9}{(12.221,1.935)}
\gpcolor{color=gp lt color border}
\node[gp node right,font={\fontsize{8pt}{9.6pt}\selectfont}] at (11.579,1.627) {BASE, batch size 1500};
\gpcolor{rgb color={0.596,0.306,0.639}}
\draw[gp path] (11.763,1.627)--(12.679,1.627);
\draw[gp path] (1.452,0.985)--(1.485,1.587)--(1.648,2.497)--(1.811,2.833)--(1.974,2.988)%
  --(2.137,3.034)--(2.300,3.170)--(2.463,3.198)--(2.625,3.261)--(2.788,3.247)--(2.951,3.292)%
  --(3.114,3.324)--(3.277,3.334)--(3.440,3.420)--(3.603,3.466)--(3.766,3.451)--(3.929,3.532)%
  --(4.091,3.594)--(4.254,3.601)--(4.417,3.576)--(4.580,3.587)--(4.743,3.683)--(4.906,3.599)%
  --(5.069,3.585)--(5.232,3.682)--(5.394,3.655)--(5.557,3.648)--(5.720,3.667)--(5.883,3.639)%
  --(6.046,3.683)--(6.209,3.763)--(6.372,3.713)--(6.535,3.703)--(6.697,3.711)--(6.860,3.797)%
  --(7.023,3.842)--(7.186,3.830)--(7.349,3.787)--(7.512,3.785)--(7.675,3.766)--(7.838,3.847)%
  --(8.000,3.878)--(8.163,3.798)--(8.326,3.717)--(8.489,3.931)--(8.652,3.752)--(8.815,3.837)%
  --(8.978,3.931)--(9.141,3.907)--(9.304,3.891)--(9.466,3.882)--(9.629,3.813)--(9.792,3.898)%
  --(9.955,3.925)--(10.118,3.952)--(10.281,3.776)--(10.444,3.887)--(10.607,3.913)--(10.769,3.959)%
  --(10.932,3.952)--(11.095,3.822)--(11.258,3.933)--(11.421,3.941)--(11.584,3.838)--(11.747,3.892)%
  --(11.910,3.944)--(12.072,3.930)--(12.235,3.886)--(12.398,3.922)--(12.561,3.881)--(12.724,3.906)%
  --(12.887,3.891)--(13.047,3.943);
\gppoint{gp mark 11}{(12.235,3.886)}
\gppoint{gp mark 11}{(12.221,1.627)}
\gpcolor{color=gp lt color border}
\node[gp node right,font={\fontsize{8pt}{9.6pt}\selectfont}] at (11.579,1.319) {BASE, batch size 1000};
\gpcolor{rgb color={1.000,0.498,0.000}}
\draw[gp path] (11.763,1.319)--(12.679,1.319);
\draw[gp path] (1.468,0.985)--(1.485,1.265)--(1.648,2.039)--(1.811,2.221)--(1.974,2.197)%
  --(2.137,2.475)--(2.300,2.572)--(2.463,2.615)--(2.626,2.638)--(2.788,2.751)--(2.951,2.786)%
  --(3.114,2.794)--(3.277,2.705)--(3.440,2.890)--(3.603,2.743)--(3.766,2.847)--(3.929,2.860)%
  --(4.091,2.805)--(4.254,3.021)--(4.417,2.916)--(4.580,2.900)--(4.743,2.999)--(4.906,2.918)%
  --(5.069,3.001)--(5.232,2.994)--(5.394,3.038)--(5.557,3.035)--(5.720,3.051)--(5.883,3.029)%
  --(6.209,2.959)--(6.372,3.064)--(6.535,3.051)--(6.697,3.077)--(6.860,3.118)--(7.023,3.164)%
  --(7.186,3.075)--(7.349,3.135)--(7.512,3.270)--(7.675,3.068)--(7.838,3.101)--(8.000,3.170)%
  --(8.163,3.090)--(8.326,3.089)--(8.489,3.063)--(8.652,3.128)--(8.815,3.138)--(8.978,3.207)%
  --(9.141,3.227)--(9.303,3.120)--(9.466,3.297)--(9.629,3.218)--(9.792,3.116)--(9.955,3.124)%
  --(10.118,3.193)--(10.281,3.214)--(10.444,3.138)--(10.607,3.274)--(10.769,3.219)--(10.932,3.245)%
  --(11.095,3.187)--(11.258,3.247)--(11.421,3.232)--(11.584,3.208)--(11.747,3.227)--(11.910,3.228)%
  --(12.072,3.238)--(12.235,3.114)--(12.398,3.182)--(12.561,3.292)--(12.724,3.308)--(12.887,3.248)%
  --(13.047,3.349);
\gppoint{gp mark 13}{(12.398,3.182)}
\gppoint{gp mark 13}{(12.221,1.319)}
\gpcolor{color=gp lt color border}
\gpsetlinetype{gp lt border}
\gpsetlinewidth{1.00}
\draw[gp path] (1.320,5.631)--(1.320,0.985)--(13.047,0.985)--(13.047,5.631)--cycle;
\gpdefrectangularnode{gp plot 1}{\pgfpoint{1.320cm}{0.985cm}}{\pgfpoint{13.047cm}{5.631cm}}
\end{tikzpicture}

%% file: plots/1GPU-big-m70-batch-size.tex
\begin{tikzpicture}[gnuplot]
\path (0.000,0.000) rectangle (13.600,5.500);
\gpcolor{color=gp lt color border}
\gpsetlinetype{gp lt border}
\gpsetlinewidth{1.00}
\draw[gp path] (1.320,0.985)--(1.500,0.985);
\draw[gp path] (13.047,0.985)--(12.867,0.985);
\node[gp node right] at (1.136,0.985) { 0};
\draw[gp path] (1.320,1.849)--(1.500,1.849);
\draw[gp path] (13.047,1.849)--(12.867,1.849);
\node[gp node right] at (1.136,1.849) { 5};
\draw[gp path] (1.320,2.713)--(1.500,2.713);
\draw[gp path] (13.047,2.713)--(12.867,2.713);
\node[gp node right] at (1.136,2.713) { 10};
\draw[gp path] (1.320,3.576)--(1.500,3.576);
\draw[gp path] (13.047,3.576)--(12.867,3.576);
\node[gp node right] at (1.136,3.576) { 15};
\draw[gp path] (1.320,4.440)--(1.500,4.440);
\draw[gp path] (13.047,4.440)--(12.867,4.440);
\node[gp node right] at (1.136,4.440) { 20};
\draw[gp path] (1.320,0.985)--(1.320,1.165);
\draw[gp path] (1.320,5.131)--(1.320,4.951);
\node[gp node center] at (1.320,0.677) { 0};
\draw[gp path] (2.995,0.985)--(2.995,1.165);
\draw[gp path] (2.995,5.131)--(2.995,4.951);
\node[gp node center] at (2.995,0.677) { 5};
\draw[gp path] (4.671,0.985)--(4.671,1.165);
\draw[gp path] (4.671,5.131)--(4.671,4.951);
\node[gp node center] at (4.671,0.677) { 10};
\draw[gp path] (6.346,0.985)--(6.346,1.165);
\draw[gp path] (6.346,5.131)--(6.346,4.951);
\node[gp node center] at (6.346,0.677) { 15};
\draw[gp path] (8.021,0.985)--(8.021,1.165);
\draw[gp path] (8.021,5.131)--(8.021,4.951);
\node[gp node center] at (8.021,0.677) { 20};
\draw[gp path] (9.696,0.985)--(9.696,1.165);
\draw[gp path] (9.696,5.131)--(9.696,4.951);
\node[gp node center] at (9.696,0.677) { 25};
\draw[gp path] (11.372,0.985)--(11.372,1.165);
\draw[gp path] (11.372,5.131)--(11.372,4.951);
\node[gp node center] at (11.372,0.677) { 30};
\draw[gp path] (13.047,0.985)--(13.047,1.165);
\draw[gp path] (13.047,5.131)--(13.047,4.951);
\node[gp node center] at (13.047,0.677) { 35};
\draw[gp path] (1.320,5.131)--(1.320,0.985)--(13.047,0.985)--(13.047,5.131)--cycle;
\node[gp node center,rotate=-270] at (0.246,3.058) {BLEU};
\node[gp node center] at (7.183,0.215) {Training time (hours)};
\node[gp node right,font={\fontsize{8pt}{9.6pt}\selectfont}] at (11.579,2.859) {BIG, batch size 2000};
\gpcolor{rgb color={0.894,0.102,0.110}}
\gpsetlinetype{gp lt plot 0}
\gpsetlinewidth{2.00}
\draw[gp path] (11.763,2.859)--(12.679,2.859);
\draw[gp path] (1.327,0.985)--(1.662,2.057)--(1.997,2.758)--(2.332,3.370)--(2.667,3.696)%
  --(3.002,3.977)--(3.337,4.051)--(3.672,4.173)--(4.007,4.258)--(4.342,4.331)--(4.677,4.383)%
  --(5.013,4.469)--(5.348,4.553)--(5.683,4.549)--(6.018,4.591)--(6.353,4.616)--(6.688,4.678)%
  --(7.023,4.665)--(7.358,4.669)--(7.693,4.707)--(8.028,4.723)--(8.363,4.730)--(8.698,4.767)%
  --(9.033,4.811)--(9.369,4.808)--(9.704,4.817)--(10.039,4.841)--(10.374,4.853)--(10.709,4.853)%
  --(11.044,4.889)--(11.379,4.857)--(11.714,4.925)--(12.049,4.930)--(12.384,4.933)--(12.719,4.932)%
  --(13.047,4.951);
\gpsetpointsize{8.00}
\gppoint{gp mark 5}{(11.044,4.889)}
\gppoint{gp mark 5}{(12.221,2.859)}
\gpcolor{color=gp lt color border}
\node[gp node right,font={\fontsize{8pt}{9.6pt}\selectfont}] at (11.579,2.551) {BIG, batch size 1500};
\gpcolor{rgb color={0.216,0.494,0.722}}
\draw[gp path] (11.763,2.551)--(12.679,2.551);
\draw[gp path] (1.327,0.985)--(1.662,2.097)--(1.997,2.612)--(2.332,3.302)--(2.667,3.531)%
  --(3.002,3.828)--(3.337,3.916)--(3.672,4.068)--(4.008,4.117)--(4.343,4.266)--(4.678,4.276)%
  --(5.013,4.384)--(5.348,4.384)--(5.683,4.458)--(6.018,4.504)--(6.353,4.524)--(6.688,4.528)%
  --(7.023,4.597)--(7.358,4.627)--(7.693,4.652)--(8.028,4.652)--(8.363,4.651)--(8.699,4.713)%
  --(9.034,4.724)--(9.369,4.754)--(9.704,4.719)--(10.039,4.776)--(10.374,4.792)--(10.709,4.781)%
  --(11.044,4.801)--(11.379,4.795)--(11.714,4.824)--(12.049,4.823)--(12.384,4.833)--(12.719,4.853)%
  --(13.047,4.855);
\gppoint{gp mark 7}{(10.374,4.792)}
\gppoint{gp mark 7}{(12.221,2.551)}
\gpcolor{color=gp lt color border}
\node[gp node right,font={\fontsize{8pt}{9.6pt}\selectfont}] at (11.579,2.243) {BIG, batch size 1450};
\gpcolor{rgb color={0.302,0.686,0.290}}
\draw[gp path] (11.763,2.243)--(12.679,2.243);
\draw[gp path] (1.327,0.985)--(1.662,2.054)--(1.997,2.619)--(2.332,3.242)--(2.669,3.551)%
  --(3.002,3.785)--(3.337,3.949)--(3.672,4.097)--(4.008,4.168)--(4.678,4.310)--(5.013,4.323)%
  --(5.348,4.441)--(5.683,4.463)--(6.018,4.487)--(6.353,4.496)--(6.688,4.553)--(7.023,4.539)%
  --(7.358,4.617)--(7.693,4.658)--(8.028,4.682)--(8.364,4.657)--(8.699,4.681)--(9.034,4.709)%
  --(9.369,4.655)--(9.704,4.679)--(10.039,4.716)--(10.374,4.723);
\gppoint{gp mark 9}{(10.039,4.716)}
\gppoint{gp mark 9}{(12.221,2.243)}
\gpcolor{color=gp lt color border}
\node[gp node right,font={\fontsize{8pt}{9.6pt}\selectfont}] at (11.579,1.935) {BIG, batch size 1400};
\gpcolor{rgb color={0.596,0.306,0.639}}
\draw[gp path] (11.763,1.935)--(12.679,1.935);
\draw[gp path] (1.327,0.985)--(1.662,2.014)--(1.997,1.891)--(2.332,1.578)--(2.667,1.814)%
  --(3.002,1.808)--(3.337,1.823)--(3.672,1.893)--(4.007,1.995)--(4.342,2.007)--(4.677,2.144)%
  --(5.013,2.269)--(5.348,2.387)--(5.683,2.434)--(6.018,2.522)--(6.353,2.617)--(6.688,2.611)%
  --(7.023,2.790)--(7.358,2.821)--(7.693,2.927)--(8.028,2.988)--(8.363,3.059)--(8.698,3.121)%
  --(9.033,3.147)--(9.368,3.304);
\gppoint{gp mark 11}{(7.693,2.927)}
\gppoint{gp mark 11}{(12.221,1.935)}
\gpcolor{color=gp lt color border}
\node[gp node right,font={\fontsize{8pt}{9.6pt}\selectfont}] at (11.579,1.627) {BIG, batch size 1300};
\gpcolor{rgb color={1.000,0.498,0.000}}
\draw[gp path] (11.763,1.627)--(12.679,1.627);
\draw[gp path] (1.327,0.985)--(1.662,2.015)--(1.997,1.751)--(2.332,1.600)--(2.667,1.649)%
  --(3.002,1.538)--(3.337,1.562)--(3.672,1.510)--(4.007,1.661)--(4.342,1.637)--(4.677,1.610)%
  --(5.013,1.731)--(5.348,1.720)--(5.683,1.779)--(6.018,1.795)--(6.353,1.861)--(6.688,1.945)%
  --(7.023,1.938)--(7.358,2.038)--(7.693,2.045)--(8.028,2.116)--(8.363,2.054)--(8.698,2.149);
\gppoint{gp mark 13}{(7.693,2.045)}
\gppoint{gp mark 13}{(12.221,1.627)}
\gpcolor{color=gp lt color border}
\node[gp node right,font={\fontsize{8pt}{9.6pt}\selectfont}] at (11.579,1.319) {BIG, batch size 1000};
\gpcolor{rgb color={0.651,0.337,0.157}}
\draw[gp path] (11.763,1.319)--(12.679,1.319);
\draw[gp path] (1.327,0.985)--(1.662,1.750)--(1.997,1.058)--(2.332,1.072)--(2.667,1.067)%
  --(3.002,1.050)--(3.337,1.047)--(3.672,1.014)--(4.007,1.040)--(4.342,1.023)--(4.677,1.023)%
  --(5.012,1.030);
\gppoint{gp mark 4}{(4.342,1.023)}
\gppoint{gp mark 4}{(12.221,1.319)}
\gpcolor{color=gp lt color border}
\gpsetlinetype{gp lt border}
\gpsetlinewidth{1.00}
\draw[gp path] (1.320,5.131)--(1.320,0.985)--(13.047,0.985)--(13.047,5.131)--cycle;
\gpdefrectangularnode{gp plot 1}{\pgfpoint{1.320cm}{0.985cm}}{\pgfpoint{13.047cm}{5.131cm}}
\end{tikzpicture}

%% file: plots/1GPU-czeng1-learning-rate.tex
\begin{tikzpicture}[gnuplot]
\path (0.000,0.000) rectangle (13.600,5.000);
\gpcolor{color=gp lt color border}
\gpsetlinetype{gp lt border}
\gpsetlinewidth{1.00}
\draw[gp path] (1.320,0.985)--(1.500,0.985);
\draw[gp path] (13.047,0.985)--(12.867,0.985);
\node[gp node right] at (1.136,0.985) { 0};
\draw[gp path] (1.320,1.729)--(1.500,1.729);
\draw[gp path] (13.047,1.729)--(12.867,1.729);
\node[gp node right] at (1.136,1.729) { 5};
\draw[gp path] (1.320,2.473)--(1.500,2.473);
\draw[gp path] (13.047,2.473)--(12.867,2.473);
\node[gp node right] at (1.136,2.473) { 10};
\draw[gp path] (1.320,3.217)--(1.500,3.217);
\draw[gp path] (13.047,3.217)--(12.867,3.217);
\node[gp node right] at (1.136,3.217) { 15};
\draw[gp path] (1.320,3.961)--(1.500,3.961);
\draw[gp path] (13.047,3.961)--(12.867,3.961);
\node[gp node right] at (1.136,3.961) { 20};
\draw[gp path] (1.320,0.985)--(1.320,1.165);
\draw[gp path] (1.320,4.631)--(1.320,4.451);
\node[gp node center] at (1.320,0.677) { 0};
\draw[gp path] (2.623,0.985)--(2.623,1.165);
\draw[gp path] (2.623,4.631)--(2.623,4.451);
\node[gp node center] at (2.623,0.677) { 5};
\draw[gp path] (3.926,0.985)--(3.926,1.165);
\draw[gp path] (3.926,4.631)--(3.926,4.451);
\node[gp node center] at (3.926,0.677) { 10};
\draw[gp path] (5.229,0.985)--(5.229,1.165);
\draw[gp path] (5.229,4.631)--(5.229,4.451);
\node[gp node center] at (5.229,0.677) { 15};
\draw[gp path] (6.532,0.985)--(6.532,1.165);
\draw[gp path] (6.532,4.631)--(6.532,4.451);
\node[gp node center] at (6.532,0.677) { 20};
\draw[gp path] (7.835,0.985)--(7.835,1.165);
\draw[gp path] (7.835,4.631)--(7.835,4.451);
\node[gp node center] at (7.835,0.677) { 25};
\draw[gp path] (9.138,0.985)--(9.138,1.165);
\draw[gp path] (9.138,4.631)--(9.138,4.451);
\node[gp node center] at (9.138,0.677) { 30};
\draw[gp path] (10.441,0.985)--(10.441,1.165);
\draw[gp path] (10.441,4.631)--(10.441,4.451);
\node[gp node center] at (10.441,0.677) { 35};
\draw[gp path] (11.744,0.985)--(11.744,1.165);
\draw[gp path] (11.744,4.631)--(11.744,4.451);
\node[gp node center] at (11.744,0.677) { 40};
\draw[gp path] (13.047,0.985)--(13.047,1.165);
\draw[gp path] (13.047,4.631)--(13.047,4.451);
\node[gp node center] at (13.047,0.677) { 45};
\draw[gp path] (1.320,4.631)--(1.320,0.985)--(13.047,0.985)--(13.047,4.631)--cycle;
\node[gp node center,rotate=-270] at (0.246,2.808) {BLEU};
\node[gp node center] at (7.183,0.215) {Training time (hours)};
\node[gp node right,font={\fontsize{8pt}{9.6pt}\selectfont}] at (11.579,2.551) {learning rate 0.25};
\gpcolor{rgb color={0.894,0.102,0.110}}
\gpsetlinetype{gp lt plot 0}
\gpsetlinewidth{2.00}
\draw[gp path] (11.763,2.551)--(12.679,2.551);
\draw[gp path] (1.335,0.985)--(1.596,2.072)--(1.856,2.462)--(2.117,3.015)--(2.378,3.351)%
  --(2.638,3.527)--(2.899,3.581)--(3.159,3.734)--(3.420,3.809)--(3.681,3.848)--(3.941,3.881)%
  --(4.202,3.915)--(4.462,3.978)--(4.723,4.003)--(4.984,3.987)--(5.244,4.098)--(5.505,4.081)%
  --(5.765,4.134)--(6.026,4.163)--(6.287,4.178)--(6.547,4.154)--(6.808,4.195);
\gpsetpointsize{8.00}
\gppoint{gp mark 5}{(6.287,4.178)}
\gppoint{gp mark 5}{(12.221,2.551)}
\gpcolor{color=gp lt color border}
\node[gp node right,font={\fontsize{8pt}{9.6pt}\selectfont}] at (11.579,2.243) {learning rate 0.20};
\gpcolor{rgb color={0.216,0.494,0.722}}
\draw[gp path] (11.763,2.243)--(12.679,2.243);
\draw[gp path] (1.335,0.985)--(1.596,2.063)--(1.856,2.714)--(2.117,3.176)--(2.378,3.401)%
  --(2.638,3.557)--(2.899,3.688)--(3.159,3.737)--(3.420,3.868)--(3.681,3.929)--(3.941,3.958)%
  --(4.202,4.019)--(4.462,4.040)--(4.723,4.062)--(4.984,4.081)--(5.244,4.163)--(5.505,4.180)%
  --(5.765,4.193)--(6.026,4.202)--(6.287,4.210)--(6.547,4.247)--(6.808,4.232)--(7.068,4.260)%
  --(7.329,4.261)--(7.590,4.303)--(7.850,4.291)--(8.111,4.294)--(8.371,4.304)--(8.632,4.327)%
  --(8.893,4.333)--(9.153,4.390)--(9.414,4.345)--(9.674,4.428)--(9.935,4.361)--(10.196,4.428)%
  --(10.456,4.395)--(10.717,4.375)--(10.977,4.383)--(11.238,4.400)--(11.499,4.423)--(11.759,4.382)%
  --(12.020,4.418)--(12.280,4.435)--(12.541,4.412)--(12.802,4.446)--(13.047,4.453);
\gppoint{gp mark 7}{(10.196,4.428)}
\gppoint{gp mark 7}{(12.221,2.243)}
\gpcolor{color=gp lt color border}
\node[gp node right,font={\fontsize{8pt}{9.6pt}\selectfont}] at (11.579,1.935) {learning rate 0.10};
\gpcolor{rgb color={0.302,0.686,0.290}}
\draw[gp path] (11.763,1.935)--(12.679,1.935);
\draw[gp path] (1.335,0.985)--(1.596,1.879)--(1.856,2.769)--(2.117,3.222)--(2.378,3.427)%
  --(2.638,3.563)--(2.899,3.690)--(3.159,3.826)--(3.420,3.867)--(3.681,3.924)--(3.941,3.983)%
  --(4.202,4.043)--(4.462,4.039)--(4.723,4.051)--(4.984,4.154)--(5.244,4.141)--(5.505,4.186)%
  --(5.765,4.210)--(6.026,4.199)--(6.287,4.226)--(6.547,4.273)--(6.808,4.261);
\gppoint{gp mark 9}{(5.765,4.210)}
\gppoint{gp mark 9}{(12.221,1.935)}
\gpcolor{color=gp lt color border}
\node[gp node right,font={\fontsize{8pt}{9.6pt}\selectfont}] at (11.579,1.627) {learning rate 0.05};
\gpcolor{rgb color={0.596,0.306,0.639}}
\draw[gp path] (11.763,1.627)--(12.679,1.627);
\draw[gp path] (1.335,0.985)--(1.596,1.426)--(1.856,2.631)--(2.117,3.109)--(2.378,3.310)%
  --(2.638,3.511)--(2.899,3.586)--(3.159,3.715)--(3.420,3.811)--(3.681,3.848)--(3.941,3.953)%
  --(4.202,3.943)--(4.462,4.013)--(4.723,4.044)--(4.984,4.103)--(5.244,4.079)--(5.505,4.101)%
  --(5.765,4.163)--(6.026,4.155)--(6.287,4.139)--(6.547,4.171)--(6.808,4.211)--(7.068,4.252)%
  --(7.329,4.228)--(7.590,4.252)--(7.850,4.254)--(8.111,4.262)--(8.371,4.297)--(8.632,4.314)%
  --(8.893,4.307)--(9.153,4.291)--(9.414,4.337)--(9.674,4.318)--(9.935,4.355)--(10.196,4.334)%
  --(10.456,4.356)--(10.717,4.379)--(10.977,4.352)--(11.238,4.375)--(11.499,4.383)--(11.759,4.422)%
  --(12.020,4.389)--(12.280,4.427)--(12.541,4.404)--(12.802,4.411)--(13.047,4.402);
\gppoint{gp mark 11}{(10.717,4.379)}
\gppoint{gp mark 11}{(12.221,1.627)}
\gpcolor{color=gp lt color border}
\node[gp node right,font={\fontsize{8pt}{9.6pt}\selectfont}] at (11.579,1.319) {learning rate 0.01};
\gpcolor{rgb color={1.000,0.498,0.000}}
\draw[gp path] (11.763,1.319)--(12.679,1.319);
\draw[gp path] (1.335,0.985)--(1.596,1.097)--(1.856,1.388)--(2.117,1.806)--(2.378,2.216)%
  --(2.638,2.452)--(2.899,2.680)--(3.159,2.819)--(3.420,2.887)--(3.681,3.032)--(3.941,3.061)%
  --(4.202,3.157)--(4.462,3.163)--(4.723,3.242)--(4.984,3.274)--(5.244,3.291)--(5.505,3.353)%
  --(5.765,3.402)--(6.026,3.371)--(6.287,3.444)--(6.547,3.470)--(6.808,3.471)--(7.068,3.497)%
  --(7.329,3.522)--(7.590,3.562)--(7.850,3.585)--(8.111,3.611)--(8.371,3.598)--(8.632,3.631)%
  --(8.893,3.661)--(9.153,3.660)--(9.414,3.695)--(9.674,3.686)--(9.935,3.722)--(10.196,3.706)%
  --(10.456,3.768)--(10.717,3.755)--(10.977,3.754)--(11.238,3.754)--(11.499,3.727)--(11.759,3.771)%
  --(12.020,3.788)--(12.280,3.803)--(12.541,3.816)--(12.802,3.797)--(13.047,3.821);
\gppoint{gp mark 13}{(11.499,3.727)}
\gppoint{gp mark 13}{(12.221,1.319)}
\gpcolor{color=gp lt color border}
\gpsetlinetype{gp lt border}
\gpsetlinewidth{1.00}
\draw[gp path] (1.320,4.631)--(1.320,0.985)--(13.047,0.985)--(13.047,4.631)--cycle;
\gpdefrectangularnode{gp plot 1}{\pgfpoint{1.320cm}{0.985cm}}{\pgfpoint{13.047cm}{4.631cm}}
\end{tikzpicture}

%% file: plots/1GPU-czeng1-warmup-steps.tex
\begin{tikzpicture}[gnuplot]
\path (0.000,0.000) rectangle (13.600,5.000);
\gpcolor{color=gp lt color border}
\gpsetlinetype{gp lt border}
\gpsetlinewidth{1.00}
\draw[gp path] (1.320,0.985)--(1.500,0.985);
\draw[gp path] (13.047,0.985)--(12.867,0.985);
\node[gp node right] at (1.136,0.985) { 0};
\draw[gp path] (1.320,1.729)--(1.500,1.729);
\draw[gp path] (13.047,1.729)--(12.867,1.729);
\node[gp node right] at (1.136,1.729) { 5};
\draw[gp path] (1.320,2.473)--(1.500,2.473);
\draw[gp path] (13.047,2.473)--(12.867,2.473);
\node[gp node right] at (1.136,2.473) { 10};
\draw[gp path] (1.320,3.217)--(1.500,3.217);
\draw[gp path] (13.047,3.217)--(12.867,3.217);
\node[gp node right] at (1.136,3.217) { 15};
\draw[gp path] (1.320,3.961)--(1.500,3.961);
\draw[gp path] (13.047,3.961)--(12.867,3.961);
\node[gp node right] at (1.136,3.961) { 20};
\draw[gp path] (1.320,0.985)--(1.320,1.165);
\draw[gp path] (1.320,4.631)--(1.320,4.451);
\node[gp node center] at (1.320,0.677) { 0};
\draw[gp path] (2.623,0.985)--(2.623,1.165);
\draw[gp path] (2.623,4.631)--(2.623,4.451);
\node[gp node center] at (2.623,0.677) { 5};
\draw[gp path] (3.926,0.985)--(3.926,1.165);
\draw[gp path] (3.926,4.631)--(3.926,4.451);
\node[gp node center] at (3.926,0.677) { 10};
\draw[gp path] (5.229,0.985)--(5.229,1.165);
\draw[gp path] (5.229,4.631)--(5.229,4.451);
\node[gp node center] at (5.229,0.677) { 15};
\draw[gp path] (6.532,0.985)--(6.532,1.165);
\draw[gp path] (6.532,4.631)--(6.532,4.451);
\node[gp node center] at (6.532,0.677) { 20};
\draw[gp path] (7.835,0.985)--(7.835,1.165);
\draw[gp path] (7.835,4.631)--(7.835,4.451);
\node[gp node center] at (7.835,0.677) { 25};
\draw[gp path] (9.138,0.985)--(9.138,1.165);
\draw[gp path] (9.138,4.631)--(9.138,4.451);
\node[gp node center] at (9.138,0.677) { 30};
\draw[gp path] (10.441,0.985)--(10.441,1.165);
\draw[gp path] (10.441,4.631)--(10.441,4.451);
\node[gp node center] at (10.441,0.677) { 35};
\draw[gp path] (11.744,0.985)--(11.744,1.165);
\draw[gp path] (11.744,4.631)--(11.744,4.451);
\node[gp node center] at (11.744,0.677) { 40};
\draw[gp path] (13.047,0.985)--(13.047,1.165);
\draw[gp path] (13.047,4.631)--(13.047,4.451);
\node[gp node center] at (13.047,0.677) { 45};
\draw[gp path] (1.320,4.631)--(1.320,0.985)--(13.047,0.985)--(13.047,4.631)--cycle;
\node[gp node center,rotate=-270] at (0.246,2.808) {BLEU};
\node[gp node center] at (7.183,0.215) {Training time (hours)};
\node[gp node right,font={\fontsize{8pt}{9.6pt}\selectfont}] at (11.579,2.551) {warmup steps 12k};
\gpcolor{rgb color={0.894,0.102,0.110}}
\gpsetlinetype{gp lt plot 0}
\gpsetlinewidth{2.00}
\draw[gp path] (11.763,2.551)--(12.679,2.551);
\draw[gp path] (1.326,0.985)--(1.586,1.810)--(1.847,1.375)--(2.107,1.238)--(2.368,1.240)%
  --(2.628,1.261)--(2.889,1.235)--(3.150,1.249)--(3.410,1.237)--(3.671,1.285)--(3.932,1.260);
\gpsetpointsize{8.00}
\gppoint{gp mark 5}{(3.410,1.237)}
\gppoint{gp mark 5}{(12.221,2.551)}
\gpcolor{color=gp lt color border}
\node[gp node right,font={\fontsize{8pt}{9.6pt}\selectfont}] at (11.579,2.243) {warmup steps 14k};
\gpcolor{rgb color={0.216,0.494,0.722}}
\draw[gp path] (11.763,2.243)--(12.679,2.243);
\draw[gp path] (1.326,0.985)--(1.586,2.038)--(1.847,2.716)--(2.107,3.168)--(2.368,3.377)%
  --(2.629,3.555)--(2.889,3.684)--(3.150,3.704)--(3.410,3.839)--(3.671,3.861)--(3.932,3.903)%
  --(4.192,4.018)--(4.453,3.970)--(4.713,4.029)--(4.974,4.051)--(5.235,4.099)--(5.495,4.140);
\gppoint{gp mark 7}{(4.974,4.051)}
\gppoint{gp mark 7}{(12.221,2.243)}
\gpcolor{color=gp lt color border}
\node[gp node right,font={\fontsize{8pt}{9.6pt}\selectfont}] at (11.579,1.935) {warmup steps 16k};
\gpcolor{rgb color={0.302,0.686,0.290}}
\draw[gp path] (11.763,1.935)--(12.679,1.935);
\draw[gp path] (1.335,0.985)--(1.596,2.063)--(1.856,2.714)--(2.117,3.176)--(2.378,3.401)%
  --(2.638,3.557)--(2.899,3.688)--(3.159,3.737)--(3.420,3.868)--(3.681,3.929)--(3.941,3.958)%
  --(4.202,4.019)--(4.462,4.040)--(4.723,4.062)--(4.984,4.081)--(5.244,4.163)--(5.505,4.180)%
  --(5.765,4.193)--(6.026,4.202)--(6.287,4.210)--(6.547,4.247)--(6.808,4.232)--(7.068,4.260)%
  --(7.329,4.261)--(7.590,4.303)--(7.850,4.291)--(8.111,4.294)--(8.371,4.304)--(8.632,4.327)%
  --(8.893,4.333)--(9.153,4.390)--(9.414,4.345)--(9.674,4.428)--(9.935,4.361)--(10.196,4.428)%
  --(10.456,4.395)--(10.717,4.375)--(10.977,4.383)--(11.238,4.400)--(11.499,4.423)--(11.759,4.382)%
  --(12.020,4.418)--(12.280,4.435)--(12.541,4.412)--(12.802,4.446)--(13.047,4.453);
\gppoint{gp mark 9}{(10.977,4.383)}
\gppoint{gp mark 9}{(12.221,1.935)}
\gpcolor{color=gp lt color border}
\node[gp node right,font={\fontsize{8pt}{9.6pt}\selectfont}] at (11.579,1.627) {warmup steps 32k};
\gpcolor{rgb color={0.596,0.306,0.639}}
\draw[gp path] (11.763,1.627)--(12.679,1.627);
\draw[gp path] (1.320,0.985)--(1.580,1.642)--(1.841,2.595)--(2.101,2.943)--(2.362,3.197)%
  --(2.622,3.497)--(2.857,3.486)--(3.117,3.596)--(3.378,3.736)--(3.638,3.766)--(3.899,3.935)%
  --(4.159,3.971)--(4.420,3.970)--(4.681,4.089)--(4.941,4.096)--(5.202,4.105)--(5.463,4.151)%
  --(5.723,4.174)--(5.984,4.203);
\gppoint{gp mark 11}{(4.681,4.089)}
\gppoint{gp mark 11}{(12.221,1.627)}
\gpcolor{color=gp lt color border}
\node[gp node right,font={\fontsize{8pt}{9.6pt}\selectfont}] at (11.579,1.319) {warmup steps 48k};
\gpcolor{rgb color={1.000,0.498,0.000}}
\draw[gp path] (11.763,1.319)--(12.679,1.319);
\draw[gp path] (1.326,0.985)--(1.586,1.310)--(1.847,2.387)--(2.107,2.917)--(2.368,3.109)%
  --(2.629,3.234)--(2.889,3.481)--(3.150,3.598)--(3.410,3.696)--(3.671,3.808)--(3.932,3.895)%
  --(4.192,3.981)--(4.453,4.032)--(4.713,4.050)--(4.974,4.107)--(5.235,4.130)--(5.495,4.164)%
  --(5.756,4.185)--(6.016,4.209)--(6.277,4.198)--(6.538,4.214)--(6.798,4.281)--(7.059,4.270)%
  --(7.320,4.285)--(7.580,4.332)--(7.841,4.327)--(8.101,4.270)--(8.363,4.317)--(8.623,4.333)%
  --(8.883,4.407)--(9.144,4.351)--(9.404,4.429)--(9.665,4.416)--(9.926,4.427)--(10.186,4.460)%
  --(10.447,4.461)--(10.708,4.435)--(10.968,4.432)--(11.229,4.411)--(11.489,4.416)--(11.750,4.429)%
  --(12.011,4.462)--(12.271,4.425)--(12.532,4.397)--(12.792,4.482)--(13.047,4.510);
\gppoint{gp mark 13}{(9.926,4.427)}
\gppoint{gp mark 13}{(12.221,1.319)}
\gpcolor{color=gp lt color border}
\gpsetlinetype{gp lt border}
\gpsetlinewidth{1.00}
\draw[gp path] (1.320,4.631)--(1.320,0.985)--(13.047,0.985)--(13.047,4.631)--cycle;
\gpdefrectangularnode{gp plot 1}{\pgfpoint{1.320cm}{0.985cm}}{\pgfpoint{13.047cm}{4.631cm}}
\end{tikzpicture}

%% file: plots/1268GPU-czeng57-b15.tex
\begin{tikzpicture}[gnuplot]
\path (0.000,0.000) rectangle (13.600,6.500);
\gpcolor{color=gp lt color border}
\gpsetlinetype{gp lt border}
\gpsetlinewidth{1.00}
\draw[gp path] (1.688,1.109)--(1.868,1.109);
\draw[gp path] (13.047,1.109)--(12.867,1.109);
\node[gp node right] at (1.504,1.109) { 23.5};
\draw[gp path] (1.688,1.729)--(1.868,1.729);
\draw[gp path] (13.047,1.729)--(12.867,1.729);
\node[gp node right] at (1.504,1.729) { 24};
\draw[gp path] (1.688,2.350)--(1.868,2.350);
\draw[gp path] (13.047,2.350)--(12.867,2.350);
\node[gp node right] at (1.504,2.350) { 24.5};
\draw[gp path] (1.688,2.970)--(1.868,2.970);
\draw[gp path] (13.047,2.970)--(12.867,2.970);
\node[gp node right] at (1.504,2.970) { 25};
\draw[gp path] (1.688,3.590)--(1.868,3.590);
\draw[gp path] (13.047,3.590)--(12.867,3.590);
\node[gp node right] at (1.504,3.590) { 25.5};
\draw[gp path] (1.688,4.210)--(1.868,4.210);
\draw[gp path] (13.047,4.210)--(12.867,4.210);
\node[gp node right] at (1.504,4.210) { 26};
\draw[gp path] (1.688,4.831)--(1.868,4.831);
\draw[gp path] (13.047,4.831)--(12.867,4.831);
\node[gp node right] at (1.504,4.831) { 26.5};
\draw[gp path] (1.688,5.451)--(1.868,5.451);
\draw[gp path] (13.047,5.451)--(12.867,5.451);
\node[gp node right] at (1.504,5.451) { 27};
\draw[gp path] (1.688,0.985)--(1.688,1.165);
\node[gp node center] at (1.688,0.677) { 0};
\draw[gp path] (3.311,0.985)--(3.311,1.165);
\node[gp node center] at (3.311,0.677) { 50};
\draw[gp path] (4.933,0.985)--(4.933,1.165);
\node[gp node center] at (4.933,0.677) { 100};
\draw[gp path] (6.556,0.985)--(6.556,1.165);
\node[gp node center] at (6.556,0.677) { 150};
\draw[gp path] (8.179,0.985)--(8.179,1.165);
\node[gp node center] at (8.179,0.677) { 200};
\draw[gp path] (9.802,0.985)--(9.802,1.165);
\node[gp node center] at (9.802,0.677) { 250};
\draw[gp path] (11.424,0.985)--(11.424,1.165);
\node[gp node center] at (11.424,0.677) { 300};
\draw[gp path] (13.047,0.985)--(13.047,1.165);
\node[gp node center] at (13.047,0.677) { 350};
\draw[gp path] (2.467,5.575)--(2.467,5.395);
\node[gp node center] at (2.467,5.883) { 1};
\draw[gp path] (3.246,5.575)--(3.246,5.395);
\node[gp node center] at (3.246,5.883) { 2};
\draw[gp path] (4.025,5.575)--(4.025,5.395);
\node[gp node center] at (4.025,5.883) { 3};
\draw[gp path] (4.804,5.575)--(4.804,5.395);
\node[gp node center] at (4.804,5.883) { 4};
\draw[gp path] (5.583,5.575)--(5.583,5.395);
\node[gp node center] at (5.583,5.883) { 5};
\draw[gp path] (6.362,5.575)--(6.362,5.395);
\node[gp node center] at (6.362,5.883) { 6};
\draw[gp path] (7.140,5.575)--(7.140,5.395);
\node[gp node center] at (7.140,5.883) { 7};
\draw[gp path] (7.919,5.575)--(7.919,5.395);
\node[gp node center] at (7.919,5.883) { 8};
\draw[gp path] (8.698,5.575)--(8.698,5.395);
\node[gp node center] at (8.698,5.883) { 9};
\draw[gp path] (9.477,5.575)--(9.477,5.395);
\node[gp node center] at (9.477,5.883) { 10};
\draw[gp path] (10.256,5.575)--(10.256,5.395);
\node[gp node center] at (10.256,5.883) { 11};
\draw[gp path] (11.035,5.575)--(11.035,5.395);
\node[gp node center] at (11.035,5.883) { 12};
\draw[gp path] (11.814,5.575)--(11.814,5.395);
\node[gp node center] at (11.814,5.883) { 13};
\draw[gp path] (12.593,5.575)--(12.593,5.395);
\node[gp node center] at (12.593,5.883) { 14};
\draw[gp path] (1.688,5.575)--(1.688,0.985)--(13.047,0.985)--(13.047,5.575)--cycle;
\node[gp node center,rotate=-270] at (0.246,3.280) {BLEU};
\node[gp node center] at (7.367,0.215) {Training time (hours)};
\node[gp node center] at (7.367,6.344) {Training time (days)};
\node[gp node right,font={\fontsize{8pt}{9.6pt}\selectfont}] at (11.579,2.551) {8GPU};
\gpcolor{rgb color={0.894,0.102,0.110}}
\gpsetlinetype{gp lt plot 0}
\gpsetlinewidth{2.00}
\draw[gp path] (11.763,2.551)--(12.679,2.551);
\draw[gp path] (1.990,0.985)--(2.014,1.323)--(2.047,1.530)--(2.079,1.799)--(2.112,1.625)%
  --(2.144,1.796)--(2.177,2.132)--(2.209,2.003)--(2.242,2.149)--(2.274,2.781)--(2.307,2.880)%
  --(2.339,2.746)--(2.371,2.717)--(2.404,3.140)--(2.436,2.820)--(2.469,3.340)--(2.501,3.166)%
  --(2.534,3.143)--(2.566,3.031)--(2.599,3.264)--(2.631,3.252)--(2.664,3.458)--(2.696,3.063)%
  --(2.728,3.430)--(2.761,3.316)--(2.793,3.639)--(2.826,3.672)--(2.858,3.686)--(2.891,3.605)%
  --(2.923,3.702)--(2.956,3.543)--(2.988,3.790)--(3.021,3.857)--(3.053,3.954)--(3.085,4.100)%
  --(3.118,3.975)--(3.150,3.860)--(3.183,3.648)--(3.215,4.141)--(3.248,3.887)--(3.280,4.051)%
  --(3.313,3.946)--(3.345,4.080)--(3.378,3.615)--(3.410,4.133)--(3.442,4.119)--(3.475,4.176)%
  --(3.507,4.121)--(3.540,4.383)--(3.572,3.841)--(3.605,4.037)--(3.637,4.089)--(3.670,4.362)%
  --(3.702,4.117)--(3.735,4.435)--(3.767,4.109)--(3.799,3.950)--(3.832,3.897)--(3.864,4.393)%
  --(3.897,4.147)--(3.929,4.359)--(3.962,4.192)--(3.994,4.157)--(4.027,3.919)--(4.059,4.087)%
  --(4.092,4.126)--(4.124,4.281)--(4.156,4.227)--(4.189,4.054)--(4.221,4.491)--(4.254,4.462)%
  --(4.286,4.308)--(4.319,4.453)--(4.351,4.337)--(4.384,4.491)--(4.416,4.366)--(4.449,4.641)%
  --(4.481,4.531)--(4.513,4.668)--(4.546,4.679)--(4.578,4.668)--(4.611,4.429)--(4.643,4.463)%
  --(4.676,4.502)--(4.708,4.487)--(4.741,4.442)--(4.773,4.515)--(4.806,4.571)--(4.838,4.517)%
  --(4.870,4.565)--(4.903,4.421)--(4.935,4.624)--(4.968,4.734)--(5.000,4.546)--(5.033,4.637)%
  --(5.065,4.677)--(5.098,4.491)--(5.130,4.488)--(5.163,4.464)--(5.195,4.549)--(5.227,4.367)%
  --(5.260,4.394)--(5.292,4.629)--(5.325,4.326)--(5.357,4.627)--(5.390,4.440)--(5.422,4.680)%
  --(5.455,4.727)--(5.487,4.574)--(5.519,4.639)--(5.552,4.632)--(5.584,4.562)--(5.617,4.543)%
  --(5.649,4.408)--(5.682,4.539)--(5.714,4.489)--(5.747,4.658)--(5.779,4.543)--(5.812,4.602)%
  --(5.844,4.629)--(5.876,4.241)--(5.909,4.826)--(5.941,4.702)--(5.974,4.772)--(6.006,4.747)%
  --(6.039,4.669)--(6.071,4.844)--(6.104,4.796)--(6.136,4.811)--(6.169,4.817)--(6.201,4.804)%
  --(6.233,4.579)--(6.266,4.612)--(6.298,4.690)--(6.331,4.708)--(6.363,4.748)--(6.396,4.593)%
  --(6.428,4.721)--(6.461,4.744)--(6.493,4.838)--(6.526,4.700)--(6.558,4.640)--(6.590,4.757)%
  --(6.623,4.922)--(6.655,4.659)--(6.688,4.862)--(6.720,5.028)--(6.753,4.756)--(6.785,4.874)%
  --(6.818,5.053)--(6.850,4.696)--(6.883,4.476)--(6.915,4.533)--(6.947,4.720)--(6.980,4.571)%
  --(7.012,4.757)--(7.045,4.801)--(7.077,4.874)--(7.110,4.766)--(7.142,4.683)--(7.175,4.743)%
  --(7.207,4.830)--(7.240,4.945)--(7.272,4.790)--(7.304,4.879)--(7.337,5.009)--(7.369,4.958)%
  --(7.402,4.996)--(7.434,4.888)--(7.467,4.695)--(7.499,4.614)--(7.532,4.836)--(7.564,4.754)%
  --(7.597,4.942)--(7.629,4.682)--(7.661,4.733)--(7.694,5.060)--(7.726,4.858)--(7.759,4.690)%
  --(7.791,4.996)--(7.824,4.892)--(7.856,5.035)--(7.889,4.794)--(7.921,4.894)--(7.954,4.923)%
  --(7.986,4.941)--(8.018,5.107)--(8.051,5.004)--(8.083,5.157)--(8.116,4.762)--(8.148,4.882)%
  --(8.181,4.884)--(8.213,5.016)--(8.246,4.971)--(8.278,4.922)--(8.311,4.933)--(8.343,4.846)%
  --(8.375,4.769)--(8.408,4.921)--(8.440,5.083)--(8.473,4.886)--(8.505,5.089)--(8.538,4.897)%
  --(8.570,4.947)--(8.603,5.012)--(8.635,4.937)--(8.668,4.865)--(8.700,4.926)--(8.732,5.095)%
  --(8.765,5.049)--(8.797,5.071)--(8.830,5.165)--(8.862,5.048)--(8.895,5.179)--(8.927,5.067)%
  --(8.960,5.305)--(8.992,5.181)--(9.025,5.046)--(9.057,5.088)--(9.089,5.068)--(9.122,5.319)%
  --(9.154,5.020)--(9.187,5.003)--(9.219,4.980)--(9.252,5.015)--(9.284,4.894)--(9.317,5.083)%
  --(9.349,5.020)--(9.382,4.876)--(9.414,5.229)--(9.446,4.986)--(9.479,5.033)--(9.511,5.063)%
  --(9.544,4.949)--(9.576,5.027)--(9.609,4.723)--(9.641,5.160)--(9.674,5.248)--(9.706,5.049)%
  --(9.739,5.170)--(9.771,5.196)--(9.803,5.117)--(9.836,5.017)--(9.868,5.044)--(9.901,4.905)%
  --(9.933,5.125)--(9.966,4.850)--(9.998,4.935)--(10.031,5.237)--(10.063,5.159)--(10.096,5.231)%
  --(10.128,5.168)--(10.160,5.041)--(10.193,5.076)--(10.225,5.033)--(10.258,5.222)--(10.290,5.309)%
  --(10.323,5.115)--(10.355,5.142)--(10.388,5.457)--(10.420,5.210)--(10.453,5.053)--(10.485,5.033)%
  --(10.517,5.032)--(10.550,5.047)--(10.582,4.843)--(10.615,5.017)--(10.647,5.087)--(10.680,5.038)%
  --(10.712,5.086)--(10.745,5.199)--(10.777,4.984)--(10.810,5.098)--(10.842,4.986)--(10.874,5.232)%
  --(10.907,5.130)--(10.939,5.112)--(10.972,5.211)--(11.004,4.863)--(11.037,5.093)--(11.069,4.921)%
  --(11.102,4.775)--(11.134,4.956)--(11.167,4.850)--(11.199,5.006)--(11.231,4.790)--(11.264,4.969)%
  --(11.296,4.880)--(11.329,4.701)--(11.361,5.036)--(11.394,4.888)--(11.426,4.851)--(11.459,4.989)%
  --(11.491,4.937)--(11.524,5.080)--(11.556,5.091)--(11.588,5.018)--(11.621,5.003)--(11.653,4.847)%
  --(11.686,4.673)--(11.718,4.919)--(11.751,4.888)--(11.783,4.899)--(11.816,4.938)--(11.848,4.805)%
  --(11.881,5.097)--(11.913,5.208)--(11.945,4.764)--(11.978,4.968)--(12.010,5.024)--(12.043,4.941)%
  --(12.075,5.068)--(12.108,5.082)--(12.140,5.033)--(12.173,4.990)--(12.205,5.125)--(12.238,4.970)%
  --(12.270,5.005)--(12.302,4.821)--(12.335,4.918)--(12.367,4.826)--(12.400,4.874)--(12.432,4.936)%
  --(12.465,4.855)--(12.497,4.951)--(12.530,5.045)--(12.562,4.987)--(12.595,4.954)--(12.627,4.951)%
  --(12.659,4.985)--(12.692,5.283)--(12.724,4.704)--(12.757,5.381)--(12.789,5.130)--(12.822,5.250)%
  --(12.854,5.120)--(12.887,4.985)--(12.919,4.861)--(12.952,4.798)--(12.984,5.204)--(13.016,4.952)%
  --(13.047,5.136);
\gpsetpointsize{8.00}
\gppoint{gp mark 5}{(11.881,5.097)}
\gppoint{gp mark 5}{(12.221,2.551)}
\gpcolor{color=gp lt color border}
\node[gp node right,font={\fontsize{8pt}{9.6pt}\selectfont}] at (11.579,2.243) {6GPU};
\gpcolor{rgb color={0.216,0.494,0.722}}
\gpsetlinetype{gp lt plot 3}
\draw[gp path] (11.763,2.243)--(12.679,2.243);
\draw[gp path] (2.051,0.985)--(2.079,1.365)--(2.112,1.673)--(2.144,1.496)--(2.177,1.763)%
  --(2.209,2.119)--(2.242,2.031)--(2.274,2.066)--(2.307,1.699)--(2.339,2.340)--(2.371,2.400)%
  --(2.404,2.555)--(2.436,2.547)--(2.469,2.830)--(2.501,2.764)--(2.534,2.999)--(2.566,3.021)%
  --(2.599,2.958)--(2.631,3.048)--(2.664,3.564)--(2.696,3.142)--(2.728,3.428)--(2.761,3.255)%
  --(2.793,3.317)--(2.826,3.613)--(2.858,3.432)--(2.891,3.526)--(2.923,3.583)--(2.956,3.731)%
  --(2.988,3.524)--(3.021,3.594)--(3.053,3.508)--(3.085,3.746)--(3.118,3.869)--(3.150,3.730)%
  --(3.183,3.917)--(3.215,3.803)--(3.248,3.824)--(3.280,3.570)--(3.313,3.736)--(3.345,3.986)%
  --(3.378,3.822)--(3.410,3.980)--(3.442,3.542)--(3.475,3.931)--(3.507,3.910)--(3.540,3.879)%
  --(3.572,3.897)--(3.605,3.732)--(3.637,4.171)--(3.670,4.297)--(3.702,3.976)--(3.735,4.130)%
  --(3.767,4.183)--(3.799,4.077)--(3.832,4.012)--(3.864,4.156)--(3.897,4.090)--(3.929,4.125)%
  --(3.962,4.092)--(3.994,4.158)--(4.027,3.871)--(4.059,3.967)--(4.092,4.210)--(4.124,4.047)%
  --(4.156,3.872)--(4.189,4.127)--(4.221,4.279)--(4.254,3.998)--(4.286,4.195)--(4.319,4.438)%
  --(4.351,4.440)--(4.384,4.080)--(4.416,4.047)--(4.449,4.237)--(4.481,4.360)--(4.513,4.318)%
  --(4.546,4.305)--(4.578,4.307)--(4.611,4.201)--(4.643,4.219)--(4.676,4.239)--(4.708,4.316)%
  --(4.741,4.364)--(4.773,3.957)--(4.806,4.360)--(4.838,4.657)--(4.870,4.513)--(4.903,4.448)%
  --(4.935,4.806)--(4.968,4.439)--(5.000,4.615)--(5.033,4.402)--(5.065,4.559)--(5.098,4.503)%
  --(5.130,4.433)--(5.163,4.479)--(5.195,4.637)--(5.227,4.553)--(5.260,4.641)--(5.292,4.297)%
  --(5.325,4.281)--(5.357,4.480)--(5.390,4.412)--(5.422,4.422)--(5.455,4.699)--(5.487,4.630)%
  --(5.519,4.563)--(5.552,4.389)--(5.584,4.557)--(5.617,4.467)--(5.649,4.322)--(5.682,4.679)%
  --(5.714,4.616)--(5.747,4.340)--(5.779,4.459)--(5.812,4.504)--(5.844,4.576)--(5.876,4.575)%
  --(5.909,4.481)--(5.941,4.542)--(5.974,4.230)--(6.006,4.195)--(6.039,4.829)--(6.071,4.657)%
  --(6.104,4.332)--(6.136,4.665)--(6.169,4.518)--(6.201,4.445)--(6.233,4.542)--(6.266,4.451)%
  --(6.298,4.771)--(6.331,4.327)--(6.363,4.379)--(6.396,4.447)--(6.428,4.517)--(6.461,4.485)%
  --(6.493,4.581)--(6.526,4.684)--(6.558,4.399)--(6.590,4.752)--(6.623,4.479)--(6.655,4.601)%
  --(6.688,4.707)--(6.720,4.532)--(6.753,4.714)--(6.785,4.661)--(6.818,4.574)--(6.850,4.624)%
  --(6.883,4.573)--(6.915,4.470)--(6.947,4.606)--(6.980,4.683)--(7.012,4.527)--(7.045,4.536)%
  --(7.077,4.668)--(7.110,4.530)--(7.142,4.403)--(7.175,4.459)--(7.207,4.607)--(7.240,4.665)%
  --(7.272,4.706)--(7.304,4.541)--(7.337,4.776)--(7.369,4.644)--(7.402,4.735)--(7.434,4.898)%
  --(7.467,4.919)--(7.499,4.599)--(7.532,4.698)--(7.564,4.718)--(7.597,4.937)--(7.629,4.696);
\gppoint{gp mark 7}{(7.175,4.459)}
\gppoint{gp mark 7}{(12.221,2.243)}
\gpcolor{color=gp lt color border}
\node[gp node right,font={\fontsize{8pt}{9.6pt}\selectfont}] at (11.579,1.935) {2GPU};
\gpcolor{rgb color={0.302,0.686,0.290}}
\gpsetlinetype{gp lt plot 0}
\draw[gp path] (11.763,1.935)--(12.679,1.935);
\draw[gp path] (2.490,0.985)--(2.501,1.049)--(2.513,0.985);
\draw[gp path] (2.595,0.985)--(2.599,1.015)--(2.631,1.429)--(2.661,0.985);
\draw[gp path] (2.675,0.985)--(2.696,1.064)--(2.728,1.060)--(2.761,1.669)--(2.793,1.285)%
  --(2.826,1.253)--(2.858,1.370)--(2.891,1.666)--(2.923,2.008)--(2.956,1.637)--(2.988,1.691)%
  --(3.021,1.679)--(3.053,1.443)--(3.085,1.946)--(3.118,1.857)--(3.150,2.047)--(3.183,1.895)%
  --(3.215,2.301)--(3.248,2.401)--(3.280,2.422)--(3.313,2.273)--(3.345,2.719)--(3.378,2.410)%
  --(3.410,2.306)--(3.442,2.341)--(3.475,2.234)--(3.507,2.668)--(3.540,2.033)--(3.572,2.256)%
  --(3.605,2.390)--(3.637,2.330)--(3.670,2.546)--(3.702,2.282)--(3.735,2.782)--(3.767,2.453)%
  --(3.799,2.736)--(3.832,2.579)--(3.864,2.563)--(3.897,2.724)--(3.929,2.574)--(3.962,2.636)%
  --(3.994,2.614)--(4.027,3.167)--(4.059,2.947)--(4.092,2.813)--(4.124,2.748)--(4.156,2.664)%
  --(4.189,2.739)--(4.221,2.827)--(4.254,2.978)--(4.286,3.069)--(4.319,2.623)--(4.351,3.027)%
  --(4.384,3.094)--(4.416,3.276)--(4.449,3.051)--(4.481,2.881)--(4.513,2.966)--(4.546,3.030)%
  --(4.578,2.836)--(4.611,3.053)--(4.643,2.774)--(4.676,2.893)--(4.708,3.169)--(4.741,3.122)%
  --(4.773,3.049)--(4.806,3.170)--(4.838,3.049)--(4.870,3.401)--(4.903,3.362)--(4.935,3.322)%
  --(4.968,2.801)--(5.000,2.928)--(5.033,3.295)--(5.065,3.277)--(5.098,3.474)--(5.130,3.238)%
  --(5.163,3.478)--(5.195,3.258)--(5.227,3.407)--(5.260,3.459)--(5.292,3.081)--(5.325,3.288)%
  --(5.357,3.325)--(5.390,3.454)--(5.422,3.592)--(5.455,3.217)--(5.487,3.344)--(5.519,3.318)%
  --(5.552,3.384)--(5.584,3.205)--(5.617,3.476)--(5.649,3.296)--(5.682,3.528)--(5.714,3.490)%
  --(5.747,3.394)--(5.779,3.574)--(5.812,3.528)--(5.844,3.673)--(5.876,3.643)--(5.909,3.292)%
  --(5.941,3.293)--(5.974,3.494)--(6.006,3.676)--(6.039,3.517)--(6.071,3.322)--(6.104,3.566)%
  --(6.136,3.380)--(6.169,3.683)--(6.201,3.541)--(6.233,3.454)--(6.266,3.308)--(6.298,3.666)%
  --(6.331,3.695)--(6.363,3.374)--(6.396,3.512)--(6.428,3.354)--(6.461,3.601)--(6.493,3.668)%
  --(6.526,3.388)--(6.558,3.626)--(6.590,3.886)--(6.623,3.776)--(6.655,3.931)--(6.688,3.786)%
  --(6.720,3.721)--(6.753,3.893)--(6.785,3.912)--(6.818,3.736)--(6.850,3.690)--(6.883,3.678)%
  --(6.915,3.695)--(6.947,3.864)--(6.980,3.994)--(7.012,3.877)--(7.045,3.795)--(7.077,3.815)%
  --(7.110,3.664)--(7.142,3.721)--(7.175,3.961)--(7.207,3.825)--(7.240,3.858)--(7.272,3.925)%
  --(7.304,3.922)--(7.337,3.711)--(7.369,3.719)--(7.402,3.630)--(7.434,3.805)--(7.467,3.694)%
  --(7.499,3.981)--(7.532,4.291)--(7.564,3.820)--(7.597,3.565)--(7.629,3.852)--(7.661,3.711)%
  --(7.694,4.007)--(7.726,3.805)--(7.759,4.002)--(7.791,4.115)--(7.824,3.827)--(7.856,3.765)%
  --(7.889,3.800)--(7.921,3.715)--(7.954,4.065)--(7.986,3.628)--(8.018,3.655)--(8.051,3.783)%
  --(8.083,3.791)--(8.116,3.846)--(8.148,3.862)--(8.181,3.805)--(8.213,3.937)--(8.246,3.663)%
  --(8.278,3.854)--(8.311,3.768)--(8.343,3.758)--(8.375,3.777)--(8.408,4.115)--(8.440,4.110)%
  --(8.473,4.015)--(8.505,3.938)--(8.538,3.930)--(8.570,3.923)--(8.603,4.092)--(8.635,4.310)%
  --(8.668,4.206)--(8.700,4.218)--(8.732,3.923)--(8.765,4.010)--(8.797,4.332)--(8.830,4.099)%
  --(8.862,4.039)--(8.895,4.247)--(8.927,4.064)--(8.960,4.168)--(8.992,4.155)--(9.025,3.933)%
  --(9.057,4.299)--(9.089,4.169)--(9.122,4.221)--(9.154,4.121)--(9.187,4.194)--(9.219,3.999)%
  --(9.252,4.041)--(9.284,4.225)--(9.317,3.956)--(9.349,3.925)--(9.382,4.298)--(9.414,3.967)%
  --(9.446,4.374)--(9.479,3.931)--(9.511,4.189)--(9.544,4.096)--(9.576,4.139)--(9.609,3.905)%
  --(9.641,4.159)--(9.674,4.143)--(9.706,4.072)--(9.739,4.000)--(9.771,4.154)--(9.803,4.036)%
  --(9.836,4.177)--(9.868,4.129)--(9.901,4.098)--(9.933,4.035)--(9.966,4.193)--(9.998,4.227)%
  --(10.031,4.040)--(10.063,3.933)--(10.096,3.874)--(10.128,4.103)--(10.160,3.799)--(10.193,4.223)%
  --(10.225,4.265)--(10.258,4.068)--(10.290,4.280)--(10.323,4.110)--(10.355,4.082)--(10.388,4.244)%
  --(10.420,4.169)--(10.453,4.332)--(10.485,4.218)--(10.517,4.155)--(10.550,4.208)--(10.582,4.138)%
  --(10.615,4.454)--(10.647,4.186)--(10.680,4.515)--(10.712,4.118)--(10.745,4.235)--(10.777,4.277)%
  --(10.810,4.259)--(10.842,4.215)--(10.874,4.524)--(10.907,4.284)--(10.939,4.378)--(10.972,4.056)%
  --(11.004,4.513)--(11.037,4.287)--(11.069,4.425)--(11.102,4.031)--(11.134,4.082)--(11.167,4.164)%
  --(11.199,3.989)--(11.231,4.394)--(11.264,4.010)--(11.296,4.191)--(11.329,4.378)--(11.361,4.254)%
  --(11.394,4.372)--(11.426,4.101)--(11.459,4.252)--(11.491,4.424)--(11.524,4.250)--(11.556,4.320)%
  --(11.588,4.284)--(11.621,4.057)--(11.653,4.436)--(11.686,4.221)--(11.718,4.102)--(11.751,4.293)%
  --(11.783,4.032)--(11.816,4.241)--(11.848,4.298)--(11.881,4.358)--(11.913,4.137)--(11.945,4.170)%
  --(11.978,4.172)--(12.010,4.295)--(12.043,4.158)--(12.075,4.280)--(12.108,4.306)--(12.140,4.317)%
  --(12.173,3.961)--(12.205,4.437)--(12.238,4.123)--(12.270,4.376)--(12.302,4.314)--(12.335,4.086)%
  --(12.367,4.294)--(12.400,4.289)--(12.432,4.196)--(12.465,4.405)--(12.497,4.368)--(12.530,4.596)%
  --(12.562,4.510)--(12.595,4.127)--(12.627,4.287)--(12.659,4.183)--(12.692,4.353)--(12.724,4.652)%
  --(12.757,4.454)--(12.789,4.346)--(12.822,4.457)--(12.854,4.543)--(12.887,4.380)--(12.919,4.501)%
  --(12.952,4.515)--(12.984,4.603)--(13.016,4.064)--(13.047,4.145);
\gppoint{gp mark 9}{(11.881,4.358)}
\gppoint{gp mark 9}{(12.221,1.935)}
\gpcolor{color=gp lt color border}
\node[gp node right,font={\fontsize{8pt}{9.6pt}\selectfont}] at (11.579,1.627) {1GPU};
\gpcolor{rgb color={0.596,0.306,0.639}}
\draw[gp path] (11.763,1.627)--(12.679,1.627);
\draw[gp path] (3.609,0.985)--(3.637,1.122)--(3.657,0.985);
\draw[gp path] (3.807,0.985)--(3.832,1.309)--(3.856,0.985);
\draw[gp path] (3.890,0.985)--(3.897,1.018)--(3.929,1.301)--(3.962,1.094)--(3.994,1.056)%
  --(4.027,1.503)--(4.059,1.235)--(4.092,1.376)--(4.124,1.350)--(4.156,1.674)--(4.189,1.156)%
  --(4.221,1.383)--(4.254,1.604)--(4.286,1.503)--(4.319,1.601)--(4.351,1.343)--(4.384,1.552)%
  --(4.416,1.450)--(4.449,1.518)--(4.481,1.630)--(4.513,1.642)--(4.546,1.612)--(4.578,1.837)%
  --(4.611,1.323)--(4.643,1.675)--(4.676,1.548)--(4.708,1.675)--(4.741,1.947)--(4.773,2.040)%
  --(4.806,1.994)--(4.838,1.657)--(4.870,1.389)--(4.903,1.879)--(4.935,1.746)--(4.968,1.518)%
  --(5.000,1.750)--(5.033,1.652)--(5.065,1.921)--(5.098,1.659)--(5.130,1.723)--(5.163,1.905)%
  --(5.195,1.955)--(5.227,1.969)--(5.260,1.630)--(5.292,1.966)--(5.325,1.999)--(5.357,2.037)%
  --(5.390,2.144)--(5.422,2.221)--(5.455,2.023)--(5.487,1.835)--(5.519,1.995)--(5.552,2.020)%
  --(5.584,1.776)--(5.617,1.858)--(5.649,2.039)--(5.682,2.204)--(5.714,2.510)--(5.747,2.038)%
  --(5.779,2.311)--(5.812,2.238)--(5.844,2.292)--(5.876,2.348)--(5.909,2.088)--(5.941,2.261)%
  --(5.974,2.201)--(6.006,2.250)--(6.039,2.250)--(6.071,2.274)--(6.104,2.350)--(6.136,2.006)%
  --(6.169,2.270)--(6.201,2.123)--(6.233,2.004)--(6.266,2.122)--(6.298,2.276)--(6.331,2.174)%
  --(6.363,2.260)--(6.396,2.480)--(6.428,2.315)--(6.461,2.686)--(6.493,2.560)--(6.526,2.542)%
  --(6.558,2.595)--(6.590,2.433)--(6.623,2.197)--(6.655,2.204)--(6.688,2.272)--(6.720,2.140)%
  --(6.753,2.350)--(6.785,2.523)--(6.818,2.521)--(6.850,2.545)--(6.883,2.258)--(6.915,2.375)%
  --(6.947,2.488)--(6.980,2.510)--(7.012,2.750)--(7.045,2.293)--(7.077,2.592)--(7.110,2.477)%
  --(7.142,2.266)--(7.175,2.767)--(7.207,2.443)--(7.240,2.562)--(7.272,2.638)--(7.304,2.875)%
  --(7.337,2.507)--(7.369,2.579)--(7.402,2.788)--(7.434,2.629)--(7.467,2.466)--(7.499,2.637)%
  --(7.532,2.587)--(7.564,2.584)--(7.597,2.505)--(7.629,2.340)--(7.661,2.458)--(7.694,2.390)%
  --(7.726,2.447)--(7.759,2.922)--(7.791,2.653)--(7.824,2.632)--(7.856,3.126)--(7.889,2.627)%
  --(7.921,3.031)--(7.954,2.759)--(7.986,3.090)--(8.018,2.867)--(8.051,2.740)--(8.083,2.901)%
  --(8.116,2.807)--(8.148,3.240)--(8.181,2.809)--(8.213,2.761)--(8.246,2.805)--(8.278,2.526)%
  --(8.311,2.663)--(8.343,2.353)--(8.375,2.706)--(8.408,2.344)--(8.440,2.725)--(8.473,2.785)%
  --(8.505,2.592)--(8.538,2.731)--(8.570,2.588)--(8.603,2.665)--(8.635,2.859)--(8.668,2.915)%
  --(8.700,2.870)--(8.732,2.560)--(8.765,2.625)--(8.797,2.929)--(8.830,2.451)--(8.862,2.952)%
  --(8.895,2.899)--(8.927,2.946)--(8.960,2.771)--(8.992,2.751)--(9.025,2.711)--(9.057,2.656)%
  --(9.089,2.805)--(9.122,2.913)--(9.154,2.967)--(9.187,2.878)--(9.219,2.827)--(9.252,2.834)%
  --(9.284,2.932)--(9.317,3.073)--(9.349,3.337)--(9.382,2.977)--(9.414,2.873)--(9.446,2.899)%
  --(9.479,2.668)--(9.511,2.896)--(9.544,2.899)--(9.576,2.994)--(9.609,2.903)--(9.641,2.822)%
  --(9.674,2.733)--(9.706,3.137)--(9.739,2.894)--(9.771,2.925)--(9.803,2.779)--(9.836,2.683)%
  --(9.868,3.069)--(9.901,3.119)--(9.933,3.185)--(9.966,2.762)--(9.998,2.902)--(10.031,3.044)%
  --(10.063,2.751)--(10.096,2.863)--(10.128,2.898)--(10.160,2.854)--(10.193,3.005)--(10.225,2.928)%
  --(10.258,3.084)--(10.290,3.104)--(10.323,3.158)--(10.355,2.999)--(10.388,2.823)--(10.420,2.839)%
  --(10.453,2.805)--(10.485,2.797)--(10.517,3.167)--(10.550,3.167)--(10.582,3.181)--(10.615,3.110)%
  --(10.647,3.204)--(10.680,2.900)--(10.712,3.394)--(10.745,3.268)--(10.777,3.262)--(10.810,3.113)%
  --(10.842,2.936)--(10.874,3.116)--(10.907,3.152)--(10.939,3.415)--(10.972,3.201)--(11.004,3.192)%
  --(11.037,2.969)--(11.069,3.279)--(11.102,3.296)--(11.134,3.176)--(11.167,2.993)--(11.199,3.063)%
  --(11.231,3.130)--(11.264,2.728)--(11.296,2.892)--(11.329,3.267)--(11.361,3.079)--(11.394,3.062)%
  --(11.426,3.296)--(11.459,3.088)--(11.491,3.189)--(11.524,3.238)--(11.556,3.197)--(11.588,3.303)%
  --(11.621,3.242)--(11.653,3.147)--(11.686,3.356)--(11.718,3.298)--(11.751,2.957)--(11.783,3.520)%
  --(11.816,3.421)--(11.848,3.221)--(11.881,3.409)--(11.913,3.425)--(11.945,3.487)--(11.978,3.298)%
  --(12.010,3.202)--(12.043,3.061)--(12.075,3.081)--(12.108,3.285)--(12.140,3.305)--(12.173,3.305)%
  --(12.205,3.515)--(12.238,3.162)--(12.270,3.170)--(12.302,3.129)--(12.335,3.185)--(12.367,3.277)%
  --(12.400,3.351)--(12.432,3.408)--(12.465,3.006)--(12.497,3.108)--(12.530,3.070)--(12.562,3.177)%
  --(12.595,3.593)--(12.627,3.340)--(12.659,3.535)--(12.692,3.297)--(12.724,3.396)--(12.757,3.201)%
  --(12.789,3.418)--(12.822,3.315)--(12.854,3.372)--(12.887,3.345)--(12.919,3.563)--(12.952,3.688)%
  --(12.984,3.533)--(13.016,3.325)--(13.047,3.569);
\gppoint{gp mark 11}{(11.881,3.409)}
\gppoint{gp mark 11}{(12.221,1.627)}
\gpcolor{color=gp lt color border}
\node[gp node right,font={\fontsize{8pt}{9.6pt}\selectfont}] at (11.579,1.319) {25.6};
\gpcolor{rgb color={0.000,0.000,0.000}}
\gpsetlinetype{gp lt plot 4}
\gpsetlinewidth{1.00}
\draw[gp path] (11.763,1.319)--(12.679,1.319);
\draw[gp path] (1.688,3.714)--(1.803,3.714)--(1.917,3.714)--(2.032,3.714)--(2.147,3.714)%
  --(2.262,3.714)--(2.376,3.714)--(2.491,3.714)--(2.606,3.714)--(2.721,3.714)--(2.835,3.714)%
  --(2.950,3.714)--(3.065,3.714)--(3.180,3.714)--(3.294,3.714)--(3.409,3.714)--(3.524,3.714)%
  --(3.639,3.714)--(3.753,3.714)--(3.868,3.714)--(3.983,3.714)--(4.097,3.714)--(4.212,3.714)%
  --(4.327,3.714)--(4.442,3.714)--(4.556,3.714)--(4.671,3.714)--(4.786,3.714)--(4.901,3.714)%
  --(5.015,3.714)--(5.130,3.714)--(5.245,3.714)--(5.360,3.714)--(5.474,3.714)--(5.589,3.714)%
  --(5.704,3.714)--(5.819,3.714)--(5.933,3.714)--(6.048,3.714)--(6.163,3.714)--(6.277,3.714)%
  --(6.392,3.714)--(6.507,3.714)--(6.622,3.714)--(6.736,3.714)--(6.851,3.714)--(6.966,3.714)%
  --(7.081,3.714)--(7.195,3.714)--(7.310,3.714)--(7.425,3.714)--(7.540,3.714)--(7.654,3.714)%
  --(7.769,3.714)--(7.884,3.714)--(7.999,3.714)--(8.113,3.714)--(8.228,3.714)--(8.343,3.714)%
  --(8.458,3.714)--(8.572,3.714)--(8.687,3.714)--(8.802,3.714)--(8.916,3.714)--(9.031,3.714)%
  --(9.146,3.714)--(9.261,3.714)--(9.375,3.714)--(9.490,3.714)--(9.605,3.714)--(9.720,3.714)%
  --(9.834,3.714)--(9.949,3.714)--(10.064,3.714)--(10.179,3.714)--(10.293,3.714)--(10.408,3.714)%
  --(10.523,3.714)--(10.638,3.714)--(10.752,3.714)--(10.867,3.714)--(10.982,3.714)--(11.096,3.714)%
  --(11.211,3.714)--(11.326,3.714)--(11.441,3.714)--(11.555,3.714)--(11.670,3.714)--(11.785,3.714)%
  --(11.900,3.714)--(12.014,3.714)--(12.129,3.714)--(12.244,3.714)--(12.359,3.714)--(12.473,3.714)%
  --(12.588,3.714)--(12.703,3.714)--(12.818,3.714)--(12.932,3.714)--(13.047,3.714);
\gpcolor{color=gp lt color border}
\gpsetlinetype{gp lt border}
\draw[gp path] (1.688,5.575)--(1.688,0.985)--(13.047,0.985)--(13.047,5.575)--cycle;
\gpdefrectangularnode{gp plot 1}{\pgfpoint{1.688cm}{0.985cm}}{\pgfpoint{13.047cm}{5.575cm}}
\end{tikzpicture}

%% file: plots/6GPU-czeng57-averaging.tex
\begin{tikzpicture}[gnuplot]
\path (0.000,0.000) rectangle (13.600,5.000);
\gpcolor{color=gp lt color border}
\gpsetlinetype{gp lt border}
\gpsetlinewidth{1.00}
\draw[gp path] (1.688,1.390)--(1.868,1.390);
\draw[gp path] (13.047,1.390)--(12.867,1.390);
\node[gp node right] at (1.504,1.390) { 26};
\draw[gp path] (1.688,2.200)--(1.868,2.200);
\draw[gp path] (13.047,2.200)--(12.867,2.200);
\node[gp node right] at (1.504,2.200) { 26.2};
\draw[gp path] (1.688,3.011)--(1.868,3.011);
\draw[gp path] (13.047,3.011)--(12.867,3.011);
\node[gp node right] at (1.504,3.011) { 26.4};
\draw[gp path] (1.688,3.821)--(1.868,3.821);
\draw[gp path] (13.047,3.821)--(12.867,3.821);
\node[gp node right] at (1.504,3.821) { 26.6};
\draw[gp path] (1.688,4.631)--(1.868,4.631);
\draw[gp path] (13.047,4.631)--(12.867,4.631);
\node[gp node right] at (1.504,4.631) { 26.8};
\draw[gp path] (2.810,0.985)--(2.810,1.165);
\draw[gp path] (2.810,4.631)--(2.810,4.451);
\node[gp node center] at (2.810,0.677) { 110};
\draw[gp path] (4.212,0.985)--(4.212,1.165);
\draw[gp path] (4.212,4.631)--(4.212,4.451);
\node[gp node center] at (4.212,0.677) { 120};
\draw[gp path] (5.615,0.985)--(5.615,1.165);
\draw[gp path] (5.615,4.631)--(5.615,4.451);
\node[gp node center] at (5.615,0.677) { 130};
\draw[gp path] (7.017,0.985)--(7.017,1.165);
\draw[gp path] (7.017,4.631)--(7.017,4.451);
\node[gp node center] at (7.017,0.677) { 140};
\draw[gp path] (8.419,0.985)--(8.419,1.165);
\draw[gp path] (8.419,4.631)--(8.419,4.451);
\node[gp node center] at (8.419,0.677) { 150};
\draw[gp path] (9.822,0.985)--(9.822,1.165);
\draw[gp path] (9.822,4.631)--(9.822,4.451);
\node[gp node center] at (9.822,0.677) { 160};
\draw[gp path] (11.224,0.985)--(11.224,1.165);
\draw[gp path] (11.224,4.631)--(11.224,4.451);
\node[gp node center] at (11.224,0.677) { 170};
\draw[gp path] (12.626,0.985)--(12.626,1.165);
\draw[gp path] (12.626,4.631)--(12.626,4.451);
\node[gp node center] at (12.626,0.677) { 180};
\draw[gp path] (1.688,4.631)--(1.688,0.985)--(13.047,0.985)--(13.047,4.631)--cycle;
\node[gp node center,rotate=-270] at (0.246,2.808) {BLEU};
\node[gp node center] at (7.367,0.215) {Training time (hours)};
\node[gp node right,font={\fontsize{8pt}{9.6pt}\selectfont}] at (11.579,1.935) {averaging 16 checkpoints};
\gpcolor{rgb color={0.894,0.102,0.110}}
\gpsetlinetype{gp lt plot 0}
\gpsetlinewidth{2.00}
\draw[gp path] (11.763,1.935)--(12.679,1.935);
\draw[gp path] (1.688,2.974)--(1.696,2.974)--(1.836,3.156)--(1.977,3.235)--(2.117,3.479)%
  --(2.257,3.154)--(2.397,2.967)--(2.538,2.928)--(2.678,2.940)--(2.818,3.026)--(2.958,2.914)%
  --(3.099,3.083)--(3.239,3.311)--(3.379,3.007)--(3.519,2.851)--(3.659,3.120)--(3.800,2.810)%
  --(3.940,2.860)--(4.080,3.123)--(4.220,3.112)--(4.361,3.296)--(4.501,3.048)--(4.641,2.700)%
  --(4.781,2.819)--(4.922,2.738)--(5.062,2.919)--(5.202,3.360)--(5.342,3.502)--(5.483,3.508)%
  --(5.623,3.616)--(5.763,3.611)--(5.903,3.324)--(6.043,3.220)--(6.184,3.209)--(6.324,3.651)%
  --(6.464,3.348)--(6.604,3.227)--(6.745,3.229)--(6.885,3.464)--(7.025,3.204)--(7.165,3.381)%
  --(7.306,3.580)--(7.446,3.317)--(7.586,3.532)--(7.726,3.346)--(7.867,2.966)--(8.007,3.507)%
  --(8.147,3.083)--(8.287,3.036)--(8.427,3.152)--(8.568,3.157)--(8.708,2.761)--(8.848,2.515)%
  --(8.988,2.623)--(9.129,2.519)--(9.269,2.656)--(9.409,2.767)--(9.549,3.089)--(9.690,3.405)%
  --(9.830,3.298)--(9.970,3.269)--(10.110,3.149)--(10.250,2.943)--(10.391,3.002)--(10.531,3.064)%
  --(10.671,3.208)--(10.811,3.286)--(10.952,3.668)--(11.092,3.437)--(11.232,3.407)--(11.372,3.431)%
  --(11.513,3.600)--(11.653,3.822)--(11.793,3.948)--(11.933,4.061)--(12.074,3.744)--(12.214,3.591)%
  --(12.354,3.989)--(12.494,3.811)--(12.634,3.761)--(12.775,4.247)--(12.915,3.930)--(13.047,3.991);
\gpsetpointsize{8.00}
\gppoint{gp mark 5}{(11.653,3.822)}
\gppoint{gp mark 5}{(12.221,1.935)}
\gpcolor{color=gp lt color border}
\node[gp node right,font={\fontsize{8pt}{9.6pt}\selectfont}] at (11.579,1.627) {averaging 8 checkpoints};
\gpcolor{rgb color={0.216,0.494,0.722}}
\gpsetlinetype{gp lt plot 3}
\draw[gp path] (11.763,1.627)--(12.679,1.627);
\draw[gp path] (1.688,3.303)--(1.696,3.305)--(1.836,3.144)--(1.977,3.364)--(2.117,3.374)%
  --(2.257,3.061)--(2.397,3.062)--(2.538,2.807)--(2.678,2.998)--(2.818,3.061)--(2.958,2.952)%
  --(3.099,2.858)--(3.239,2.722)--(3.379,2.685)--(3.519,3.007)--(3.659,3.002)--(3.800,3.374)%
  --(3.940,2.729)--(4.080,2.872)--(4.220,3.000)--(4.361,2.643)--(4.501,2.982)--(4.641,3.353)%
  --(4.781,2.799)--(4.922,3.063)--(5.062,2.971)--(5.202,2.852)--(5.342,3.212)--(5.483,3.197)%
  --(5.623,3.099)--(5.763,3.352)--(5.903,3.215)--(6.043,3.132)--(6.184,3.879)--(6.324,3.298)%
  --(6.464,3.550)--(6.604,3.023)--(6.745,3.520)--(6.885,3.215)--(7.025,3.483)--(7.165,2.962)%
  --(7.306,3.093)--(7.446,3.226)--(7.586,3.314)--(7.726,3.146)--(7.867,2.749)--(8.007,3.026)%
  --(8.147,2.738)--(8.287,2.918)--(8.427,2.984)--(8.568,2.852)--(8.708,3.138)--(8.848,3.049)%
  --(8.988,3.080)--(9.129,3.185)--(9.269,3.455)--(9.409,2.843)--(9.549,2.951)--(9.690,2.955)%
  --(9.830,2.725)--(9.970,2.792)--(10.110,3.149)--(10.250,3.057)--(10.391,3.116)--(10.531,3.072)%
  --(10.671,2.998)--(10.811,3.046)--(10.952,3.355)--(11.092,3.472)--(11.232,3.639)--(11.372,3.310)%
  --(11.513,3.357)--(11.653,3.353)--(11.793,3.456)--(11.933,3.527)--(12.074,3.723)--(12.214,3.957)%
  --(12.354,3.904)--(12.494,4.325)--(12.634,4.163)--(12.775,4.196)--(12.915,4.109)--(13.047,4.131);
\gppoint{gp mark 7}{(11.653,3.353)}
\gppoint{gp mark 7}{(12.221,1.627)}
\gpcolor{color=gp lt color border}
\node[gp node right,font={\fontsize{8pt}{9.6pt}\selectfont}] at (11.579,1.319) {no averaging};
\gpcolor{rgb color={0.302,0.686,0.290}}
\gpsetlinetype{gp lt plot 4}
\draw[gp path] (11.763,1.319)--(12.679,1.319);
\draw[gp path] (1.688,2.679)--(1.696,2.713)--(1.836,2.016)--(1.977,2.528)--(2.117,2.345)%
  --(2.257,2.116)--(2.397,2.266)--(2.538,2.784)--(2.678,2.507)--(2.818,2.796)--(2.958,1.673)%
  --(3.099,1.621)--(3.239,2.272)--(3.379,2.050)--(3.519,2.081)--(3.659,2.986)--(3.800,2.761)%
  --(3.940,2.542)--(4.080,1.973)--(4.220,2.521)--(4.361,2.228)--(4.501,1.754)--(4.641,2.921)%
  --(4.781,2.716)--(4.922,1.814)--(5.062,2.203)--(5.202,2.349)--(5.342,2.583)--(5.483,2.579)%
  --(5.623,2.274)--(5.763,2.471)--(5.903,1.456)--(6.043,1.341)--(6.184,3.410)--(6.324,2.849)%
  --(6.464,1.786)--(6.604,2.876)--(6.745,2.395)--(6.885,2.155)--(7.025,2.473)--(7.165,2.175)%
  --(7.306,3.222)--(7.446,1.769)--(7.586,1.942)--(7.726,2.164)--(7.867,2.392)--(8.007,2.286)%
  --(8.147,2.601)--(8.287,2.937)--(8.427,2.007)--(8.568,3.158)--(8.708,2.268)--(8.848,2.665)%
  --(8.988,3.012)--(9.129,2.440)--(9.269,3.034)--(9.409,2.860)--(9.549,2.578)--(9.690,2.740)%
  --(9.830,2.574)--(9.970,2.239)--(10.110,2.681)--(10.250,2.933)--(10.391,2.423)--(10.531,2.452)%
  --(10.671,2.885)--(10.811,2.435)--(10.952,2.019)--(11.092,2.202)--(11.232,2.685)--(11.372,2.876)%
  --(11.513,3.008)--(11.653,2.470)--(11.793,3.236)--(11.933,2.807)--(12.074,3.104)--(12.214,3.634)%
  --(12.354,3.705)--(12.494,2.660)--(12.634,2.983)--(12.775,3.048)--(12.915,3.764)--(13.047,3.023);
\gppoint{gp mark 9}{(11.653,2.470)}
\gppoint{gp mark 9}{(12.221,1.319)}
\gpcolor{color=gp lt color border}
\gpsetlinetype{gp lt border}
\gpsetlinewidth{1.00}
\draw[gp path] (1.688,4.631)--(1.688,0.985)--(13.047,0.985)--(13.047,4.631)--cycle;
\gpdefrectangularnode{gp plot 1}{\pgfpoint{1.688cm}{0.985cm}}{\pgfpoint{13.047cm}{4.631cm}}
\end{tikzpicture}

%% file: ms.bbl
\begin{thebibliography}{25}
\providecommand{\natexlab}[1]{#1}
\providecommand{\url}[1]{\texttt{#1}}
\expandafter\ifx\csname urlstyle\endcsname\relax
  \providecommand{\doi}[1]{doi: #1}\else
  \providecommand{\doi}{doi: \begingroup \urlstyle{rm}\Url}\fi

\bibitem[Bahdanau et~al.(2015)Bahdanau, Cho, and
  Bengio]{bahdanau:etal:attention:iclr:2015}
Bahdanau, Dzmitry, Kyunghyun Cho, and Yoshua Bengio.
\newblock Neural Machine Translation by Jointly Learning to Align and
  Translate.
\newblock In \emph{Proceedings of ICLR}, 2015.

\bibitem[Bojar et~al.(2012)Bojar, {\v{Z}}abokrtsk{\'{y}}, Du{\v{s}}ek,
  Galu{\v{s}}{\v{c}}{\'{a}}kov{\'{a}}, Majli{\v{s}}, Mare{\v{c}}ek,
  Mar{\v{s}}{\'{\i}}k, Nov{\'{a}}k, Popel, and Tamchyna]{czeng10:lrec2012}
Bojar, Ond{\v{r}}ej, Zden{\v{e}}k {\v{Z}}abokrtsk{\'{y}}, Ond{\v{r}}ej
  Du{\v{s}}ek, Petra Galu{\v{s}}{\v{c}}{\'{a}}kov{\'{a}}, Martin Majli{\v{s}},
  David Mare{\v{c}}ek, Ji{\v{r}}{\'{\i}} Mar{\v{s}}{\'{\i}}k, Michal
  Nov{\'{a}}k, Martin Popel, and Ale{\v{s}} Tamchyna.
\newblock {The Joy of Parallelism with CzEng 1.0}.
\newblock In \emph{Proceedings of the Eighth International Language Resources
  and Evaluation Conference (LREC'12)}, pages 3921--3928, Istanbul, Turkey, May
  2012. ELRA, European Language Resources Association.
\newblock ISBN 978-2-9517408-7-7.

\bibitem[Bojar et~al.(2016)Bojar, Du{\v{s}}ek, Kocmi, Libovick{\'{y}},
  Nov{\'{a}}k, Popel, Sudarikov, and Vari{\v{s}}]{czeng16:2016}
Bojar, Ond{\v{r}}ej, Ond{\v{r}}ej Du{\v{s}}ek, Tom Kocmi, Jind{\v{r}}ich
  Libovick{\'{y}}, Michal Nov{\'{a}}k, Martin Popel, Roman Sudarikov, and
  Du{\v{s}}an Vari{\v{s}}.
\newblock {CzEng 1.6: Enlarged Czech-English Parallel Corpus with Processing
  Tools Dockered}.
\newblock In Sojka, Petr, Ale{\v{s}} Hor{\'{a}}k, Ivan Kope{\v{c}}ek, and Karel
  Pala, editors, \emph{{Text, Speech, and Dialogue: 19th International
  Conference, {TSD} 2016}}, number 9924 in Lecture Notes in Artificial
  Intelligence, pages 231--238. Masaryk University, Springer International
  Publishing, 2016.
\newblock ISBN 978-3-319-45509-9.

\bibitem[Bojar et~al.(2017{\natexlab{a}})Bojar, Chatterjee, Federmann, Graham,
  Haddow, Huck, Koehn, Logacheva, Monz, Negri, Post, Rubino, Specia, and
  Turchi]{bojar-EtAl:2017:WMT1}
Bojar, Ond{\v{r}}ej, Rajen Chatterjee, Christian Federmann, Yvette Graham,
  Barry Haddow, Matthias Huck, Philipp Koehn, Varvara Logacheva, Christof Monz,
  Matteo Negri, Matt Post, Raphael Rubino, Lucia Specia, and Marco Turchi.
\newblock {Findings of the 2017 Conference on Machine Translation (WMT17)}.
\newblock In \emph{{Proceedings of the Second Conference on Machine
  Translation}}, Copenhagen, Denmark, September 2017{\natexlab{a}}. ACL.

\bibitem[Bojar et~al.(2017{\natexlab{b}})Bojar, Graham, and
  Kamran]{bojar-graham-kamran:2017:WMT}
Bojar, Ond{\v{r}}ej, Yvette Graham, and Amir Kamran.
\newblock {Results of the WMT17 Metrics Shared Task}.
\newblock In \emph{{Proceedings of the Second Conference on Machine
  Translation}}, Copenhagen, Denmark, September 2017{\natexlab{b}}. ACL.

\bibitem[Bottou(2012)]{bottou-2012}
Bottou, L{\'e}on.
\newblock \emph{Stochastic Gradient Descent Tricks}, pages 421--436.
\newblock Springer Berlin Heidelberg, Berlin, Heidelberg, 2012.
\newblock ISBN 978-3-642-35289-8.
\newblock \doi{10.1007/978-3-642-35289-8_25}.
\newblock URL \url{https://doi.org/10.1007/978-3-642-35289-8_25}.

\bibitem[{Bottou} et~al.(2016){Bottou}, {Curtis}, and
  {Nocedal}]{bottou-et-al:2016}
{Bottou}, L., F.~E. {Curtis}, and J.~{Nocedal}.
\newblock {Optimization Methods for Large-Scale Machine Learning}.
\newblock \emph{ArXiv e-prints}, June 2016.
\newblock URL \url{https://arxiv.org/abs/1606.04838}.

\bibitem[Cettolo et~al.(2017)Cettolo, Federico, Bentivogli, Niehues,
  St{\"u}ker, Sudoh, Yoshino, and Federmann]{iwslt:overview:2017}
Cettolo, Mauro, Marcello Federico, Luisa Bentivogli, Jan Niehues, Sebastian
  St{\"u}ker, Katsuhito Sudoh, Koichiro Yoshino, and Christian Federmann.
\newblock {Overview of the IWSLT 2017 Evaluation Campaign}.
\newblock In \emph{{Proceedings of the 14th International Workshop on Spoken
  Language Translation (IWSLT)}}, pages 2--14, Tokyo, Japan, 2017.

\bibitem[Goyal et~al.(2017)Goyal, Doll{\'{a}}r, Girshick, Noordhuis,
  Wesolowski, Kyrola, Tulloch, Jia, and He]{goyal-et-al:2017}
Goyal, Priya, Piotr Doll{\'{a}}r, Ross~B. Girshick, Pieter Noordhuis, Lukasz
  Wesolowski, Aapo Kyrola, Andrew Tulloch, Yangqing Jia, and Kaiming He.
\newblock Accurate, Large Minibatch {SGD:} Training ImageNet in 1 Hour.
\newblock \emph{CoRR}, 2017.
\newblock URL \url{http://arxiv.org/abs/1706.02677}.

\bibitem[Hoffer et~al.(2017)Hoffer, Hubara, and Soudry]{hoffer-et-al-2017}
Hoffer, Elad, Itay Hubara, and Daniel Soudry.
\newblock Train longer, generalize better: closing the generalization gap in
  large batch training of neural networks.
\newblock In Guyon, I., U.~V. Luxburg, S.~Bengio, H.~Wallach, R.~Fergus,
  S.~Vishwanathan, and R.~Garnett, editors, \emph{Advances in Neural
  Information Processing Systems 30}, pages 1731--1741. Curran Associates,
  Inc., 2017.
\newblock URL
  \href{http://papers.nips.cc/paper/6770-train-longer-generalize-better-closing-the-generalization-gap-in-large-batch-training-of-neural-networks.pdf}{http://papers.nips.cc/paper/6770-train-longer-generalize-better-closing-the-generalization-gap-in-large-batch-training-of-neural-networks.pdf}.

\bibitem[Ioffe and Szegedy(2015)]{batch-norm}
Ioffe, Sergey and Christian Szegedy.
\newblock Batch Normalization: Accelerating Deep Network Training by Reducing
  Internal Covariate Shift.
\newblock \emph{CoRR}, abs/1502.03167, 2015.
\newblock URL \url{http://arxiv.org/abs/1502.03167}.

\bibitem[Jastrzebski et~al.(2017)Jastrzebski, Kenton, Arpit, Ballas, Fischer,
  Bengio, and Storkey]{jastrzebski-et-al:2017}
Jastrzebski, Stanislaw, Zachary Kenton, Devansh Arpit, Nicolas Ballas, Asja
  Fischer, Yoshua Bengio, and Amos~J. Storkey.
\newblock Three Factors Influencing Minima in {SGD}.
\newblock \emph{CoRR}, abs/1711.04623, 2017.
\newblock URL \url{http://arxiv.org/abs/1711.04623}.

\bibitem[Keskar et~al.(2017)Keskar, Mudigere, Nocedal, Smelyanskiy, and
  Tang]{keskar:etal:minibatches:iclr:2017}
Keskar, Nitish~Shirish, Dheevatsa Mudigere, Jorge Nocedal, Mikhail Smelyanskiy,
  and Ping Tak~Peter Tang.
\newblock On Large-Batch Training for Deep Learning: Generalization Gap and
  Sharp Minima.
\newblock In \emph{Proceedings of ICLR}, 2017.
\newblock URL \url{http://arxiv.org/abs/1609.04836}.

\bibitem[Krizhevsky(2014)]{krizhevsky:2014}
Krizhevsky, Alex.
\newblock One weird trick for parallelizing convolutional neural networks.
\newblock \emph{CoRR}, abs/1404.5997, 2014.
\newblock URL \url{http://arxiv.org/abs/1404.5997}.

\bibitem[Lee et~al.(2016)Lee, Cho, and Hofmann]{lee:etal:charlevel:2016}
Lee, Jason, Kyunghyun Cho, and Thomas Hofmann.
\newblock Fully Character-Level Neural Machine Translation without Explicit
  Segmentation.
\newblock \emph{CoRR}, 2016.
\newblock URL \url{http://arxiv.org/abs/1610.03017}.

\bibitem[{Lei Ba} et~al.(2016){Lei Ba}, {Kiros}, and {Hinton}]{layer-norm}
{Lei Ba}, J., J.~R. {Kiros}, and G.~E. {Hinton}.
\newblock {Layer Normalization}.
\newblock \emph{ArXiv e-prints}, July 2016.

\bibitem[Papineni et~al.(2002)Papineni, Roukos, Ward, and Zhu]{papineni:2002}
Papineni, Kishore, Salim Roukos, Todd Ward, and Wei-Jing Zhu.
\newblock {BLEU: a Method for Automatic Evaluation of Machine Translation}.
\newblock In \emph{{Proceedings of ACL 2002}}, pages 311--318, Philadelphia,
  Pennsylvania, 2002.

\bibitem[Popovi\'{c}(2015)]{popovic-2015}
Popovi\'{c}, Maja.
\newblock chrF: character n-gram F-score for automatic MT evaluation.
\newblock In \emph{Proceedings of the Tenth Workshop on Statistical Machine
  Translation}, pages 392--395, Lisbon, Portugal, September 2015. ACL.
\newblock URL \url{http://aclweb.org/anthology/W15-3049}.

\bibitem[Sennrich et~al.(2016)Sennrich, Haddow, and
  Birch]{sennrich-haddow-birch:2016:P16-12}
Sennrich, Rico, Barry Haddow, and Alexandra Birch.
\newblock Neural Machine Translation of Rare Words with Subword Units.
\newblock In \emph{Proceedings of ACL 2016}, pages 1715--1725, Berlin, Germany,
  August 2016. ACL.
\newblock URL \url{http://www.aclweb.org/anthology/P16-1162}.

\bibitem[{Shazeer} and {Stern}(2018)]{adafactor}
{Shazeer}, N. and M.~{Stern}.
\newblock {Adafactor: Adaptive Learning Rates with Sublinear Memory Cost}.
\newblock \emph{ArXiv e-prints}, Apr. 2018.
\newblock URL \url{https://arxiv.org/abs/1804.04235}.

\bibitem[Smith and Le(2017)]{smith:le:generalization:2017}
Smith, Samuel~L. and Quoc~V. Le.
\newblock A Bayesian Perspective on Generalization and Stochastic Gradient
  Descent.
\newblock In \emph{Proceedings of Second workshop on Bayesian Deep Learning
  (NIPS 2017)}, Long Beach, CA, USA, 2017.
\newblock URL \url{http://arxiv.org/abs/1710.06451}.

\bibitem[Smith et~al.(2017)Smith, Kindermans, and
  Le]{smith:etal:lr:batchsize:arxiv:2017}
Smith, Samuel~L., Pieter{-}Jan Kindermans, and Quoc~V. Le.
\newblock Don't Decay the Learning Rate, Increase the Batch Size.
\newblock \emph{CoRR}, 2017.
\newblock URL \url{http://arxiv.org/abs/1711.00489}.

\bibitem[Vaswani et~al.(2017)Vaswani, Shazeer, Parmar, Uszkoreit, Jones, Gomez,
  Kaiser, and Polosukhin]{vaswani-et-al:2017}
Vaswani, Ashish, Noam Shazeer, Niki Parmar, Jakob Uszkoreit, Llion Jones,
  Aidan~N Gomez, {\L}ukasz Kaiser, and Illia Polosukhin.
\newblock Attention is All you Need.
\newblock In Guyon, I., U.~V. Luxburg, S.~Bengio, H.~Wallach, R.~Fergus,
  S.~Vishwanathan, and R.~Garnett, editors, \emph{Advances in Neural
  Information Processing Systems 30}, pages 6000--6010. Curran Associates,
  Inc., 2017.
\newblock URL
  \url{http://papers.nips.cc/paper/7181-attention-is-all-you-need.pdf}.

\bibitem[Wu et~al.(2016)Wu, Schuster, Chen, Le, Norouzi, Macherey, Krikun, Cao,
  Gao, Macherey, Klingner, Shah, Johnson, Liu, Kaiser, Gouws, Kato, Kudo,
  Kazawa, Stevens, Kurian, Patil, Wang, Young, Smith, Riesa, Rudnick, Vinyals,
  Corrado, Hughes, and Dean]{google:bridging:gap:2016:arxiv}
Wu, Yonghui, Mike Schuster, Zhifeng Chen, Quoc~V. Le, Mohammad Norouzi,
  Wolfgang Macherey, Maxim Krikun, Yuan Cao, Qin Gao, Klaus Macherey, Jeff
  Klingner, Apurva Shah, Melvin Johnson, Xiaobing Liu, Lukasz Kaiser, Stephan
  Gouws, Yoshikiyo Kato, Taku Kudo, Hideto Kazawa, Keith Stevens, George
  Kurian, Nishant Patil, Wei Wang, Cliff Young, Jason Smith, Jason Riesa, Alex
  Rudnick, Oriol Vinyals, Greg Corrado, Macduff Hughes, and Jeffrey Dean.
\newblock Google's Neural Machine Translation System: Bridging the Gap between
  Human and Machine Translation.
\newblock \emph{CoRR}, abs/1609.08144, 2016.
\newblock URL \url{http://arxiv.org/abs/1609.08144}.

\bibitem[You et~al.(2017)You, Gitman, and Ginsburg]{you-et-al:2017}
You, Yang, Igor Gitman, and Boris Ginsburg.
\newblock Scaling {SGD} Batch Size to 32K for ImageNet Training.
\newblock \emph{CoRR}, abs/1708.03888, 2017.
\newblock URL \url{http://arxiv.org/abs/1708.03888}.

\end{thebibliography}
